\documentclass[twoside,11pt]{article}

\usepackage{blindtext}

\usepackage{jmlr2e}

\usepackage{wrapfig}
\usepackage{microtype}
\usepackage{graphicx}
\usepackage{subfigure}
\usepackage{booktabs} 
\usepackage{amsmath}
\usepackage{amssymb}
\usepackage{mathtools}
\usepackage{paralist}
\usepackage{multirow}
\usepackage{textcomp}
\usepackage{stfloats}
\usepackage{url}
\usepackage{verbatim}
\usepackage{hyperref}
\hypersetup{
    colorlinks,
    breaklinks,
    linkcolor = blue,
    citecolor = blue,
    urlcolor  = blue,
}

\DeclareMathOperator*{\argmin}{arg\,min}
\usepackage{bbding}
\usepackage{booktabs} 
\usepackage{amsfonts} 
\usepackage{nicefrac} 
\usepackage{microtype} 
\usepackage{xcolor} 
\usepackage{graphicx}
\usepackage{makecell}
\usepackage{algorithm,algorithmic}
\usepackage{paralist,amsmath, amssymb,bm}
\usepackage{multirow}

\newtheorem{assumption}{Assumption}
\newtheorem{rem}{Remark}
\usepackage{enumitem}
\usepackage{textcase}
\usepackage{booktabs}
\usepackage{fancybox}
\newcommand{\Norm}[1]{\left\|#1\right\|}
\def \E {\mathbb{E}}
\def \R {\mathbb{R}}

\def \g {\mathbf{g}}

\def \L {\mathbf{L}}
\def \tr {\operatorname{tr}}

\def \n {\mathbf{n}}

\def \v {\mathbf{v}}
\def \w {\mathbf{w}}
\def \x {\mathbf{x}}
\def \y {\mathbf{y}}

\def \A {\mathcal{A}}

\usepackage{lastpage}

\ShortHeadings{Efficient  Preconditioned Optimization Methods with Frequent Directions}{Yang, Wan, Wang, and Zhang}
\firstpageno{1}

\begin{document}

\title{Efficient  Preconditioned Optimization Methods\\ with Frequent Directions}

\author{\name Sifan Yang \email yangsf@lamda.nju.edu.cn \\
       \addr State Key Laboratory of Novel Software Technology, Nanjing University, Nanjing, China\\
School of Artificial Intelligence, Nanjing University, Nanjing, China
        \AND
       \name Yuanyu Wan \email wanyy@zju.edu.cn \\
       \addr School of Software Technology, Zhejiang University, Ningbo, China\\
State Key Laboratory of Blockchain and Data Security, Zhejiang University, Hangzhou, China
       \AND
       \name Yibo Wang \email wangyb@lamda.nju.edu.cn \\
    \name Lijun Zhang \email zhanglj@lamda.nju.edu.cn \\
       \addr State Key Laboratory of Novel Software Technology, Nanjing University, Nanjing, China\\
School of Artificial Intelligence, Nanjing University, Nanjing, China
       }


\maketitle

\begin{abstract}%
Adaptive optimization methods have received considerable attention in recent years due to their strong empirical performance. However, they generally require maintaining a preconditioning matrix, which incurs prohibitive computational overhead in large-scale applications. To address these limitations, we investigate the acceleration of preconditioned optimization methods through frequent directions (FD), a widely used matrix sketching technique.  We first propose FTSL by integrating FD with adaptive subgradient methods under \textit{the primal-dual framework} and adding back the cumulative information discarded by FD. FTSL obtains a tighter regret bound than existing state-of-the-art results.  We then apply a doubling trick that further reduces the time complexity of FTSL, while maintaining the same regret guarantee.
We also consider optimization over matrix-valued decision variables and exploit their structural properties to construct sketched preconditioners. We develop FTSL-Shampoo by adapting FD to Shampoo and give a novel analysis under the primal-dual framework. To further mitigate its computational complexity, we design one-sided variants of FTSL-Shampoo under both mirror descent and primal-dual frameworks. Our algorithms not only reduce the computational cost of FTSL-Shampoo, but also achieve tighter regret bounds. We extend our methods to offline stochastic convex optimization and exploit smoothness to achieve faster convergence rates.  Finally, we introduce practical variants of our algorithms and empirically demonstrate their efficiency and effectiveness across a broad range of tasks.
\end{abstract}

\section{Introduction}
Adaptive optimization methods have been extensively studied over the past decade, owing to their ability to reduce the need for laborious learning-rate tuning while retaining theoretical guarantees comparable to those of carefully tuned algorithms \citep{mcmahan2010adaptive,coltduchi,gupta2017unified,kingma2014adam,agarwal2019efficient}. The pioneering work of \cite{duchi2011adaptive} introduces adaptive subgradient methods with full matrices (ADA-FULL) within both the mirror descent framework \citep{duchi2010composite} and the primal-dual  framework \citep{xiao2009dual}.  However, ADA-FULL requires maintaining a preconditioning matrix of shape $d\times d$, resulting in $O(d^2)$ space complexity and $O(d^3)$ time complexity per round, where $d$ is the dimensionality. Consequently, ADA-FULL is impractical for large-scale machine learning tasks involving high-dimensional data.

Meanwhile, in many modern machine learning applications, especially in deep learning tasks, the optimization variables are naturally matrix-valued \citep{lecun2015deep}, i.e., $X_t \in \R^{m\times n}$. Directly applying ADA-FULL requires vectorizing the parameter matrix, which overlooks the matrix structure and still incurs excessive space and time complexities. This limitation has motivated a growing body of work on structured preconditioners that exploit the geometry of matrix-valued variables \citep{gupta2018shampoo,anil2020scalable,vyas2025soap,xie25icml,an2025asgo}. As a representative example, Shampoo \citep{gupta2018shampoo} constructs a Kronecker-factored preconditioner along each axis of the gradient matrix, which lowers the memory and computational cost of ADA-FULL.  Nevertheless, its regret guarantee depends on the rank of the gradients. In addition, maintaining a separate preconditioner for each axis can still incur considerable overhead.    
Thus, a natural question arises: \textit{can we develop efficient preconditioned optimization methods while retaining the performance of full-matrix methods?}

In this paper, we provide an affirmative answer to this question by systematically investigating computationally efficient preconditioned optimization methods via frequent directions (FD) \citep{ghashami2016frequent}. Let $G_T$ denote the sum of gradient outer products over $T$ rounds, $\tau \ll d$ denote the sketching size, and $\rho_t$ denote the eigenvalue discarded by FD in round $t$. Although several studies have developed efficient variants of ADA-FULL with FD \citep{IJCAI:2018:Wan,wan2021efficient,feinberg2024sketchy}, the state-of-the-art algorithm, S-ADA \citep{feinberg2024sketchy}, still suffers from two drawbacks: \textit{(i)} its regret bound of $O(\operatorname{tr}(G_T^{1/2}) + \sqrt{d(d-\tau)\rho_{1:T}})$ grows \textit{linearly} with $d$; \textit{(ii)} its $O(\tau^2 d)$ time complexity depends \textit{quadratically} on the sketching size $\tau$.  To overcome these drawbacks, we first develop an algorithm termed \textbf{F}ollow-\textbf{t}he-\textbf{S}ketchy-\textbf{L}eader~(FTSL). The key idea is to integrate FD with ADA-FULL under \textit{the primal-dual framework} and add the cumulative discarded eigenvalues back into the preconditioner. FTSL  attains a \textit{dimension-free} regret bound of $O(\operatorname{tr}(G_T^{1/2}) + \sqrt{\rho_{1:T}})$. We further propose an accelerated variant named FTFSL, which doubles the sketching size to lower the frequency of expensive SVD computations. FTFSL retains the same regret bound and the space complexity of FTSL, while reducing the per-round time complexity to $O(\tau d)$ when $\tau \leq \sqrt{d}$. This acceleration technique can also be applied to S-ADA, though its regret bound remains inferior to that of FTFSL. We summarize our results  in Table~\ref{table:1} and Table~\ref{table:3}.

\begin{table*}[t!]
\centering
\small
\renewcommand{\arraystretch}{2}
\caption{Comparison of  ADA-FULL and its FD-based variants, where (M) and (P) represent the primal-dual  framework and the mirror descent framework, respectively.  }
\begin{tabular}{cccc}
\specialrule{1pt}{0pt}{0pt} 
\textbf{$\qquad$$\quad$Algorithms$\qquad$$\quad$} & {$\qquad$\textbf{Regret Bounds}$\qquad$} & {\textbf{Space}} & {\textbf{Time}} \\ 
\hline
ADA-FULL \citep{duchi2011adaptive} & $O\left(\operatorname{tr}\left(G_T^{1/2}\right)\right)$ & $O(d^2)$ & $O(d^3)$  \\
\hline
ADA-FD (P) \citep{IJCAI:2018:Wan}& $O(\operatorname{tr}(G_T^{1/2}) + \sum_{t=1}^T\textcolor{red}{\sqrt{\rho_{t}}})$ & $O(\tau d)$ & $O(\tau^2d)$  \\
\hline
ADA-FD (M) \citep{IJCAI:2018:Wan}& $O(\operatorname{tr}(G_T^{1/2}) + \sum_{t=1}^T\textcolor{red}{\tau}\textcolor{red}{\sqrt{\rho_{t}}})$ & $O(\tau d)$ & $O(\tau^2d)$  \\
\hline
S-ADA (M) \citep{feinberg2024sketchy}& $O(\operatorname{tr}(G_T^{1/2}) + \sqrt{\textcolor{red}{d(d-\tau)}\rho_{1:T}})$ & $O(\tau d)$ & $O(\tau^2d)$  \\
\hline
\textbf{FTSL (P)} (this work) &  $O(\operatorname{tr}(G_T^{1/2}) + \sqrt{\rho_{1:T}})$ & $O(\tau d)$ & $O(\tau^2d)$  \\
\specialrule{1pt}{0pt}{0pt}
\end{tabular}
\label{table:1}
\end{table*}

\begin{table*}[t!]
\centering
\small
\renewcommand{\arraystretch}{2}
\caption{Comparison of FD-based variants of ADA-FULL, where (M) and (P) represent the primal-dual  framework and the mirror descent framework, respectively.}
\begin{tabular}{cccc}
\specialrule{1pt}{0pt}{0pt} 
\textbf{$\qquad$$\quad$Algorithms$\qquad$$\quad$} & {$\qquad$\textbf{Regret Bounds}$\qquad$} & {\textbf{Space}} & {\textbf{Time}} \\ 
\hline
ADA-FFD (P) \citep{wan2021efficient} & $O(\operatorname{tr}(G_T^{1/2}) + \sum_{t=1}^T\textcolor{red}{\sqrt{\rho_{t}}})$ & $O(\tau d)$ & $O(\tau d)$  \\
\hline
ADA-FFD (M) \citep{wan2021efficient}& $O(\operatorname{tr}(G_T^{1/2}) + \sum_{t=1}^T\textcolor{red}{\tau} \textcolor{red}{\sqrt{\rho_{t}}})$ & $O(\tau d)$ & $O(\tau d)$  \\
\hline
\textbf{Fast S-ADA (M) } (this work)& $O(\operatorname{tr}(G_T^{1/2}) + \sqrt{{d(d-\tau)}\rho_{1:T}})$ & $O(\tau d)$ & $O(\tau d)$  \\
\hline
\textbf{FTFSL (P) } (this work) & $O(\operatorname{tr}(G_T^{1/2}) + \sqrt{\rho_{1:T}})$ & $O(\tau d)$ & $O(\tau d)$  \\
\specialrule{1pt}{0pt}{0pt} 
\end{tabular}
\label{table:3}
\end{table*}

\begin{table*}[t!]
\centering
\footnotesize
\caption{Comparison of adaptive methods when the decision has a matrix structure $X_t \in \mathbb{R}^{m\times n}$, where (M) and (P) represent the primal-dual  framework and the mirror descent framework, respectively. Without loss of generality,  we assume $m \leq n$. We denote the gradient by $G_t^{X} \in \mathbb{R}^{m \times n }$ and let $r$ be the largest rank of $G^{X}_t$.}
\renewcommand{\arraystretch}{2.6}
\resizebox{\textwidth}{!}{
\begin{tabular}{cccc}
\specialrule{1pt}{0pt}{0pt} 
\textbf{Algorithms} & {\textbf{Regret Bounds}} & {\textbf{Space}} & {\textbf{Time}} \\ 
\hline
\multirow{1}{*}{\makecell{Shampoo \\ \citep{gupta2018shampoo}}}& $O(\sqrt{r} \operatorname{tr}(L_T^{1/4}) \operatorname{tr}(R_T^{1/4}))$ & $O(m^2 + n^2)$ & $O(m^3 + n^3)$  \\ 
\hline
\multirow{1}{*}{\makecell{S-Shampoo (M)  \\ \citep{feinberg2024sketchy}}} & $O(\sqrt{r} (\operatorname{tr}(L_T^{1/4}) + \textcolor{red}{m} (\rho_{1:T}^{L})^{1/4}) (\operatorname{tr}(R_T^{1/4}) + \textcolor{red}{n} (\rho_{1:T}^R)^{1/4}))$ & $O(\tau (m+n))$ & $O(\tau^2 mn)$  \\
\hline
\multirow{1}{*}{\makecell{\textbf{FTSL-Shampoo (P) } \\ (this work)}} & $O(\sqrt{r}(\operatorname{tr}(L_T^{1/4}) + (\rho_{1:T}^{L})^{1/4})(\operatorname{tr}(R_T^{1/4}) + (\rho_{1:T}^R)^{1/4}))$  & $O(\tau (m+n)) $ & $O(\tau^2 mn)$ \\
\hline
\multirow{1}{*}{\makecell{ASGO\\ \citep{an2025asgo}}}   & $O\left(\operatorname{tr}(L_T^{1/2})\right)$  & $O(m^2) $ & $O(m^3)$ \\
\hline
\multirow{1}{*}{\makecell{\textbf{S-ASGO (M)} \\ (this work)}}  & $O\left(\operatorname{tr}(L_T^{1/2}) + m\sqrt{\rho_{1:T}^{L}}\right)$  & $O(\tau m) $ & $O(\tau^2 mn)$ \\
\hline
\multirow{1}{*}{\makecell{\textbf{S-ASGO (P)} \\ (this work)}}  & $O\left(\operatorname{tr}(L_T^{1/2}) + \sqrt{\rho_{1:T}^{L}}\right)$  & $O(\tau m) $ & $O(\tau^2 mn)$ \\
\specialrule{1pt}{0pt}{0pt} 
\end{tabular}
}
\label{table:2}
\end{table*}  

Second, we consider optimization over matrix-valued decision variables, a scenario commonly encountered in deep learning tasks. Let $G_t^X \in \mathbb{R}^{m\times n}$ be the gradient at round $t$, $r$ be the upper bound on the rank of $G_t^X$, $L_T = \sum_{t=1}^T G_t^X (G_t^X)^\top$ and $R_T = \sum_{t=1}^T (G_t^X)^\top G_t^X$ be its left and right covariance matrices. \cite{feinberg2024sketchy} applies FD to Shampoo  and reduces its space complexity to $O(\tau(m+n))$. Nevertheless, its regret bound of $O(\sqrt{r}(\operatorname{tr}(L_T^{1/4}) + m(\rho_{1:T}^L)^{1/4})(\operatorname{tr}(R_T^{1/4}) + n(\rho_{1:T}^R)^{1/4}))$ again depends on the dimensions $m$ and $n$, where $\rho_{1:T}^L$ and $\rho_{1:T}^R$ denote the cumulative eigenvalues discarded by FD when sketching $L_T$ and $R_T$, respectively. To eliminate this dimension dependence, we develop FTSL-Shampoo by combining FD with a primal-dual variant of Shampoo and provide a novel analysis. FTSL-Shampoo attains a dimension-free regret bound of $O(\sqrt{r}(\operatorname{tr}(L_T^{1/4}) + (\rho_{1:T}^L)^{1/4})(\operatorname{tr}(R_T^{1/4}) + (\rho_{1:T}^R)^{1/4}))$, while preserving the same space complexity. Despite this improvement, FTSL-Shampoo still maintains a separate preconditioner for each axis. Consequently, when extended to higher order parameters, its memory and computational overhead grows with the number of axes.

To handle the aforementioned limitations, we incorporate FD into ASGO \citep{xie25icml,an2025asgo} and propose Sketchy ASGO (S-ASGO). Instead of keeping the preconditioner for each axis, S-ASGO maintains a low-rank approximation of the preconditioner only along the smaller dimension, which reduces the space complexity of FTSL-Shampoo from $O(\tau(m+n))$ to $O(\tau m)$. We provide a novel analysis of S-ASGO under both the mirror descent and the primal-dual frameworks and establish the regret bounds $O(\operatorname{tr}(L_T^{1/2}) + m\sqrt{\rho_{1:T}^L})$ and $O(\operatorname{tr}(L_T^{1/2}) + \sqrt{\rho_{1:T}^L})$, respectively. Both bounds are independent of the rank $r$, and their leading term $O(\operatorname{tr}(L_T^{1/2}))$ is no larger than that of FTSL-Shampoo, which indicates that a single preconditioner already suffices to capture the dominant curvature information. Furthermore, we extend all of the above methods to the offline stochastic convex optimization setting. We provide convergence rates for both non-smooth and smooth objectives, and exploit the smoothness of the objective to attain a faster rate.  A detailed comparison with prior methods \citep{gupta2018shampoo,feinberg2024sketchy,an2025asgo} is given in Table~\ref{table:2}.  To reduce the runtime overhead of SVD in practice, we adopt a more efficient singular-value computation procedure \citep{gong2026aronewlensmatrix} and develop a practical variant of S-ASGO for large-scale model training. Extensive experiments demonstrate that our methods attain performance comparable to that of existing approaches while consistently reducing computational cost.

A preliminary version of this paper was presented at the 42nd International Conference on Machine Learning (ICML 2025) \citep{yangdimension}. The conference version focused on the online setting and primarily built on ADA-FULL, developing FTSL, FTFSL and Fast S-ADA for vector-valued decisions, together with FTSL-Shampoo for matrix-valued decisions. In this paper, we significantly expand the conference version in terms of algorithm design, theoretical analysis, and empirical studies. The main extensions are summarized as follows.

\begin{compactenum}
    \item We propose S-ASGO by incorporating FD into ASGO under both the mirror descent and the primal-dual frameworks (Algorithm~\ref{alg:sasgo}). S-ASGO reduces the space complexity of FTSL-Shampoo from $O(\tau(m+n))$ to $O(\tau m)$, while attaining tighter regret bounds (Theorems~\ref{thm:jmlr1} and~\ref{thm:jmlr1md}).
    \item We extend all of our methods to the offline stochastic convex optimization setting and establish convergence rates for both non-smooth and smooth objectives (Theorems~\ref{thm:p1}, \ref{thm:p2}, \ref{thm:p4}, and \ref{thm:jmlr2}). The dimension-free advantage of our online analysis carries over to the offline setting, and our methods attain faster rates on smooth objectives.
    \item We develop a practical variant of S-ASGO tailored to large-scale model pretraining (Algorithm~\ref{alg:sasgo2}). This variant substantially improves the efficiency of S-ASGO while preserving its empirical effectiveness.
    \item We conduct extensive experiments to validate the effectiveness and efficiency of our methods, in particular on large language model pretraining tasks.
\end{compactenum}

\paragraph{Organization.} The remainder of the paper is structured 
as follows. Section~\ref{section2} briefly reviews related work. 
Section~\ref{section3} presents the preliminaries, including the notation, assumptions, as well as the technical 
tools that underpin our analysis. Section~\ref{section4} illustrates 
our main results, together with their extension to
offline stochastic optimization. Section~\ref{section5} establishes 
our results for optimization with matrix structure. 
Section~\ref{section6} provides the analysis of all theorems. 
Section~\ref{section7} empirically validates the efficiency of our 
algorithms through experiments on a variety of tasks. 
Section~\ref{section8} concludes the paper and discusses future work.

\section{Related Work}\label{section2}
In this section, we briefly review the related work on adaptive subgradient methods, their efficient variants based on matrix sketching,  and optimization with matrix structure.

\subsection{Online Convex Optimization and Adaptive Subgradient Methods}
Online convex optimization (OCO) is a powerful paradigm for solving sequential decision-making problems \citep{hazan2016introduction,orabona2019modern,NeurIPS:2018:Zhang:B,ICML:2022:Zhang}. Specifically, it is typically formulated as an iterative game between a player and an adversary. In each round $t \in [T]$, the player begins by selecting a decision $\x_t \in \mathbb{R}^d$. After that, the adversary chooses a convex loss function $f_t(\cdot)\colon \mathbb{R}^d  \mapsto \mathbb{R}$, and the player incurs a loss $f_t(\x_t)$.  The goal of the player is to minimize the cumulative loss $\sum_{t=1}^T f_t(\x_t)$ over $T$ rounds, which is equivalent to minimizing the regret \citep{zinkevich2003online}
\begin{equation}\label{regret}
    R(T) = \sum_{t=1}^T f_t(\x_t) - \sum_{t=1}^T f_t(\mathbf{x}^*),
\end{equation}
defined as the excess loss suffered by the player compared to the loss of the fixed optimal choice $\x^* \in \arg\min_{\mathbf{x} \in \mathbb{R}^d} \sum_{t=1}^T f_t(\mathbf{x})$. Although \cite{zinkevich2003online} establishes the optimal regret bound of  $O(\sqrt{T})$ for convex loss functions, it is data-independent. In the following, we will introduce ADAGRAD \citep{coltduchi,duchi2011adaptive}, a widely-used adaptive subgradient method,  in both the primal-dual subgradient framework \citep{xiao2009dual} and the mirror descent framework \citep{duchi2010composite}, which achieves a data-dependent regret bound.

ADAGRAD can be categorized into two forms based on how the preconditioner $\tilde{G}_t$ is computed: ADAGRAD with full matrices (ADA-FULL) and ADAGRAD with diagonal matrices (ADA-DIAG). Let $\g_t$ be a particular vector in the subdifferential set $\partial f_t(\x_t)$. Since we do not require the loss function to be smooth, we will not explicitly distinguish subgradients and gradients in the  subsequent discussion. ADA-FULL first calculates the outer product matrix of the past gradients $G_t = \sum_{i=1}^t \g_i \g_i^{\top}$, and further defines a symmetric matrix $\tilde{G}_t =\epsilon  I_{d\times d} + G_t^{1/2}$, where $\epsilon > 0$ is a hyper-parameter introduced to ensure the invertibility of $\tilde{G}_t$.  According to the primal-dual  framework, the update rule is given by
\begin{align*}
\x_{t+1} &= \arg\min_{\x \in \mathbb{R}^d} \left\{ \eta \left\langle \frac{1}{t} \overline{\g}_t, \x \right\rangle + \ \frac{1}{t} \Psi_t(\x) \right\} \\
&= -\eta \tilde{G}_t^{-1} \overline{\g}_t,
\end{align*}
where $\eta$ is the learning rate, $\overline{\g}_t = \sum_{i=1}^t \g_i$ is the sum of the received gradients and $\Psi_t(\x) = \frac{1}{2}\langle \x, \tilde{G}_t\x\rangle$ is the proximal term. The mirror descent version updates the decision as follows
\begin{align*}
\x_{t+1} &= \arg\min_{\x \in \mathbb{R}^d} \left\{ \eta \left\langle \g_t, \x \right\rangle +  B_{\Psi_t}(\x,\x_t) \right\} \\
&= \x_t -\eta \tilde{G}_t^{-1} \g_t,
\end{align*}
where $B_{\Psi_t}(\mathbf{x}, \mathbf{y}) = \Psi_t(\mathbf{x}) - \Psi_t(\mathbf{y}) - \left\langle \nabla \Psi_t(\mathbf{y}), \mathbf{x} - \mathbf{y} \right\rangle$ is the Bregman divergence associated with $\Psi_t (\cdot)$. ADA-FULL achieves an $O(\operatorname{tr}(G_T^{1/2}))$  regret bound within  both frameworks. However, ADA-FULL stores the past gradient outer products, i.e., $G_t$, and computes $G_t^{-1/2}$, resulting in $O(d^2)$ and $O(d^3)$ space and time complexities, respectively.  The  high computational cost of ADA-FULL prohibits its out-of-the-box use in typical machine learning problems, such as training neural networks \citep{sagun2017empirical,ghorbani2019investigation,sankar2021deeper}.

Different from ADA-FULL, ADA-DIAG only utilizes the diagonal elements of the gradient outer product matrix, i.e., redefining $ \tilde{G}_t = \epsilon I_{d \times d} + \operatorname{diag}(G_t)^{1/2}$, which thus is computationally more efficient. However, since the preconditioning matrix of ADA-DIAG only contains limited information, the regret bound of ADA-DIAG is worse than that of ADA-FULL when high-dimensional data is dense and has a low-rank structure.

\subsection{Adaptive Subgradient Methods with Matrix Sketching}
To alleviate the computational burden of ADA-FULL, several works have employed the sketching techniques to reduce its  space and time complexities \citep{krummenacher2016scalable,PAKDD:2018:Wan,IJCAI:2018:Wan,wan2021efficient,feinberg2024sketchy}.  \cite{krummenacher2016scalable}  propose ADA-LR to enhance the computational efficiency of ADA-FULL  by using random projection \citep{indyk1998approximate,achlioptas2003database}. While ADA-LR reduces the time complexity to $O(\tau d^2)$, its space complexity remains at $O(d^2)$, where  $\tau \ll d$ is the sketching size. To further improve the efficiency, they develop RADAGRAD, which incorporates a more randomized approximation, achieving space and time complexities of $O(\tau d)$ and $O(\tau^2 d)$, respectively. However, RADAGRAD  is not supported by rigorous theoretical analysis. Later, \cite{PAKDD:2018:Wan} develop ADA-DP based on  random projection, which achieves space and time complexities of $O(\tau d)$ and $O(\tau^2 d)$, while providing theoretical guarantees.

Another class of sketching-based adaptive subgradient methods adopts frequent directions (FD) \citep{ghashami2016frequent}, a stable matrix sketching technique. \cite{IJCAI:2018:Wan} first apply FD with ADA-FULL under both the  primal-dual subgradient framework and the mirror descent framework by maintaining a  matrix $B_t \in \mathbb{R}^{d \times \tau}$, such that $B_tB_t^{\top} \approx G_t\in \mathbb{R}^{d\times d}$.  Their approach, ADA-FD, obtains  space and time complexities of $O(\tau d)$ and $O(\tau^2 d)$, respectively. ADA-FD achieves regret bounds of $O(\operatorname{tr}(G_T^{1/2}) + \sum_{t=1}^T\sqrt{\textcolor{black}{\rho_{t}}})$ and $O(\operatorname{tr}(G_T^{1/2}) + \sum_{t=1}^T \tau\sqrt{\textcolor{black}{\rho_{t}}})$ under the  primal-dual subgradient framework and the mirror descent framework, where $\rho_t$ is the removed eigenvalue of FD in round $t$. Furthermore, \cite{wan2021efficient} introduce  a fast variant of ADA-FD, named ADA-FFD,  by doubling the sketching size. ADA-FFD improves the time complexity to $O(\tau d)$ while keeping the same regret bounds. Although ADA-FD and ADA-FFD enjoy  better space and time complexities, as mentioned by \cite{feinberg2024sketchy},  their regret bounds are $\Omega(T^{3/4})$ in some cases. For this reason, \cite{feinberg2024sketchy} propose S-ADA by adding back the discarded information of FD to the FD-based preconditioner instead of utilizing a fixed diagonal regularization. While S-ADA enjoys an $O(\operatorname{tr}(G_T^{1/2}) +\sqrt{d(d-\tau)\rho_{1:T}})$ regret bound,  it suffers from a linear dependence on the dimensionality $d$. Moreover, it only achieves an unsatisfactory time complexity of $O(\tau^2 d)$.

 Additionally,  FD has been utilized to accelerate the online Newton step (ONS) algorithm \citep{hazan2007logarithmic} for exponentially concave functions, and the LinUCB algorithm \citep{chu2011contextual} in the linear contextual bandit setting, both of which also need to maintain a covariance matrix. \citet{luo2016efficient} first apply FD in ONS to construct a low-rank approximation of the  matrix.  To reduce the approximation error of FD, \citet{luo2019robust} develop a new sketching scheme called robust frequent directions (RFD), which is the first method that compensates the discarded singular values  back into the second-order matrix.  They utilize RFD to propose a hyperparameter-free variant of ONS, which is more robust than FD-SON. In the linear contextual bandit setting, \citet{ijcai2020p588} design spectral compensation frequent directions (SCFD) to approximate the covariance matrices, which adds up the total mass of subtracted values during the FD procedure. SCFD can approximate a sequence of incremental covariance matrices while keeping the positive definite monotonicity. 

\subsection{Adaptive Preconditioned Optimization with Matrix Structure}\label{section:shampoo}

Optimization over matrix-valued decision variables has emerged as
a central topic in modern machine learning, owing to the ubiquity
of matrix-structured parameters in deep neural networks \citep{lecun2015deep}. Shampoo \citep{gupta2018shampoo} is an adaptive optimization method that first takes the structure of the parameter space into  consideration and thus is more efficient than ADA-FULL in scenarios where the decision is a matrix. Shampoo has attracted significant attention due to its \emph{Kronecker-factored} preconditioning strategy.
Specifically, Shampoo maintains a set of preconditioning matrices, each of which operates on one dimension, while aggregating information across the remaining dimensions. For example, for a parameter matrix $X_t \in \mathbb{R}^{m\times n}$ and its gradient  $G_t^X \in \mathbb{R}^{m \times n}$, ADA-FULL  treats the matrix-shaped gradient as a vector of size $mn$ and its preconditioner $\tilde{G}_t$ has  size $mn \times mn$, which leads to $O(m^2n^2)$ and  $O(m^3n^3)$ space and time complexities, respectively. In contrast, Shampoo constructs two smaller matrices,  $L_t \in  \mathbb{R}^{m \times m}$ and $R_t \in\mathbb{R}^{n \times n}$, to precondition the rows and columns of $G_t^X$, respectively, which only requires an $O(m^2+n^2)$ memory cost and an $O(m^3+n^3)$ computational complexity.  Since the parameters in deep learning tasks often have matrix structures, Shampoo has strong empirical performance and has received much attention \citep{anil2020scalable,liu2023sophia,eschenhagen2024kronecker}. 

In recent years, a growing body of work has sought to improve Shampoo from complementary perspectives. CASPR \citep{duvvuricombining} designs a new
preconditioning matrix to improve the regret bound of Shampoo. \cite{vyas2025soap} introduce SOAP, which performs Adam \citep{kingma2014adam} in the eigenspace of the gradient, achieving better performance than Adam and Shampoo. Another line of research focuses on reducing optimizer-state memory through low-rank gradient projection.
GaLore \citep{zhao2024galore} projects gradients onto a low-rank
subspace and maintains the preconditioner only within this reduced
space to reduce the memory cost. However, as highlighted by
\cite{he2025subspace}, GaLore provably fails to converge in the regime of large stochastic noise. To address this limitation, they propose GoLore, which replaces the SVD-based projection in GaLore with a random projection, thereby restoring convergence guarantees. From a different perspective, Muon \citep{jordan2024muon} formulates the update as steepest descent under the spectral norm and has been successfully applied to training large foundation models at scale. Building on low-rank gradient projection, \citet{pan2025unbiased} reduce the memory requirements of Muon by applying layerwise low-rank projections \citep{pan2024lisa}. 
\cite{nguyen2025improving} propose AdaDiag, which can be viewed as a simplified variant of SOAP that performs singular value decomposition without accumulating gradient covariance statistics. 

However, the memory demands of Shampoo may still be prohibitive for large-scale neural networks. \cite{anil2020scalable} address the memory cost of Shampoo by introducing two variants. The first variant partitions the variable $X_t$ into several blocks but it depends on the specific ordering of neurons in the hidden layers. The second variant relies on one-sided covariance upper bounds, which cannot effectively handle vector parameters. \cite{feinberg2024sketchy} first incorporate FD into Shampoo and reduce its memory to $O(\tau(m + n))$. Their method, named S-Shampoo, uses two low-rank matrices $\hat{L}_t \in \mathbb{R}^{m\times \tau}$ and $\hat{R}_t \in \mathbb{R}^{n\times \tau}$  to approximate the preconditioning matrices $L_t$ and $R_t$, respectively. While S-Shampoo improves the space complexity of Shampoo, its regret bound relies on the dimensions $m,n$,  leading to the unsatisfactory performance when the dimensions are high. More recently, two concurrent works \citep{xie25icml,an2025asgo} independently developed ASGO (also referred to as One-Sided Shampoo), which maintains a preconditioner for only one axis of the gradient matrix. Despite this simpler structure, ASGO achieves improved regret bounds compared to Shampoo at a lower computational cost.

\section{Preliminaries}\label{section3}

We begin by introducing the notation, followed by the assumptions 
underlying our analysis and the sketching technique that serves as 
the algorithmic foundation of our methods.

\textbf{Notation.} We use $\x \in \mathbb{R}^d$ to denote a vector and $X \in \mathbb{R}^{m\times n}$ to represent a matrix. Without loss of generality, we assume $m \leq n$. For simplicity, we use $\Norm{\cdot}$ for the $\ell_2$ norm by default. For any vector $\x \in \mathbb{R}^{d}$ and a positive semi-definite matrix $A \in \mathbb{R}^{d\times d}$, $\Norm{\x}_A^2 = \left\langle \x, A\x\right\rangle$. We use $[\x]_i$ to represent the $i$-th element of a vector $\x$. For a matrix $A$, $A^{-1}$ is the inverse of  $A$ if  $A$ is full rank; otherwise, $A^{-1}$ is taken to be the Moore-Penrose pseudoinverse. We denote by $\Norm{A}_2$ the spectral norm of a matrix $A$, by $\Norm{A}_{F}$ the Frobenius norm of $A$, and by $\Norm{A}_{*}$ the trace norm of $A$. For two symmetric matrices $A,B$, $A \preceq B$ if and only if $B-A$ is positive semi-definite. For two matrices of the same size, we denote their Frobenius inner product by $\langle A,B\rangle=\operatorname{tr}(A^\top B)$. We use $A\otimes B$ to denote Kronecker product \citep{petersen2008matrix} and $\overline{\operatorname{vec}}(A)$ to denote row-major order vectorization of matrix $A$. We define the space complexity and time complexity as the memory and time usage in each round, respectively.  $\lambda_{[1:\tau]}$ is a sequence containing $\tau$ elements, and $\max\{\lambda_{[1:\tau]}, a\}$ represents a new sequence, where each element is the maximum of $\lambda_i$ and $a$, and $\text{diag}(\lambda_{[1:\tau]})$ is a diagonal matrix where the $i$-th diagonal element is $\lambda_i$, $1\leq i \leq\tau$.

We adopt a standard assumption of online convex optimization \citep{hazan2016introduction}.
\begin{assumption}[Convexity]\label{convex}  
    \textnormal{All loss functions $f_t(\cdot)$ are convex.}
\end{assumption}




We then briefly introduce the key technique behind our methods.

\paragraph{Frequent Directions.} Frequent directions (FD)~\citep{ghashami2016frequent} is a deterministic matrix sketching technique that extends the well-known algorithm for approximating item frequencies in online data streams \citep{misra1982finding}.   For a given matrix $A \in \mathbb{R}^{d \times t}$, FD aims to generate a matrix $B \in \mathbb{R}^{d \times \tau}$ such that $BB^{\top} \approx AA^{\top}$, where $\tau \ll \min\{t,d\}$ is the sketching size. The procedure is summarized in  Algorithm \ref{alg:fd1}. In each round $t$, we denote the low-rank matrix $B_{t-1} = [\mathbf{b}_{1},\mathbf{b}_{2},...,\mathbf{b}_{\tau-1},\textbf{0}_d]\in\mathbb{R}^{d\times \tau}$, where the last column is $\textbf{0}_{d}$.  Upon receiving the new gradient $\g_t \in \mathbb{R}^d$, it is inserted into the last column. Next, we perform singular value decomposition (SVD) on $B_{t-1} = U_t \sqrt{\operatorname{diag}(\lambda^{(t)}_{[1:\tau]})} V^{\top}_t$, and the matrix $B_t$ is computed as $B_{t} = U_t \sqrt{\text{diag}(\lambda^{(t)}_{[1:\tau]} - \lambda_{\tau}^{(t)})}$ with its last column set to  $\textbf{0}_{d}$.  The time complexity of FD is $O(\tau^2 d)$ for each iteration, which is dominated by computing the SVD of $B_{t-1}$, causing a quadratic dependence on the sketching size $\tau$.

To further reduce the time complexity of FD, \cite{ghashami2016frequent} propose fast frequent directions (FFD) by expanding the space of $B_t$. Specifically, FFD maintains a matrix $B_{0} = \textbf{0}_{d\times 2\tau} \in \mathbb{R}^{d\times 2\tau}$. In each round $t$, we insert the received gradient $\g_t$ into the first all-zero column of $B_{t-1}$. Once $B_t$ no longer contains any all-zero columns, we perform SVD and obtain $B_t=U_t\sqrt{\operatorname{diag}(\lambda^{(t)}_{[1:2\tau]})} V^{\top}_t$. $B_t$ is then updated as $B_t = U_t\sqrt{\operatorname{diag}(\max\{\lambda^{(t)}_{[1:2\tau]} - \lambda_{\tau}^{(t)},0\})}$, ensuring that the last  $\tau+1$ columns are set to $\textbf{0}_d$. As we only need to update $B_t$ every $\tau+1$ rounds, the time complexity of FFD is $O(\tau d)$.

Since FD removes a singular value per round, the matrix $B_tB_t^{\top}$ does not preserve monotonicity. To resolve this limitation, \cite{ijcai2020p588} propose spectral compensation frequent directions (SCFD), which adds up the total mass of subtracted values $\sum_{i=1}^t\lambda_{\tau}^{(i)}$ during the FD procedure. SCFD is able to approximate a sequence of high-dimensional matrices while preserving positive definite monotonicity.

\section{Efficient Preconditioned Methods for Vector-Valued Decisions}\label{section4}

In this section, we first present  FTSL, which enjoys a better dimension-free regret bound. Furthermore, we accelerate FTSL by employing a doubling trick for FD, achieving  enhanced computational efficiency. We demonstrate that this technique can be applied to expedite S-ADA. Finally, we apply our methods to offline stochastic optimization.

\begin{algorithm}[t]
   \caption{Frequent Directions (FD)}
   \label{alg:fd1}
\begin{algorithmic}[1]
    \STATE {\bfseries Input:} Sketching matrix $B_{t-1} \in \mathbb{R}^{d \times \tau}$ (with its last column as $\textbf{0}_d$), new gradient vector $\g_t \in \mathbb{R}^{d}$
    \STATE Insert the gradient $\g_t$ into the last column of $B_{t-1}$
    \STATE Perform SVD to $B_{t-1} = U_t \sqrt{\text{diag}(\lambda^{(t)}_{[1:\tau]})}V^{\top}_{t}$, where $U_t \in \mathbb{R}^{d \times \tau}$
    \STATE Compute $B_{t} = U_t \sqrt{\text{diag}(\lambda^{(t)}_{[1:\tau]} - \lambda_{\tau}^{(t)})}$
    \STATE {\bfseries Return:} $B_t$ and $\lambda_{\tau}^{(t)}$
\end{algorithmic}
\end{algorithm}

\subsection{Our Improved Results}

Before introducing our algorithms, we first briefly discuss  why the regret bound of S-ADA \citep{feinberg2024sketchy} relies on the dimensionality $d$, offering motivation
for the methods we design. Since S-ADA is under the  mirror descent framework, its regret bound contains  the Bregman divergence  term,  that is,
\begin{align*}
    &O\left( \sum_{t=0}^{T-1} \left[ B_{\Psi_{t+1}} (\x^*, \x_{t+1}) - B_{\Psi_t} (\x^*, \x_{t+1}) \right]\right)=O\left(\sum_{t=0}^{T-1} \left\| \x_{t+1} - \x^* \right\|^2_{\widetilde{G}_{t+1}^{1/2} - \widetilde{G}_{t}^{1/2}}\right),
\end{align*}
where $\Psi_t(\x) = \frac{1}{2} \langle\x,\tilde{G}_t^{1/2} \x \rangle$ is the proximal term, 
$\tilde{G}_t$ is the preconditioning matrix and $B_{\Psi_t}(\mathbf{x}, \mathbf{y}) = \Psi_t(\mathbf{x}) - \Psi_t(\mathbf{y}) - \left\langle \nabla \Psi_t(\mathbf{y}), \mathbf{x} - \mathbf{y} \right\rangle$. 
To facilitate summation, they upper bound this term by $O(\sum_{t=0}^{T-1}\text{tr}(\tilde{G}_{t+1}^{1/2}-\tilde{G}_{t}^{1/2}))$ and then exploit the additivity of the trace, which yields a bound of $O(\text{tr}(\tilde{G}_T^{1/2}))$. 
\cite{feinberg2024sketchy} add  the cumulative removed eigenvalues of FD into $\tilde{G}_t$. Consequently, the  Bregman divergence term is $O(\operatorname{tr}((B_TB_T^{\top} + \rho_{1:T}I_{d\times d})^{1/2}))$, which is further bounded by $ O(\operatorname{tr}(G_T^{1/2}) + \sqrt{d(d-\tau)\rho_{1:T}})$, where $B_T \in \mathbb{R}^{d\times\tau}$ is the sketching matrix and  $G_T = \sum_{t=1}^T \g_t\g_t^{\top}$. As a result, the regret bound of S-ADA exhibits a \textit{linear} dependence on $d$, resulting in an unsatisfactory performance in high-dimensional problems.

To overcome this issue, we  integrate  FD with ADA-FULL under \textit{the primal-dual framework}. Our method, which we call FTSL, is outlined in Algorithm~\ref{alg:ftrl1}. Specifically, we employ  FD to construct a low-rank approximation of the outer product matrix of gradients,  aiming to reduce the computational complexity. To ensure the monotonicity of the preconditioning matrix $\tilde{G}_t$, we also add back the cumulative escaped masses $\rho_{1:t}$ into $\tilde{G}_t$. Under the primal-dual  framework, we update the decision as follows
\begin{equation}\label{update}
\begin{aligned}
    \x_{t+1} & = \arg\min_{\x \in \mathbb{R}^d} \left\{\eta \left\langle \overline{\g}_t,\x  \right\rangle + \Psi_t(\x) \right\}\\
    &=  -\eta \tilde{G}_t^{-1/2}\overline{\g}_t,
\end{aligned}
\end{equation}
where $\overline{\g}_t = \sum_{i=1}^t \g_i$ is the sum of the past gradients. According to the analysis under the primal-dual framework, the regret of FTSL is upper bounded by the term $O(\|\tilde{G}_T^{1/2}\|_2) = O(\|(B_TB_T^{\top}+ \rho_{1:T}I_{d\times d})^{1/2}\|_2) \leq O(\|G_T^{1/2}\|_2 + \sqrt{\rho_{1:T}})$, thereby avoiding the dependence on $d$.

Formally, we present the theoretical guarantee of FTSL.

\begin{algorithm}[t]
   \caption{Follow-the-Sketchy-Leader (FTSL)}
   \label{alg:ftrl1}
\begin{algorithmic}[1]
    \STATE {\bfseries Input:} Learning rate $\eta$, sketching size $\tau$
    \STATE Initialize $\x_1 = \textbf{0}_d, \overline{\g}_0=\textbf{0}_d,  \tilde{G}_0 = \textbf{0}_{d\times d}, B_0 = \textbf{0}_{d\times \tau}$
     \FOR{$t=1$ to $T$}
     \STATE Play the decision $\x_t$ and suffer the loss $f_t(\x_t)$
     \STATE Query the gradient $\g_t = \nabla f_t(\x_t)$ and calculate $\overline{\g}_t = \overline{\g}_{t-1} + \g_t$
     \STATE Send $B_{t-1}$ and $\g_t$ to Algorithm \ref{alg:fd1}
     \STATE Receive $B_t$ and set $\rho_t = \lambda_{\tau}^{(t)}$
     \STATE Calculate $\tilde{G}_t = B_tB_t^{\top} + \rho_{1:t}I_{d\times d}$ and derive $\tilde{G}_t^{-1/2}$
     \STATE Update $\x_{t+1}$ according to (\ref{update})
     \ENDFOR
\end{algorithmic}
\end{algorithm}

\begin{theorem}\label{thm1}
    Under Assumption \ref{convex}, by setting the learning rate $\eta = \frac{D}{\sqrt{2}}$, where $D \geq \Norm{\x^{*}}$ is the upper bound on the norm of the optimal decision $\x^{*} \in \arg\min_{\x \in \R^d} \sum_{t=1}^T f_t(\x)$, FTSL ensures
    \begin{equation*}
        R(T) \leq O\left(D\left(\operatorname{tr}(G_T^{1/2}) + \sqrt{\rho_{1:T}}\right)\right),
    \end{equation*}
    where $G_T = \sum_{t=1}^T \g_t\g_t^{\top}$.
\end{theorem}
 
\begin{rem}\textnormal{
 In contrast to the previous regret bound of $O(\operatorname{tr}(G_T^{1/2}) + \sqrt{d(d-\tau)\rho_{1:T}})$ \citep{feinberg2024sketchy}, the regret bound of FTSL is \textit{dimension-free}, a benefit realized from the primal-dual framework. }
\end{rem}

\begin{rem}\textnormal{
Since we only maintain a sketching matrix $B_t \in \mathbb{R}^{d\times \tau}$, the space complexity of FTSL is  $O(\tau d)$. Its time complexity is  $O(\tau^2d)$ per round, which arises from the SVD of $B_t$ and the calculation of the matrix $\tilde{G}_t^{-1/2}$. While the time complexity of FTSL is linear  with respect to the dimensionality $d$, it still suffers from the quadratic dependence on the sketching size $\tau$. To further alleviate its computational burden, we develop a fast variant of FTSL in the next section.}
\end{rem}
    
\begin{algorithm}[t]
   \caption{Follow-the-Fast-Sketchy-Leader (FTFSL)}
   \label{alg:ffd}
\begin{algorithmic}[1]
    \STATE {\bfseries Input:} Learning rate $\eta$, sketching size $\tau$
    \STATE Initialize $\x_1 = \mathbf{0}_d , \tilde{G}_0 = \textbf{0}_{d\times d}, r_0 = 0, M_0 = \mathbf{0}_{2\tau \times 2\tau},  V_0 = \mathbf{0}_{d \times 2\tau},\overline{\g}_0=\mathbf{0}_d, \rho_1=0$
     \FOR{ $t=1$ to $T$}
     \STATE Play the decision $\x_t$ and suffer the loss $f_t(\x_t)$
     \STATE Query the gradient $\g_t = \nabla f_t(\x_t)$ and compute $\g^{\prime}_t = V_{t-1}(V_{t-1}^{\top}\g_t),\overline{\g}_t = \overline{\g}_{t-1} + \g_t$
     \IF{$\g^{\prime}_t \neq \g_t $}\label{step:6}
     \STATE Set $r_{t-1} = r_{t-1}+1$, calculate $\mathbf{v}_{r_{t-1}}^{t-1} = \frac{\mathbf{g}_t - \mathbf{g}^{\prime}_t}{\left\| \mathbf{g}_t - \mathbf{g}^{\prime}_t \right\|}$ and set the $r_{t-1}$-th column of $V_{t-1}$ as  $\mathbf{v}_{r_{t-1}}^{t-1}$
     \ENDIF\label{step:8}
     \STATE Set $r_t=r_{t-1}, V_t = V_{t-1}$
     \STATE Compute $M_t  =M_{t-1} + (V_{t-1}^{\top}\g_t)(V_{t-1}^{\top}\g_t)^{\top}$
     \STATE Perform SVD decomposition on $M_t$, which is  $ U_t\Sigma_t U_t^{\top} = U_t\text{diag}(\lambda^{(t)}_{[1:2\tau]}) U_t^{\top} = M_t$
     \STATE Calculate $\tilde{G}_t = \rho_{1:t}I_{d\times d} + V_{t}U_t\Sigma_t U_t^{\top}V_{t}^{\top}$
     \STATE Update $\x_{t+1}$ according to (\ref{update}) and set $\rho_{t+1} = 0$\label{step:13}
     \IF{$r_t = 2\tau$}\label{step:14}
     \STATE Set $\rho_{t+1} =\lambda_{\tau}^{(t)}$, $M_t = \text{diag}(\max\{\lambda^{(t)}_{[1:2\tau]}-\lambda_{\tau}^{(t)},0\})$ and $V_t = V_{t} U_t$
     \STATE Set $r_t = \tau -1$ and set the $\tau$-th to $2\tau$-th columns of $V_t$ to $\textbf{0}_d$
     \ENDIF\label{step:16}
     \ENDFOR
\end{algorithmic}
\end{algorithm}

\subsection{Our Accelerated Variant}
The time complexity of FTSL exhibits a quadratic dependence on the sketching size $\tau$, which arises from the SVD of $B_t \in \mathbb{R}^{d\times\tau}$ performed in every round of FD. To reduce this cost, drawing inspiration from previous work \citep{ijcai2020p588,wan2021efficient}, we adopt a more efficient strategy for computing the SVD of $B_t$. The resulting method, termed FTFSL, is summarized in Algorithm \ref{alg:ffd}.

In the matrix-approximation literature, \citet{ghashami2016frequent} develop fast FD to reduce the number of SVD computations in FD from one per round to one every $\tau+1$ rounds, while maintaining the same order of space complexity. FFD expands the sketching matrix $B_t$ from $\mathbb{R}^{d\times\tau}$ to $\mathbb{R}^{d\times2\tau}$. FTFSL adopts this approach but, instead of maintaining $B_t$ explicitly, represents it through two matrices $V_t$ and $M_t$. Specifically, $V_{t} =[\mathbf{v}_{1}^{t},\cdots,\mathbf{v}_{2\tau}^{t}] \in \mathbb{R}^{d \times 2\tau}$ consists of $r_{t}$ orthonormal vectors ($r_t \leq 2\tau$) and the remaining columns are zero vectors, and $M_t \in \mathbb{R}^{2\tau \times 2\tau}$ is a symmetric matrix. We require that $V_t$ and $M_t$ satisfy the condition $V_tM_t^{1/2} = B_t \in \mathbb{R}^{d \times 2\tau}$.  In each round $t$, after receiving the gradient $\g_t$, we first check whether this vector lies within the subspace spanned by $V_{t-1}$. If the vector is not contained within the subspace, we normalize it and subsequently add it to $V_{t-1}$,  thereby enlarging the span of the subspace and ensuring $V_{t-1}V_{t-1}^{\top}\g_t=\g_{t}$  (Steps \ref{step:6}-\ref{step:8}). Then we have the following  identity:
\begin{align*}
&V_{t-1} M_{t-1} V_{t-1}^\top + \mathbf{g}_t \mathbf{g}_t^\top= V_{t-1} \left( M_{t-1} + V_{t-1}^\top \mathbf{g}_t \mathbf{g}_t^\top V_{t-1} \right) V_{t-1}^\top.
\end{align*}
This implies that we only need to perform an SVD decomposition on the matrix $M_{t-1} + (V_{t-1}^{\top}\g_t)(V_{t-1}^{\top}\g_t)^{\top} \in \mathbb{R}^{2\tau\times 2\tau}$, which only takes a time complexity of $O(\tau^3)$. Next,  we incorporate the escaped masses to keep the monotonicity of the preconditioning matrix $\tilde{G}_t$.  When $r_t = 2\tau$, we need to set $r_t=\tau-1$ and set  the last $\tau+1$ columns of the sketching matrix $B_t$ to be zero (Steps \ref{step:14}-\ref{step:16}). Given the decomposition $M_t = U_t \operatorname{diag}(\lambda^{(t)}_{[1:2\tau]})U_t^{\top}$ and the relationship $V_tM_t^{1/2} = B_t$, this can be efficiently achieved by updating $M_t$ as $\text{diag}(\max\{\lambda^{(t)}_{[1:2\tau]}-\lambda_{\tau}^{(t)},0\})$, calculating $V_t = V_{t}U_t$, and setting the last $\tau+1$ columns of $V_t$ to zero.

 In the following,  we provide the theoretical guarantee of FTFSL.

\begin{theorem}\label{thm:ffd1}
    Under Assumption \ref{convex}, by setting the learning rate $\eta = \frac{D}{\sqrt{2}}$, where $D \geq \Norm{\x^{*}}$ is the upper bound on the norm of the optimal decision $\x^{*} \in \arg\min_{\x \in \R^d} \sum_{t=1}^T f_t(\x)$, FTFSL ensures
    \begin{equation*}
        R(T) \leq O\left(D\left(\operatorname{tr}(G_T^{1/2}) + \sqrt{\rho_{1:T}}\right)\right).
    \end{equation*}
\end{theorem}

\begin{rem}\textnormal{
    In each round, FTFSL computes $\g^{\prime}_t$, $\mathbf{v}_{r_{t-1}}^{t-1}$, $M_t$, the SVD of $M_t$ and updates $\x_t$, with respective time complexities of  $O(\tau d)$, $O(d)$,  $O(\tau d)$,  $O(\tau^3)$ and  $O(\tau d)$. Additionally, we only compute $V_t = V_t U_t$ every $\tau +1$ rounds,  incurring a time complexity of $O(\tau^2 d)$. When $\tau \leq O(\sqrt{d})$, the time complexity of FTFSL is $O(\tau d)$ per round.}
\end{rem}

Notably, we can also reduce the time complexity of S-ADA \citep{feinberg2024sketchy} by adopting this technique. We replace the update rule for the decision variable $\x_t$ of FTFSL (Step \ref{step:13}) with the following
\begin{align*}
    \x_{t+1} = \x_t - \eta \tilde{G}^{-1/2}_t\g_t,
\end{align*}
and analyze it under the mirror descent framework.

Then we provide the theoretical guarantee of Fast  S-ADA.
\begin{theorem}\label{thm:4}
    Under Assumption \ref{convex}, by setting the learning rate $\eta = \frac{D_1}{\sqrt{2}}$, where $D_1 = \max_{t \in [T]}\Norm{\x_t - \x^{*}}$, Fast S-ADA ensures
    \begin{align*}
        R(T) \leq  O\left(D_1\left(\operatorname{tr}(G_T^{1/2}) + \sqrt{d(d-\tau)\rho_{1:T}}\right)\right).
\end{align*}
\end{theorem}

\begin{rem}\textnormal{
    Compared to S-ADA \citep{feinberg2024sketchy}, Fast S-ADA obtains a better $O(\tau d)$ time complexity when $\tau \leq O(\sqrt{d})$, while preserving the same regret bound.}
\end{rem}
\begin{algorithm}[t]
   \caption{Online-to-Batch Conversion}
   \label{alg:o2b-sco}
\begin{algorithmic}[1]
    \STATE {\bfseries Input:} online algorithm $\A$, batch size $M$
    \FOR{$t=1$ to $T$}
     \STATE Receive the decision $\x_t$ from $\A$
     \STATE Compute the stochastic gradient $\g_t = \frac{1}{M}\sum_{i=1}^M \nabla f(\x_t;\xi_{t,i})$
     \STATE Send $\g_t$ to $\A$ as the gradient of round $t$
    \ENDFOR
    \STATE {\bfseries Return:} $\overline{\x}_T = \frac{1}{T}\sum_{t=1}^T \x_t$
\end{algorithmic}
\end{algorithm}

\subsection{Application to Offline Optimization}\label{offline-vector}

In this section, we extend our methods to the offline stochastic convex
optimization setting. Formally, we consider the problem
\begin{equation}\label{offline}
    \min_{\x \in \R^{d}} f(\x) = \E\left[f(\x;\xi)\right],
\end{equation}
where $f(\cdot)$ is convex and only noisy gradient estimates $\nabla f(\x;\xi)$ are
accessible. The random sample $\xi$ is drawn from a stochastic oracle
satisfying $\E\left[\nabla f(\x;\xi)\right] = \nabla f(\x)$.

We adopt the standard online-to-batch conversion \citep{CesaBianchi2004Generalization}, as summarized in Algorithm~\ref{alg:o2b-sco}. At each round~$t$, the
learner receives the decision $\x_t$ produced by the online algorithm $\A$,
samples a mini-batch of size $M$, and computes the stochastic gradient
$\g_t = \frac{1}{M} \sum_{i=1}^{M} \nabla f(\x_t;\xi_{t,i})$, which is
fed back to $\A$ as the gradient of round~$t$.

\paragraph{Non-smooth theory.} We first establish the convergence rates for non-smooth loss functions.
\begin{theorem}\label{thm:p1}
    Under Assumption \ref{convex}, by setting the learning rate $\eta = \frac{D}{\sqrt{2}}$ and $M=1$, where $D \geq \Norm{\x^{*}}$ is the upper bound on the norm of the optimal decision $\x^{*} \in \arg\min_{\x \in \R^d} f(\x)$, FTFSL ensures
    \begin{equation*}
        \frac{1}{T}\sum_{t=1}^T \E\left[f(\x_t)- f(\x^*)\right] \leq O\left(\frac{D\operatorname{tr}(G_T^{1/2})}{T} + \frac{D\sqrt{\rho_{1:T}}}{T}\right).
    \end{equation*}

Under Assumption \ref{convex}, by setting the learning rate $\eta = \frac{D_1}{\sqrt{2}}$ and $M=1$, where $D_1 = \max_{t \in [T]}\Norm{\x_t - \x^{*}}$, Fast S-ADA ensures
\begin{equation}
        \frac{1}{T}\sum_{t=1}^T \E\left[f(\x_t)- f(\x^*)\right] \leq O\left(\frac{D_1\operatorname{tr}(G_T^{1/2})}{T} + \frac{D_1\sqrt{d(d-\tau)\rho_{1:T}}}{T}\right).
\end{equation}
\end{theorem}

\begin{rem}\textnormal{
Under the standard bounded-gradient assumption $\Norm{\g_t} \leq G$ for all $t \in [T]$, both FTFSL and Fast S-ADA obtain the optimal $O(1/\sqrt{T})$ convergence rate  for convex non-smooth loss functions \citep{Agarwal2012OracleLowerBound}.} 
\end{rem}

\paragraph{Smooth theory.}   Many objectives in modern machine learning are smooth, in the sense that their gradients are Lipschitz continuous, as exemplified by logistic regression and networks with smooth activations.  Previous work has shown that smoothness can be exploited to obtain faster convergence rates \citep{NIPS2011_40008b9a,pmlr-v97-cutkosky19a,Zhang2020bGradientClipping,pmlr-v195-faw23a,pmlr-v195-wang23a,liu2024adagradanisotropicsmoothness,yu2025mirror}. Despite this progress, \textit{the analysis of memory-efficient full-matrix adaptive methods on smooth objectives remains absent}. To fill this gap, we extend our analysis to the smooth setting. Following prior work on adaptive optimization methods \citep{liu2024adagradanisotropicsmoothness}, we impose two standard assumptions.

\begin{assumption}[Anisotropic smoothness]\label{smooth1}
There exists a positive diagonal matrix 
$\L = \operatorname{diag}(L_1,\ldots,L_d)$ such that
\begin{equation*}
    \Norm{\nabla f(\x) - \nabla f(\y)}_{\L^{-1}} 
    \leq \Norm{\x - \y}_{\L}, \qquad \forall \x, \y \in \R^d,
\end{equation*}
when $f(\cdot)$ is twice differentiable, it is equivalent to 
$\nabla^2 f(\x) \preceq \L$, $ \forall \x \in \R^d$.
\end{assumption}

\begin{rem}\textnormal{
Assumption \ref{smooth1} refines the standard smoothness condition by 
capturing the coordinate-wise curvature of $f(\cdot)$. Specifically, 
it reduces to the classical $L$-smoothness with $L = \Norm{\L}_\infty$ \citep{nesterov2018lectures}, while allowing a much tighter 
characterization when the Hessian spectrum is imbalanced across 
coordinates. In such regimes, the diagonal bound $\Norm{\mathbf{L}}_1$ can be substantially 
smaller than the uniform bound $Ld$ \citep{sagun2016eigenvalues,pan2021eigencurve}.}
\end{rem}

\begin{assumption}[Bounded noise variance]\label{noise1}
Let $\n_t = \nabla f(\x_t;\xi_t) - \nabla f(\x_t)$ denote the stochastic 
gradient noise at round $t$. There exists a  vector 
$\v = [\v_1,\ldots,\v_d] \in \R^d$ such that 
$\E[\n_t] = \mathbf{0}$ and
\begin{equation*}
    \E\left[[\n_{t}]_i^2\right] \leq [\v]_i^2, \qquad \forall i \in [d].
\end{equation*}
\end{assumption}
\begin{rem}
    \textnormal{Assumption \ref{noise1} implies the standard bounded noise assumption by $\E\left[\Norm{\n_t}^2\right] \leq \Norm{\v}^2$. Intuitively, it bounds each  coordinate of $\n_t$ rather than only its norm, thereby providing finer-grained information about the noise scale.}
\end{rem}

In the following, we present our analysis for smooth loss functions.
\begin{theorem}\label{thm:p2}
    Under Assumptions \ref{convex}, \ref{smooth1} and \ref{noise1}, by setting the learning rate $\eta = \frac{D}{\sqrt{2}}$, where  $D \geq \Norm{\x^{*}}$ is the bound on the optimal decision  $\x^{*} \in \arg\min_{\x \in \R^d}f(\x)$, FTFSL ensures
    \begin{equation*}
        \frac{1}{T}\sum_{t=1}^T \E\left[f(\x_t)- f(\x^*)\right] \leq O\left(\frac{D^2\tr(\L)}{T} + \frac{D\Norm{\v}}{\sqrt{MT}} + \frac{D\sqrt{\rho_{1:T}}}{T}\right).
    \end{equation*}
Under Assumptions \ref{convex}, \ref{smooth1} and \ref{noise1}, by setting the learning rate $\eta = \frac{D_1}{\sqrt{2}}$, where $D_1 = \max_{t \in [T]}\Norm{\x_t - \x^{*}}$, Fast S-ADA ensures
    \begin{equation*}\frac{1}{T}\sum_{t=1}^T \E\left[f(\x_t)- f(\x^*)\right] \leq O\left(\frac{D^2_1\tr(\L)}{T} + \frac{D_1\Norm{\v}}{\sqrt{MT}} + \frac{D_1\sqrt{d(d-\tau)}\sqrt{\rho_{1:T}}}{T}\right).
\end{equation*}
\end{theorem}

\begin{rem}\textnormal{
Both bounds can be decomposed into three terms: a deterministic term of  order $O(\operatorname{tr}(\L)/T)$, a stochastic noise term of order 
$O(\|\mathbf{v}\|/\sqrt{MT})$, and a sketching error term involving 
$O(\sqrt{\rho_{1:T}}/T)$. The first two terms match the convergence rate of 
their non-sketched counterparts, while the third term quantifies the 
additional error introduced by FD.  Notably, FTFSL achieves a 
dimension-free sketching error term, whereas that of Fast S-ADA 
scales linearly with $d$.}
\end{rem}

\section{Efficient Preconditioned Methods for Matrix-Valued Decisions}\label{section5}

In this section, we consider a practical scenario where the decision variable is a matrix $X_t \in \mathbb{R}^{m\times n}$, which is common for parameters in deep learning tasks \citep{lecun2015deep}.

\subsection{Our Proposed Methods}
A direct application of the optimization methods developed in the 
vector setting requires vectorizing the gradient $G^X_t$ as $\g_t \in \mathbb{R}^{mn}$. However, such a reduction incurs prohibitive memory costs. In particular, ADA-FULL maintains an $mn \times mn$ preconditioner with $O((mn)^2)$ memory, while even the FD-sketched variants developed in Section~\ref{section4} reduce  this to $O(\tau mn)$, which remains impractical for large $m$  and $n$. Additionally,  if we consider the matrix structure of gradients in deep neural networks, it has been widely observed that these structured gradients are usually low-rank \citep{zhao2021zero,yang2023spectral,cosson2023lowrank}.

To exploit the structure information, \cite{gupta2018shampoo} propose  Shampoo, which retains the matrix structure of the gradient and  maintains two matrices as preconditioners of the rows and columns of $G_t^X$, yielding a space complexity of $O(m^2 + n^2)$.  While S-Shampoo \citep{feinberg2024sketchy} improves the space complexity of Shampoo to $O(\tau(m+n))$, its regret bound again relies on the dimensionality $m,n$. To further reduce its regret bound,  we propose FTSL-Shampoo by integrating FD with Shampoo under the primal-dual framework. Our method  achieves a superior dimension-free guarantee while obtaining an $O(\tau(m+n))$ space complexity, which is presented in Algorithm \ref{shampoo}.

\begin{algorithm}[t]
   \caption{Frequent Directions in General Form}
   \label{alg:fd2}
\begin{algorithmic}[1]
    \STATE {\bfseries Input:} Sketching matrix $B_{t-1} \in \mathbb{R}^{d \times \tau}$, a new symmetric PSD matrix $M_t \in \mathbb{R}^{d\times d}$
    \STATE Eigendecompose $\overline{U}_t \text{diag}(\lambda^{(t)})\overline{U}_t^\top = B_{t-1}B_{t-1}^{\top} + M_t$, let $U_t \in \mathbb{R}^{d \times \tau}$ be the first $\tau$ columns of $\overline{U}_t$ and $\lambda^{(t)}_{[1:\tau]}$ be the corresponding eigenvalues
    \STATE Compute $B_{t} = U_t \sqrt{\text{diag}(\lambda^{(t)}_{[1:\tau]}-\lambda_{\tau}^{(t)})}$
    \STATE {\bfseries Return:} $B_t$ and $\lambda_{\tau}^{(t)}$
\end{algorithmic}
\end{algorithm}

\begin{algorithm}[t]
   \caption{FTSL-Shampoo}
	\label{shampoo}
	\begin{algorithmic}[1]
     \REQUIRE Learning rate $\eta$, sketching size $\tau$, $\epsilon >0$
     \STATE Initialize $X_1 = \textbf{0}_{m \times n}, \hat{L}_0 = \textbf{0}_{m\times \tau},   \hat{R}_0 = \textbf{0}_{n\times \tau}, \tilde{L}_0 = \textbf{0}_{m\times m},   \tilde{R}_0 = \textbf{0}_{n\times n}, \overline{G}_0^{X} = \textbf{0}_{m\times n}$
     \FOR{ $t=1$ to $T$}
     \STATE Play the decision $X_t$ and suffer the loss $f_t(X_t)$
     \STATE Query the gradient $G_t^X = \nabla f_t(X_t) \in \mathbb{R}^{m \times n}$ and calculate $\overline{G}_t^{X} = \overline{G}_{t-1}^{X} + G_t^X$
     \STATE Send $\hat{L}_{t-1}$ and $G_t^X(G_t^X)^{\top}$ to Algorithm \ref{alg:fd2} and receive $\hat{L}_t, \rho^L_{t}$
    \STATE Send $\hat{R}_{t-1}$ and $(G_t^X)^{\top}G_t^X$ to Algorithm \ref{alg:fd2} and receive $ \hat{R}_t,\rho^R_{t}$
     \STATE Update $\tilde{L}_t = \hat{L}_t \hat{L}_t^{\top} +(\epsilon + \rho_{1:t}^L)I_{m\times m}$
     \STATE Update $\tilde{R}_t = \hat{R}_t \hat{R}_t^{\top}+ (\epsilon +\rho_{1:t}^R)I_{n\times n}$
     \STATE Update $X_{t+1}$ according to (\ref{shampoo:update})
     \ENDFOR
	\end{algorithmic}
\end{algorithm}

Specifically, we utilize FD to approximate the left and right preconditioning matrices for Shampoo. We maintain two matrices $\hat{L}_t \in \mathbb{R}^{m \times \tau}, \hat{R}_t \in \mathbb{R}^{n \times \tau}$ to ensure that $\hat{L}_t\hat{L}_t^{\top} \approx \sum_{i=1}^t G_i^X(G_i^X)^{\top}, \hat{R}_t\hat{R}_t^{\top} \approx \sum_{i=1}^t (G_i^X)^{\top}G_i^X$. We track the cumulative escaped masses $\rho^L_{1:t}$ and $\rho^R_{1:t}$ of the left and right preconditioning matrices separately, and then add them back into $\tilde{L}_t$ and $\tilde{R}_t$ to uphold the monotonicity. To achieve a dimension-free regret bound with FD, we update the parameters as follows:
\begin{equation}\label{shampoo:update}
    X_{t+1} =  -\eta  \tilde{L}_t^{-1/4}\overline{G}^X_t\tilde{R}_t^{-1/4},
\end{equation}
where $\overline{G}^X_t = \sum_{i=1}^t G^X_i$ is the sum of the past gradients, and we conduct the analysis under \textit{the primal-dual framework}. Before we present the regret bound for FTSL-Shampoo,  we introduce a standard assumption that is specific to Shampoo-based methods, which has also been used in prior works \citep{gupta2018shampoo, feinberg2024sketchy}.

\begin{assumption} [Bounded rank]\label{ass:2}
    The rank of gradient matrix $G_t^X$ is bounded by $r$, i.e., $\max_{t \in [T]}\operatorname{rank}(G_t^X) \leq r$.
\end{assumption}
Then we establish the guarantee for FTSL-Shampoo.

\begin{theorem}\label{thm3}
    Under Assumptions \ref{convex} and \ref{ass:2}, by setting the learning rate $\eta = \frac{D_{F}}{\sqrt{r}}$, where $D_F = \Norm{X^*}_F$ is the Frobenius norm of the optimal decision $X^*$, FTSL-Shampoo ensures
    \begin{align*}
        R(T) &\leq O\left(D_F\sqrt{r}(\operatorname{tr}(L_T^{1/4}) + (\rho^L_{1:T})^{1/4})(\operatorname{tr}(R_T^{1/4}) + (\rho^R_{1:T})^{1/4})\right),
\end{align*}
where  $L_T = \sum_{t=1}^T G_t^X(G_t^X)^{\top}, R_T =  \sum_{t=1}^T (G_t^X)^{\top}G_t^X$, $\rho_{1:T}^L$ and $\rho_{1:T}^R$ represent the sum of the removed eigenvalues of FD during the approximation of ~ $\sum_{t=1}^T G_t^X(G_t^X)^{\top}$ and $ \sum_{t=1}^T (G_t^X)^{\top}G_t^X$, respectively.
\end{theorem}

\begin{rem}
    In comparison to the previous  $O(\sqrt{r}(\operatorname{tr}(L_T^{1/4}) + {m}(\rho_{1:T}^L)^{1/4}) (\operatorname{tr}(R_T^{1/4}) + {n}(\rho^R_{1:T})^{1/4}))$ regret bound of S-Shampoo  \citep{feinberg2024sketchy}, we achieve a dimension-free regret bound while enjoying the same $O(\tau(m+n))$ space complexity.
\end{rem}

\begin{algorithm}[t]
   \caption{S-ASGO}
	\label{alg:sasgo}
	\begin{algorithmic}[1]
     \REQUIRE Learning rate $\eta$, sketching size $\tau$, $\epsilon >0$
     \STATE Initialize $X_1 = \textbf{0}_{m \times n}, \hat{L}_0 = \textbf{0}_{m\times \tau}, \tilde{L}_0 = \textbf{0}_{m\times m}, \overline{G}_0^{X} = \textbf{0}_{m\times n}$
     \FOR{ $t=1$ to $T$}
     \STATE Play the decision $X_t$ and suffer the loss $f_t(X_t)$
     \STATE Query the gradient $G_t^X = \nabla f_t(X_t)$ and calculate $\overline{G}_t^{X} = \overline{G}_{t-1}^{X} + G_t^X$
     \STATE Send $\hat{L}_{t-1}$ and $G_t^X(G_t^X)^{\top}$ to Algorithm \ref{alg:fd2} and receive $\hat{L}_t, \rho^L_{t}$
     \STATE Denote $\tilde{L}_t =  \hat{L}_t \hat{L}_t^{\top} +(\epsilon + \rho_{1:t}^L)I_{m\times m}$
     \STATE Update $X_{t+1}$ according to \eqref{asgo:m} or \eqref{asgo:p}
     \ENDFOR
	\end{algorithmic}
\end{algorithm}

Although FTSL-Shampoo achieves an $O(\tau(m+n))$ space complexity, it requires maintaining a separate preconditioner for each axis of the gradient matrix. For convolutional neural networks, where a single layer may involve parameters with multiple dimensions, this necessitates storing and updating multiple preconditioners per layer. Beyond this overhead, the benefit obtained from such axis-wise preconditioning is itself limited. When the dimension is large, the theoretical guarantee of Shampoo can be worse than those of ADA-FULL or even SGD.  Recently, ASGO \citep{xie25icml,an2025asgo} demonstrates that maintaining a preconditioner along only a single axis suffices to achieve better theoretical guarantees.

Therefore, to address the above limitations, we integrate FD with ASGO to propose Sketchy ASGO (S-ASGO) under both the mirror descent and primal-dual frameworks, which is summarized in Algorithm \ref{alg:sasgo}. Without loss of generality, we assume $m \leq n$. S-ASGO maintains only a single sketching matrix
$\hat{L}_t \in \mathbb{R}^{m \times \tau}$ to approximate the matrix $\sum_{i=1}^t G_i^X(G_i^X)^{\top}$ along the smaller axis, reducing the space complexity from $O(\tau(m+n))$ to $O(\tau m)$.  In each round $t$, the learner first plays the decision $X_t$ and receives the gradient
$G_t^X = \nabla f_t(X_t)$. Then  we send $G_t^X$ and the previous sketching matrix $\hat{L}_{t-1}$ to
Algorithm~\ref{alg:fd2} to obtain the updated sketching matrix
$\hat{L}_t$ and the discarded eigenvalue $\rho_t^L$. Similar to FTSL-Shampoo, we add back the cumulative escaped mass
$\rho_{1:t}^L$ to preserve the positive definite monotonicity, and
construct the preconditioner as
\begin{equation*}
    \tilde{L}_t = \hat{L}_t \hat{L}_t^{\top} + (\epsilon +
    \rho_{1:t}^L) I_{m \times m}.
\end{equation*}
Under the primal-dual framework, the learner updates the decision as
\begin{equation}\label{asgo:p}
    X_{t+1} =- \eta \tilde{L}_t^{-1/2} \overline{G}_t^X.
\end{equation}
We now present the regret bound of S-ASGO under the primal-dual framework.

\begin{theorem}\label{thm:jmlr1}
Under Assumption \ref{convex}, by setting the learning rate $\eta = \frac{D_F}{\sqrt{2}}$, where $D_F= \Norm{X^*}_F$ is the Frobenius norm of the optimal decision $X^*$, S-ASGO under the primal-dual framework ensures
    \begin{align*}
        R(T)\leq O\left(D_F \left(\operatorname{tr}(L_T^{1/2}) +  \sqrt{\rho_{1:T}^L} \right)\right).
    \end{align*}
\end{theorem}

We next study S-ASGO under the mirror descent framework, where the learner updates the decision as
\begin{equation}\label{asgo:m}
    X_{t+1} = X_t - \eta \tilde{L}_t^{-1/2}G_t^X.
\end{equation}
We present its theoretical guarantee in the following theorem.

\begin{theorem}\label{thm:jmlr1md}
Under Assumption \ref{convex}, by setting the learning rate $\eta = \frac{D_2}{\sqrt{2}}$, where $D_2= \max_{t\in[T]}\Norm{X_t - X^*}_2$, S-ASGO under the mirror descent framework ensures
    \begin{align*}
        R(T)\leq O\left(D_2 \left(\operatorname{tr}(L_T^{1/2}) + m \sqrt{\rho_{1:T}^L} \right)\right).
    \end{align*}
\end{theorem}

\begin{rem}\textnormal{Compared with FTSL-Shampoo, S-ASGO not only reduces
the space complexity from $O(\tau(m+n))$ to $O(\tau m)$, but also
achieves a tighter regret bound. Specifically, for any positive
semi-definite matrix $A$, it holds that
$\sqrt{\operatorname{tr}(A)} \leq \operatorname{tr}(A^{1/2})$.
Applying this inequality yields
\begin{align*}
    \operatorname{tr}(L_T^{1/4}) \operatorname{tr}(R_T^{1/4})
    &\geq \operatorname{tr}(L_T^{1/4}) \operatorname{tr}(R_T)^{1/4} = \operatorname{tr}(L_T^{1/4}) \operatorname{tr}(L_T)^{1/4} \\
    &\geq \operatorname{tr}(L_T^{1/4}) \|L_T\|_{\mathrm{2}}^{1/4}
    \geq \operatorname{tr}(L_T^{1/2}).
\end{align*}
Therefore, the leading term
$O(\operatorname{tr}(L_T^{1/2}))$ of S-ASGO is no larger than
$O(\operatorname{tr}(L_T^{1/4}) \operatorname{tr}(R_T^{1/4}))$ of
FTSL-Shampoo.  Moreover, the regret bound of S-ASGO does not depend on the rank of the gradient.}
\end{rem}

\begin{rem} \textnormal{The two versions of S-ASGO offer a trade-off
between the regret bound and the problem-dependent constant. The
mirror descent variant achieves a regret bound with the smaller
constant $D_2$, but incurs a
dimension-dependent term $O(m\sqrt{\rho_{1:T}^L})$ arising from the
Bregman divergence analysis. In contrast, the primal-dual variant
eliminates this dimension dependence and achieves a fully
\textit{dimension-free} regret bound, at the cost of a larger
constant $D_F$ that bounds the norm of the optimal
decision.}
\end{rem}

\subsection{Application to Offline Optimization}\label{offline-matrix}
In parallel with Section~\ref{offline-vector}, we extend S-ASGO to the offline stochastic setting. Formally, we consider the following stochastic optimization problem
\begin{equation}\label{offline2}
    \min_{X \in \R^{m\times n}} f(X) = \E\left[f(X;\xi)\right],
\end{equation}
where $f(\cdot)$ is a convex loss function.  We assume that only noisy estimates of its gradient $\nabla f(X;\xi)$ can be accessed, where $\xi$
 is the random sample drawn from a stochastic oracle such that $\E\left[\nabla f(X;\xi)\right] = \nabla f(X)$.

In each round $t$, the learner receives the decision $X_t$ from the online algorithm $\A$. It samples a mini-batch of $M$ samples and queries the gradient $\hat{G}_t^X= \frac{1}{M} \sum_{i=1}^M \nabla f(X_t;\xi_{t,i})$.  Then it sends the stochastic gradient to $\A$ as the gradient in round $t$.

We first establish the convergence rate for convex non-smooth loss functions.

\begin{theorem}\label{thm:p4}
Under Assumption \ref{convex}, by setting the learning rate $\eta = \frac{D_F}{\sqrt{2}}$ and $M=1$, where $D_F= \Norm{X^*}_F$ is the Frobenius norm of the optimal decision $X^* \in \argmin_{X\in \R^{m\times n}} f(X)$, S-ASGO under the primal-dual framework ensures
    \begin{equation*}
        \frac{1}{T}\sum_{t=1}^T \E\left[f(X_t)- f(X^*)\right] \leq O\left(\frac{D_F\operatorname{tr}(L_T^{1/2})}{T} + \frac{D_F\sqrt{\rho_{1:T}^L}}{T}\right).
    \end{equation*}
    Under Assumption \ref{convex}, by setting the learning rate $\eta = \frac{D_2}{\sqrt{2}}$ and $M=1$, where $D_2= \max_{t\in[T]}\Norm{X_t - X^*}_2$, S-ASGO under the mirror descent framework ensures
    \begin{equation*}
        \frac{1}{T}\sum_{t=1}^T \E\left[f(X_t)- f(X^*)\right] \leq O\left(\frac{D_2 \operatorname{tr}(L_T^{1/2})}{T} + \frac{D_2m \sqrt{\rho_{1:T}^L}}{T}\right).
    \end{equation*}
\end{theorem}

We next turn to smooth losses. In line with \citet{an2025asgo}, we impose two standard assumptions.

\begin{assumption}[Smoothness] \label{smooth2}The loss function is $1$-smooth with respect to $\Norm{\cdot}_\L$, where $\L \in \R^{m\times m}$ is a symmetric positive definite matrix and $\Norm{X}_\L= \sqrt{\operatorname{tr}\left(X^\top\L X\right)}$.
\end{assumption}

\begin{assumption}[Bounded noise variance] \label{noise2} Let $N_t = \nabla f(X_t;\xi_t) - \nabla f(X_t)$ be the stochastic gradient noise. We assume $\E [ N_t ] = 0$ and there exists a symmetric positive definite matrix $V$ such that
\begin{equation*}
    \E[N_t N_t^\top] \preceq V^2.
\end{equation*}
\end{assumption}

Then, we establish the convergence rate for smooth loss functions.

\begin{theorem}\label{thm:jmlr2}
Under Assumptions \ref{convex}, \ref{smooth2} and \ref{noise2}, by setting the learning rate $\eta = \frac{D_F}{\sqrt{2}}$ and the batch size $M$, where $D_F= \Norm{X^*}_F$ is the Frobenius norm of the optimal decision $X^*$, S-ASGO under the primal-dual framework ensures
    \begin{equation*}
        \frac{1}{T}\sum_{t=1}^T \E\left[f(X_t)- f(X^*)\right] \leq O\left(\frac{D_F^2 \operatorname{tr}(\L)}{T} + \frac{D_F \Norm{V}_{*}}{\sqrt{MT}}+ \frac{D_F\sqrt{\rho_{1:T}^L}}{T}\right).
    \end{equation*}
    Under Assumptions \ref{convex}, \ref{smooth2} and \ref{noise2}, by setting the learning rate $\eta = \frac{D_2}{\sqrt{2}}$ and the batch size $M$, where $D_2= \max_{t\in[T]}\Norm{X_t - X^*}_2$, S-ASGO under the mirror descent framework ensures
    \begin{equation*}
        \frac{1}{T}\sum_{t=1}^T \E\left[f(X_t)- f(X^*)\right] \leq O\left(\frac{D_2^2 \operatorname{tr}(\L)}{T} + \frac{D_2 \Norm{V}_{*}}{\sqrt{MT}}+ \frac{mD_2\sqrt{\rho_{1:T}^L}}{T}\right).
    \end{equation*}
\end{theorem}

\begin{rem}\textnormal{
Both bounds contain three terms: a smooth term of order 
$O(\tr(\L)/T)$, a stochastic noise term of order $O(\|V\|_*/\sqrt{MT})$, 
and a sketching error term of order $O(\sqrt{\rho_{1:T}^L}/T)$. The first 
two terms match the convergence rate of their non-sketched 
counterparts, while the third quantifies the additional error 
introduced by FD.  When the gradient covariance is approximately low-rank,  both methods recover the optimal $O(\|V\|_*/\sqrt{MT})$ rate dictated by the stochastic noise. Moreover, a faster convergence rate can be obtained by choosing  a  large batch size $M$.}
\end{rem}

\section{Analysis}\label{section6}
In this section, we present the proofs of the main theorems.

\subsection{Proof of Theorem \ref{thm1}}

Before giving the proof of Theorem \ref{thm1}, we first introduce some supporting lemmas.

\begin{lemma}[{\citealp[Proposition 2]{duchi2011adaptive}}]
\label{duchi1}
    Let $\{\x_t\}$ be the decisions of Algorithm \ref{alg:ftrl1} and $\x^* \in \arg\min_{\x \in \mathbb{R}^d} \sum_{t=1}^Tf_t(\x)$. Then we have
    \begin{equation*}
        \sum_{t=1}^T f_t\left(\x_t\right) - \sum_{t=1}^Tf_t\left(\x^*\right) \leq \frac{1}{\eta} \Psi_T\left(\x^*\right) + \frac{\eta}{2} \sum_{t=1}^T \|\g_t\|^2_{\tilde{G}^{-1/2}_{t-1}},
    \end{equation*}
    where $ \Psi_T(\x^*) = \frac{1}{2}\langle\x^*,\tilde{G}^{1/2}_T\x^* \rangle$.
\end{lemma}

\begin{lemma}[{\citealp[Lemma 10]{duchi2011adaptive}}]
\label{duchi2} Define  $G_t = \sum_{i=1}^t \g_i\g_i^{\top}$. Then we have
\begin{equation*}
\sum_{t=1}^T \left\langle \mathbf{g}_t, \left(G_t^{1/2}\right)^{-1} \mathbf{g}_t \right\rangle 
\leq 2 \sum_{t=1}^T \left\langle \mathbf{g}_t, \left(G_T^{1/2}\right)^{-1} \mathbf{g}_t \right\rangle 
= 2 \operatorname{tr}\left(G_T^{1/2}\right).
\end{equation*}
\end{lemma}

Then, we illustrate how FD approximates the original matrix.
\begin{lemma}[{\citealp[Remark 11]{feinberg2024sketchy}}]\label{lem1} By defining $G_t = \sum_{i=1}^t \g_i\g_i^{\top}$, for the FD sketch maintained in FTSL, we have
\begin{equation*}
B_tB_t^{\top} \preceq G_t\preceq \tilde{G}_t=\rho_{1:t}I_{d\times d} + B_tB_t^{\top}.
\end{equation*}
\end{lemma}

Using Lemma \ref{duchi1}, we can bound the regret of Algorithm \ref{alg:ftrl1} by
\begin{equation}\label{regret1}
    \sum_{t=1}^T f_t\left(\x_t\right) - \sum_{t=1}^Tf_t\left(\x^*\right) \leq \underbrace{\frac{1}{\eta} \Psi_T\left(\x^*\right)}_{R_D}
+ \underbrace{\frac{\eta}{2} \sum_{t=1}^T \|\g_t\|^2_{\tilde{G}^{-1/2}_{t-1}}}_{R_G}.
\end{equation}

Then we bound the terms $R_D$ and $R_G$, respectively. As for $R_D$, we have 

\begin{align*}
\Psi_T\left(\x^*\right)& = \frac{1}{2} 
    \left\langle\x^*,\tilde{G}^{1/2}_T\x^* \right\rangle= \frac{1}{2} \left\langle\x^*,\left(B_TB_T^{\top} + \rho_{1:T}I_{d\times d}\right)^{1/2}\x^* \right\rangle\\
    &\leq \frac{1}{2} \left\langle\x^*,\left(B_TB_T^{\top}\right)^{1/2}\x^* \right\rangle+ \frac{1}{2}  \left\langle\x^*,\left(\rho_{1:T}I_{d\times d}\right)^{1/2}\x^* \right\rangle\\
    &\leq \frac{1}{2} \sqrt{\rho_{1:T}}\Norm{\x^*}^2 + \frac{1}{2}  \left\langle\x^*,\left(G_T\right)^{1/2}\x^* \right\rangle\\
    & \leq \frac{1}{2} D^2 \sqrt{\rho_{1:T}} +\frac{1}{2}  D^2 \lambda_{\max}\left(\left(G_T\right)^{1/2}\right)\\
    & \leq \frac{1}{2}  D^2 \sqrt{\rho_{1:T}} + \frac{1}{2} D^2 \operatorname{tr}\left(G_T^{1/2}\right),
\end{align*}
where the first inequality is due to $\sqrt{a+b} \leq \sqrt{a} +\sqrt{b}$ and we assume $\Norm{\x^*} \leq D$.

Therefore, we have 
\begin{equation}\label{rd1}
    R_D  = \frac{1}{\eta}\Psi_T\left(\x^*\right) \leq \frac{1}{2\eta}\left(D^2 \sqrt{\rho_{1:T}} + D^2 \operatorname{tr}\left(G_T^{1/2}\right)\right).
\end{equation}

To give the bound for $R_G$, we first give the lower bound of $\tilde{G}_{t-1}$, which connects $\tilde{G}_{t-1}$ with $G_t$. We denote $a_t = \min\left\{\frac{\rho_{1:t-1}}{\Norm{\g_t}^2+\rho_{1:t-1}},1\right\} \leq 1$, and we have

\begin{align*}
    \tilde{G}_{t-1} &= B_{t-1}B_{t-1}^{\top} + \rho_{1:t-1}I_{d\times d}\\
    & \succeq a_t\left(\Norm{\g_t}^2I_{d\times d} + \rho_{1:t-1}I_{d\times d} + B_{t-1}B_{t-1}^{\top}\right) \\
    & \succeq a_t\left(\Norm{\g_t}^2I_{d\times d} + G_{t-1}\right)\\
    & \succeq a_tG_t,
\end{align*}
which means $\tilde{G}_{t-1}^{-1} \preceq \frac{1}{a_t}G_t^{-1}$.

Then, letting $C_1 = \max_{t\in[T]} \frac{1}{\sqrt{a_t}}$, we have
\begin{align*}
    \sum_{t=1}^T \Norm{\g_t}_{\tilde{G}^{-1/2}_{t-1}}^2 & = \sum_{t=1}^T\left\langle\g_t, \left(\tilde{G}^{1/2}_{t-1}\right)^{-1}\g_t\right\rangle\leq \sum_{t=1}^T\frac{1}{\sqrt{a_t}}\left\langle\g_t, \left(G^{1/2}_{t}\right)^{-1}\g_t\right\rangle\\
    &\leq C_1\sum_{t=1}^T\left\langle\g_t, \left(G^{1/2}_{t}\right)^{-1}\g_t\right\rangle\leq 2C_1\operatorname{tr}\left(G_T^{1/2}\right),
\end{align*}
where the last inequality is due to Lemma \ref{duchi2}.

Therefore, we have

\begin{equation}\label{rg1}
    R_G = \frac{\eta}{2}\sum_{t=1}^T \Norm{\g_t}_{\tilde{G}^{-1/2}_{t-1}}^2 \leq \eta C_1\operatorname{tr}\left(G_T^{1/2}\right).
\end{equation}

By combining equations (\ref{rd1}) and (\ref{rg1}), and setting $\eta = \frac{D}{\sqrt{2}}$, we obtain
\begin{align*}
    \sum_{t=1}^T f_t\left(\x_t\right) - \sum_{t=1}^T f_t\left(\x^*\right)  &\leq \eta C_1\operatorname{tr}\left(G_T^{1/2}\right) + \frac{1}{2\eta}\left(D^2 \sqrt{\rho_{1:T}} + D^2 \operatorname{tr}\left(G_T^{1/2}\right)\right)\\
    & \leq D\sqrt{2\rho_{1:T}} + \left(\frac{C_1+1}{\sqrt{2}}\right)D\operatorname{tr}\left(G_T^{1/2}\right)\\
    & =O\left(D\left(\operatorname{tr}\left(G_T^{1/2}\right) + \sqrt{\rho_{1:T}}\right)\right).
\end{align*}

\subsection{Proof of Theorem \ref{thm:ffd1}}

We first introduce some guarantees of the fast frequent directions technique.

In Algorithm \ref{alg:ffd}, we do not perform SVD on the sketching matrix $B_t$ every round. Instead, we maintain two matrices $M_t$ and $V_t$, which approximate the sketching matrix $B_t$ in  Algorithm \ref{alg:ftrl1}, that is 
\begin{equation*}
    V_tM_t^{1/2} = B_t \in \mathbb{R}^{d \times 2\tau}.
\end{equation*}
In each round $t$, after receiving a new gradient $\g_t$, we first check whether this vector lies within the subspace spanned by $V_{t-1}$. If the vector is not contained within the subspace, we normalize it and subsequently add it to $V_{t-1}$,  thereby enlarging the span of the subspace and ensuring $V_{t-1}V_{t-1}^{\top}\g_t=\g_{t}$. In our algorithm design, we require the matrix $V_t$ only contain orthonormal vectors, therefore we add $\frac{\mathbf{g}_t - \mathbf{g}'}{\|\mathbf{g}_t - \mathbf{g}'\|_2}$ to the first all zero column. In round $t$, we have 
\begin{align*}
V_{t-1} M_{t-1} V_{t-1}^\top + \mathbf{g}_t \mathbf{g}_t^\top 
&= V_{t-1} M_{t-1} V_{t-1}^\top + V_{t-1} V_{t-1}^\top \mathbf{g}_t \mathbf{g}_t^\top V_{t-1} V_{t-1}^\top \\
&= V_{t-1} \left( M_{t-1} + V_{t-1}^\top \mathbf{g}_t \mathbf{g}_t^\top V_{t-1} \right) V_{t-1}^\top\\
& = V_{t} \left( M_{t-1} + V_{t}^\top \mathbf{g}_t \mathbf{g}_t^\top V_{t} \right) V_{t}^\top.
\end{align*}

According to Lemma \ref{lem1}, we can derive 
\begin{equation*}
    V_tM_tV_t^{\top} = B_t B_t^{\top} \preceq G_t.
\end{equation*}

In the following, we will prove $G_t \preceq \tilde{G}_t$.

 Assume that we delete the eigenvalues of $M_k$ in round $k$. Note that in our method, we calculate $\tilde{G}_k$ before we delete the eigenvalues of $M_k$. In round $k$, before we update $V_k$, we have $V_k = V_{k-1}$, and we let $V_k$ represent the matrix before we delete its columns.

As we perform SVD on $V_kM_k^{1/2} \in \mathbb{R}^{d\times 2\tau}$, we delete the eigenvalues of $M_k$, update $V_k = V_kU_k$ and set the last $\tau+1$ columns to zero. In round $k+1$, we have $\tilde{G}_{k+1}=V_{k+1}(\text{diag}(\max\left\{\lambda^{(k)}_{[1:2\tau]}-\lambda_{\tau}^{(k)},0\right\})  + V_{k+1}^{\top}\g_{k+1}(V_{k+1}^{\top}\g_{k+1})^{\top})V_{k+1}^{\top} + \rho_{1:k+1}I_{d\times d}$, and we can derive
\begin{align*}
    &V_{k+1}\left(\text{diag}\left(\max\left\{\lambda^{\left(k\right)}_{\left[1:2\tau\right]}-\lambda_{\tau}^{\left(k\right)},0\right\}\right)  + V_{k+1}^{\top}\g_{k+1}\left(V_{k+1}^{\top}\g_{k+1}\right)^{\top}\right)V_{k+1}^{\top} + \rho_{k+1}I_{d\times d}\\
    =& V_{k+1}\text{diag}\left(\max\left\{\lambda^{\left(k\right)}_{\left[1:2\tau\right]}-\lambda_{\tau}^{\left(k\right)},0\right\}\right)  V_{k+1}^{\top} + \lambda_{\tau}^{\left(k\right)}I_{d\times d} + \g_{k+1}\g_{k+1}^{\top}\\
     \succeq &  V_{k}U_k\text{diag}\left(\max\left\{\lambda^{\left(k\right)}_{\left[1:2\tau\right]}-\lambda_{\tau}^{\left(k\right)},0\right\}\right) U_k^{\top} V_{k}^{\top} + \lambda_{\tau}^{\left(k\right)}\left(V_{k}U_k\right)\left(V_{k}U_k\right)^{\top} + \g_{k+1}\g_{k+1}^{\top}\\
     =&  V_{k}U_k\text{diag}\left(\max\left\{\lambda^{\left(k\right)}_{\left[1:2\tau\right]}-\lambda_{\tau}^{\left(k\right)},0\right\}+\lambda_{\tau}^{\left(k\right)}\right) U_k^{\top} V_{k}^{\top} +  \g_{k+1}\g_{k+1}^{\top}\\
    \succeq & V_{k}U_k\text{diag}\left(\lambda^{\left(k\right)}_{\left[1:2\tau\right]}\right)U_k^{\top}V_{k}^{\top}+ \g_{k+1}\g_{k+1}^{\top}\\
    = &  V_{k}M_{k}V_{k}^{\top}+ \g_{k+1}\g_{k+1}^{\top},
\end{align*}
where the second inequality is due to the fact that only the first $\tau-1$ columns of $\text{diag}(\max\left\{\lambda^{(k)}_{[1:2\tau]}-\lambda_{\tau}^{(k)},0\right\})$ are non-zero and the first $\tau-1$ columns of  $V_{k+1}$ and $V_{k}U_k$ are the same, and that $U_k$ and $V_k$ are orthogonal matrices.

If we do not delete the eigenvalues of $M_k$ in round $k$, we have  $\rho_{k+1}=0$ and
\begin{align*}
    V_{k+1} M_{k+1} V_{k+1}^{\top} + \rho_{k+1}I_{d\times d}&= V_{k+1}\left(M_{k}+ V_{k+1}^{\top} \g_{k+1}\left(V_{k+1}^{\top} \g_{k+1}\right)^{\top}\right)V_{k+1}^{\top} \\
    &= V_{k+1}M_{k}V_{k+1}^{\top} + \g_{k+1}\g_{k+1}^{\top}.
\end{align*} 

Assume there are at most $\ell$ eigenvalues in $M_k$, $\ell \leq 2\tau - 1$, at most $\ell+1$ non-zero columns in $V_{k+1}$, and at most $\ell$ non-zero columns in $V_{k}$, and the first $\ell$ columns of $V_{k+1}$ and $V_{k}$ are the same. We have the following 
\begin{equation}
    \begin{aligned}
    V_{k+1} M_{k+1} V_{k+1}^{\top}+\rho_{k+1}I_{d\times d} &=   V_{k+1}U_k\text{diag}\left(\lambda_{\left[1:2\tau\right]}^{\left(k\right)}\right)U_k^{\top}V_{k+1}^{\top} + \g_{k+1}\g_{k+1}^{\top} + 0I_{d\times d}\\
    & =   V_{k}U_k\text{diag}\left(\lambda_{\left[1:2\tau\right]}^{\left(k\right)}\right)U_k^{\top}V_{k}^{\top} + \g_{k+1}\g_{k+1}^{\top}\\
    &= V_{k}M_kV_{k}^{\top} + \g_{k+1}\g_{k+1}^{\top},
\end{aligned} 
\end{equation}
where the second equality is due to the fact that only the first $\ell$ elements in $\lambda_{[1:2\tau]}^{(k)}$ are non-zero and the first $\ell$ columns of $V_{k}$ and $V_{k+1}$ are the same.

Therefore, we have
\begin{equation*}
\tilde{G}_t = V_{t} M_{t} V_{t}^{\top}+\rho_{1:t}I_{d\times d}\succeq    V_{t-1} M_{t-1} V_{t-1}^{\top}+\rho_{1:t-1}I_{d\times d}+\g_t\g_t^{\top}= \tilde{G}_{t-1} + \g_t\g_t^{\top}.
\end{equation*}
By summing up, we have
\begin{align*}
    \tilde{G}_t \succeq \sum_{i=1}^t\g_i\g_i^{\top} = G_t.
\end{align*}
Next, we need to ensure that after FFD, the preconditioning matrix $\tilde{G}_t$ is monotone. It is easy to verify that $\tilde{G}_t$ remains monotone in the rounds where we do not delete any eigenvalue of $M_t$.

Then, we prove that $\tilde{G}_t$ remains monotone even if we delete the eigenvalues of $M_t$. Assume in round $k$, we delete the eigenvalues of $M_k$. In round $k$, the matrix $\tilde{G}_{k} = \rho_{1:k}I_{d\times d} + V_{k}U_k\text{diag}(\lambda^{(k)}_{[1:2\tau]})U_k^{\top}V_{k}^{\top}$. Moreover, $\tilde{G}_{k+1} = \rho_{1:k+1}I_{d\times d} + V_{k+1}M_{k+1}V_{k+1}^{\top}$. In round $k+1$, since we do not delete the eigenvalues of $M_{k+1}$, the first $\tau-1$ columns of  $V_{k+1}$ of round $k+1$ and $V_kU_k$ of round $k$ are the same (notably, $V_k$ is different in rounds $k$ and $k+1$).  We have $M_{k+1} = \text{diag}(\max\left\{\lambda^{(k)}_{[1:2\tau]} - \lambda_{\tau}^{(k)},0\right\}) + (V_{k+1}^{\top}\g_{k+1}) (V_{k+1}^{\top}\g_{k+1})^{\top}$. As we set $\rho_{k+1} = \lambda_{\tau}^{(k)}$, we can ensure
\begin{align*}
    \tilde{G}_{k+1} &= \rho_{1:k+1}I_{d\times d} + V_{k+1}M_{k+1}V_{k+1}^{\top}\\
     &= \rho_{1:k+1}I_{d\times d} + V_{k+1}\left(\text{diag}\left(\max\left\{\lambda^{\left(k\right)}_{\left[1:2\tau\right]} - \lambda_{\tau}^{\left(k\right)},0\right\}\right) +V_{k+1}^{\top}\g_{k+1} \left(V_{k+1}^{\top}\g_{k+1}\right)^{\top}\right)V_{k+1}^{\top}\\
    & \succeq \rho_{1:k}I_{d\times d} + \lambda_{\tau}^{\left(k\right)} I_{d \times d} + V_{k+1} \text{diag}\left(\max\left\{\lambda^{\left(k\right)}_{\left[1:2\tau\right]} - \lambda_{\tau}^{\left(k\right)},0\right\}\right)V_{k+1}^{\top}\\
        & \succeq \rho_{1:k}I_{d\times d} + V_{k} U_k \lambda_{\tau}^{\left(k\right)} I_{d \times d} U_k^{\top}V_{k}^{\top}\\
    &\quad + V_{k} U_k\text{diag}\left(\max\left\{\lambda^{\left(k\right)}_{\left[1:2\tau\right]} - \lambda_{\tau}^{\left(k\right)},0\right\}\right)U_k^{\top}V_{k}^{\top}\\
    & \succeq \rho_{1:k}I_{d\times d}  + V_{k} U_k\text{diag}\left(\lambda^{\left(k\right)}_{\left[1:2\tau\right]}\right)U_k^{\top}V_{k}^{\top}= \tilde{G}_{k}, 
\end{align*}
where the second inequality is due to the fact that the $\tau$-th to $2\tau$-th columns of the matrix $\text{diag}(\max\left\{\lambda^{(k)}_{[1:2\tau]}- \lambda_{\tau}^{(k)},0\right\})$ are zero and the first $\tau-1$ columns of $V_kU_k$ of round $k$ and $V_{k+1}$ of round $k+1$ are the same, so we can replace $V_{k+1}$ with $V_{k}U_k$, and $V_k, U_k$ are orthogonal matrices.

Therefore, we can ensure $\tilde{G}_t$ is monotone in FTFSL. Using Eq.~\eqref{regret1}, we have the following
\begin{equation*}
        \sum_{t=1}^T f_t\left(\x_t\right) - \sum_{t=1}^T f_t\left(\x^*\right) \leq \underbrace{\frac{1}{\eta} \Psi_T\left(\x^*\right)}_{R_D}
+ \underbrace{\frac{\eta}{2} \sum_{t=1}^T \|\g_t\|^2_{\tilde{G}^{-1/2}_{t-1}}}_{R_G}.
\end{equation*}

As for the term $R_D$, we can derive 
\begin{equation}\label{ffdprof1}
\begin{aligned}
    R_D & = \frac{1}{\eta}\Psi_T\left(\x^*\right) =\frac{1}{2\eta}\left\langle\x^*,\tilde{G}^{1/2}_T\x^* \right\rangle\\
    &= \frac{1}{2\eta}\left\langle\x^*,\left(V_{T}M_TV_{T}^{\top} + \rho_{1:T}I_{d \times d}\right)^{1/2}\x^* \right\rangle\\
    &\leq \frac{1}{2\eta}\left\langle\x^*,\left(V_{T}M_TV_{T}^{\top}\right)^{1/2}\x^* \right\rangle+\frac{1}{2\eta}\left\langle\x^*,\left(\rho_{1:T}I_{d \times d}\right)^{1/2}\x^* \right\rangle\\
    &\leq \frac{1}{2\eta}\Norm{\left(\rho_{1:T}I_{d\times d}\right)^{1/2}}\Norm{\x^*}^2 + \frac{1}{2\eta}\left\langle\x^*,\left(G_T\right)^{1/2}\x^* \right\rangle\\
    & \leq\frac{1}{2\eta} D^2 \sqrt{\rho_{1:T}} + \frac{1}{2\eta}D^2 \lambda_{\max}\left(G_T^{1/2}\right)\\
    & \leq \frac{1}{2\eta}D^2 \sqrt{\rho_{1:T}} + \frac{1}{2\eta}D^2 \operatorname{tr}\left(G_T^{1/2}\right),
\end{aligned}    
\end{equation}

where the second inequality is due to $V_TM_TV_T^{\top} = B_TB_T^{\top} \preceq G_T$  and we assume $\Norm{\x^*} \leq D$.

For the term $R_G$, we denote $b_t = \min\left\{\frac{\rho_{1:t-1}}{\Norm{\g_t}^2+\rho_{1:t-1}},1\right\} \leq 1$, we have

\begin{align*}
    \tilde{G}_{t-1} &=V_{t-1}M_{t-1}V_{t-1}^{\top} + \rho_{1:t-1}I_{d\times d}\\
    & \succeq b_t\left(\Norm{\g_t}^2I_{d \times d} + \rho_{1:t-1}I_{d\times d} + B_{t-1}B_{t-1}^{\top}\right) \\
    & \succeq b_t\left(\Norm{\g_t}^2I_{d \times d}  + G_{t-1}\right) \succeq b_tG_t.
\end{align*}

Then, letting $C_2= \max_{t \in [T]} \frac{1}{\sqrt{b_t}}$, we have
\begin{align*}
    \sum_{t=1}^T \Norm{\g_t}_{\tilde{G}^{-1/2}_{t-1}}^2 & = \sum_{t=1}^T\left\langle\g_t, \left(\tilde{G}^{1/2}_{t-1}\right)^{-1}\g_t\right\rangle \leq \sum_{t=1}^T\frac{1}{\sqrt{b_t}}\left\langle\g_t, \left(G^{1/2}_{t}\right)^{-1}\g_t\right\rangle\\
    &\leq C_2\sum_{t=1}^T\left\langle\g_t, \left(G^{1/2}_{t}\right)^{-1}\g_t\right\rangle \leq 2C_2\operatorname{tr}\left(G_T^{1/2}\right),
\end{align*}
where the last inequality is due to Lemma \ref{duchi2}.

Therefore, we have

\begin{equation}\label{ffdprof2}
    R_G = \frac{\eta}{2}\sum_{t=1}^T \Norm{\g_t}_{\tilde{G}^{-1/2}_{t-1}}^2 \leq \eta C_2\operatorname{tr}\left(G_T^{1/2}\right).
\end{equation}

By combining (\ref{ffdprof1}) and (\ref{ffdprof2}), and setting $\eta = \frac{D}{\sqrt{2}}$, we can derive Theorem \ref{thm:ffd1}.

\begin{align*}
    R\left(T\right) &\leq R_D+R_G\\
    &\leq \frac{1}{2\eta}D^2 \sqrt{\rho_{1:T}} + \frac{1}{2\eta}D^2 \operatorname{tr}\left(G_T^{1/2}\right) +\eta C_2\operatorname{tr}\left(G_T^{1/2}\right) \\
    &\leq O\left(D\left(\operatorname{tr}\left(G_T^{1/2}\right) + \sqrt{\rho_{1:T}}\right)\right).
\end{align*}

\subsection{Proof of Theorem \ref{thm:4}}

According to the proof of Theorem \ref{thm:ffd1}, we have the following properties in FFD:
\begin{align*}
      V_tM_t^{1/2} &= B_t,\\
     V_t U_t \Sigma_t U_t^\top& V_t^\top \preceq G_t \preceq \tilde{G}_t.
\end{align*}
Similar to the proof of Theorem 3 in \citet{feinberg2024sketchy}, we have the following lemma:
\begin{lemma}\label{lem:5}
    Let $\{\x_t\}$ be the decisions of Fast S-ADA and $\x^* \in \arg\min_{\x \in \mathbb{R}^d} \sum_{t=1}^Tf_t(\x)$. Then the regret bound of Fast S-ADA is 

    \begin{equation*}
    R(T) \leq \frac{1}{2\eta} \sum_{t=1}^T \left\|\x_t - \x^*\right\|^2_{\tilde{G}_t^{1/2} - \tilde{G}_{t-1}^{1/2}}
+ \frac{\eta}{2} \sum_{t=1}^T \left\|\g_t\right\|_{\tilde{G}_t^{-1/2}}^2.
\end{equation*}
\end{lemma}
According to Lemma \ref{lem:5}, we have
\begin{equation*}
    R(T) \leq \frac{1}{2\eta} \sum_{t=1}^T \left\|\x_t - \x^*\right\|^2_{\tilde{G}_t^{1/2} - \tilde{G}_{t-1}^{1/2}}
+ \frac{\eta}{2} \sum_{t=1}^T \left\|\g_t\right\|_{\tilde{G}_t^{-1/2}}^2.
\end{equation*}

We first bound the term $ \frac{\eta}{2} \sum_{t=1}^T \left\|\g_t\right\|_{\tilde{G}_t^{-1/2}}^2$, which is easier to bound than that in FTFSL, as we can directly apply a lemma to it. 
\begin{align*}
    \sum_{t=1}^T \Norm{\g_t}_{\tilde{G}^{-1/2}_{t}}^2 & = \sum_{t=1}^T\left\langle\g_t, (\tilde{G}^{1/2}_{t})^{-1}\g_t\right\rangle\\
    &\leq\sum_{t=1}^T\left\langle\g_t, (G^{1/2}_{t})^{-1}\g_t\right\rangle\\
    &\leq 2\operatorname{tr}(G_T^{1/2}),
\end{align*}
where the first inequality is due to $\tilde{G}_t^{-1} \preceq G_t^{-1}$ and the last inequality is due to Lemma \ref{duchi2}.

Next, we bound the term $\frac{1}{2\eta} \sum_{t=1}^T \left\|\x_t - \x^*\right\|^2_{\tilde{G}_t^{1/2} - \tilde{G}_{t-1}^{1/2}}$, which introduces the dependence on  dimensionality $d$. We have
\begin{align*}
    \frac{1}{2\eta} \sum_{t=1}^T \left\|\x_t - \x^*\right\|^2_{\tilde{G}_t^{1/2} - \tilde{G}_{t-1}^{1/2}} &\leq\frac{D_1^2}{2\eta} \operatorname{tr}\left(\tilde{G}_T^{1/2}\right),
\end{align*}
where this inequality is due to monotonicity of $\tilde{G}_t$ and $D_1 = \max_{t\in [T]}\Norm{\x_t-\x^*}$.

Then we need to bound the term $\operatorname{tr}\left(\tilde{G}_T^{1/2}\right)$.

Assume that we delete the eigenvalues of $M_k$ at round $k$. In round $k$, before we update $V_k$, $V_k = V_{k-1}$ and we use $V_k$ to represent the matrix before we delete its columns.

As we perform SVD on $V_kM_k^{1/2} \in \mathbb{R}^{d\times 2\tau}$, we denote by $V_{k}^{1:\tau} \in \mathbb{R}^{d\times \tau}$ the first $\tau$ columns of $V_{k}U_k$ before we set the last $\tau+1$ columns to $\textbf{0}_d$, and by $N_k \in \mathbb{R}^{d\times (d-\tau)}$ the complementary subspace of $V_{k}^{1:\tau}$. Note that the first $\tau-1$ columns of $V_{k+1}$ and $V_{k}^{1:\tau}$ are the same, $\rho_{k+1} = \lambda_{\tau}^{(k)}$, and $[V_{k}^{1:\tau};N_k][V_{k}^{1:\tau};N_k]^{\top} = I_{d\times d}$. 

In round $k+1$, we have
\begin{equation*}
\tilde{G}_{k+1}=V_{k+1}(\text{diag}(\max\left\{\lambda^{(k)}_{[1:2\tau]}-\lambda_{\tau}^{(k)},0\right\})  + V_{k+1}^{\top}\g_{k+1}(V_{k+1}^{\top}\g_{k+1})^{\top})V_{k+1}^{\top} + \rho_{1:k+1}I_{d\times d},
\end{equation*}
and we can derive
\begin{align*}
    &V_{k+1}(\text{diag}(\max\left\{\lambda^{(k)}_{[1:2\tau]}-\lambda_{\tau}^{(k)},0\right\})  + V_{k+1}^{\top}\g_{k+1}(V_{k+1}^{\top}\g_{k+1})^{\top})V_{k+1}^{\top} + \rho_{k+1}I_{d\times d}\\
    =& V_{k+1}\text{diag}(\max\left\{\lambda^{(k)}_{[1:2\tau]}-\lambda_{\tau}^{(k)},0\right\})  V_{k+1}^{\top} + \lambda_{\tau}^{(k)}I_{d\times d} + \g_{k+1}\g_{k+1}^{\top}\\
     =&  V_{k}U_k\text{diag}(\max\left\{\lambda^{(k)}_{[1:2\tau]}-\lambda_{\tau}^{(k)},0\right\}) U_k^{\top} V_{k}^{\top} + \lambda_{\tau}^{(k)}( V_{k}^{1:\tau} (V_{k}^{1:\tau})^{\top} + N_kN_{k}^{\top}) + \g_{k+1}\g_{k+1}^{\top}\\
     \preceq &  V_{k}U_k\text{diag}(\max\left\{\lambda^{(k)}_{[1:2\tau]}-\lambda_{\tau}^{(k)},0\right\}) U_k^{\top} V_{k}^{\top} + \lambda_{\tau}^{(k)}( V_{k}U_k ( V_{k}U_k)^{\top} + N_kN_{k}^{\top})+\g_{k+1}\g_{k+1}^{\top}\\
    \preceq & V_{k}U_k\text{diag}\left(\lambda^{(k)}_{[1:2\tau]}\right)U_kV_{k}^{\top} +  \lambda_{\tau}^{(k)}N_kN_k^{\top} + \g_{k+1}\g_{k+1}^{\top}\\
    = & \lambda_{\tau}^{(k)}N_kN_k^{\top} +  V_{k}M_{k}V_{k}^{\top}+ \g_{k+1}\g_{k+1}^{\top}\\
    = & \rho_{k+1}N_kN_k^{\top} +  V_{k}M_{k}V_{k}^{\top}+ \g_{k+1}\g_{k+1}^{\top},
\end{align*}
where the second equality is due to the fact that only the first $\tau-1$ columns of the matrix $\text{diag}\left(\max\left\{\lambda^{(k)}_{[1:2\tau]}-\lambda_{\tau}^{(k)},0\right\}\right)$ are non-zero and the first $\tau-1$ columns of  $V_{k+1}$ and $V_{k}U_k$ are the same, and the first inequality is due to the fact that $V_{k}^{1:\tau}$ contains at most $\tau$ orthonormal vectors.

If we do not delete the eigenvalues of $M_k$ in round $k$, we have $\rho_{k+1}=0$ and can derive
\begin{align*}
    V_{k+1} M_{k+1} V_{k+1}^{\top} + \rho_{k+1}I_{d\times d} &= V_{k+1}(M_{k}+ V_{k+1}^{\top} \g_{k+1}(V_{k+1}^{\top} \g_{k+1})^{\top})V_{k+1}^{\top} \\
    &= V_{k+1}M_{k}V_{k+1}^{\top} + \g_{k+1}\g_{k+1}^{\top}\\
     &= 0N_kN_k^{\top} +  V_{k+1}M_{k}V_{k+1}^{\top} + \g_{k+1}\g_{k+1}^{\top}\\
    &=\rho_{k+1}N_kN_k^{\top} +  V_{k+1}U_k\text{diag}(\lambda_{[1:2\tau]}^{(k)})U_k^{\top}V_{k+1}^{\top} + \g_{k+1}\g_{k+1}^{\top}.
\end{align*} 

Assume there are at most $\ell$ eigenvalues in $M_k$, $\ell \leq 2\tau - 1$. Therefore, there are at most $\ell+1$ non-zero columns in $V_{k+1}$ and at most $\ell$ non-zero columns in $V_{k}$, and the first $\ell$ columns of $V_{k+1}$ and $V_{k}$ are the same. We have the following 
\begin{equation}\label{eq:2}
    \begin{aligned}
    V_{k+1} M_{k+1} V_{k+1}^{\top} + \rho_{k+1}I_{d\times d} &= 0N_kN_k^{\top} +  V_{k+1}U_k\text{diag}(\lambda_{[1:2\tau]}^{(k)})U_k^{\top}V_{k+1}^{\top} + \g_{k+1}\g_{k+1}^{\top}\\
    &= 0N_kN_k^{\top} +  V_{k}U_k\text{diag}(\lambda_{[1:2\tau]}^{(k)})U_k^{\top}V_{k}^{\top} + \g_{k+1}\g_{k+1}^{\top}\\
    & = \rho_{k+1}N_kN_k^{\top} +  V_{k}U_k\text{diag}(\lambda_{[1:2\tau]}^{(k)})U_k^{\top}V_{k}^{\top} + \g_{k+1}\g_{k+1}^{\top}\\
    &= \rho_{k+1}N_kN_k^{\top} +  V_{k}M_kV_{k}^{\top} + \g_{k+1}\g_{k+1}^{\top},
\end{aligned} 
\end{equation}
where the second equality is due to the fact that there are at most $\ell$ non-zero elements in $\lambda_{[1:2\tau]}^{(k)}$  and the first $\ell$ columns of $V_{k}$ and $V_{k+1}$ are the same, and the third equality is due to $\rho_{k+1}=0$. 

Therefore, we have 
\begin{equation*}
    V_{k+1} M_{k+1} V_{k+1}^{\top} + \rho_{k+1}I_{d\times d} \preceq\rho_{k+1}N_kN_k^{\top} +  V_{k}M_kV_{k}^{\top} + \g_{k+1}\g_{k+1}^{\top}.
\end{equation*}

By induction, we have
\begin{align*}
    \tilde{G}_T & = V_T M_T V_T^{\top} + \rho_{1:T}I_{d\times d}\\
    & \preceq   \rho_{1:T-1}I_{d\times d} + V_{T-1}M_{T-1}V_{T-1}^{\top} +  \rho_{T}N_TN_T^{\top} + \g_T\g_T^{\top} \\
    & \quad ...\\
    & \preceq \sum_{t=1}^T \rho_t N_tN_t^{\top} + \sum_{t=1}^T\g_t\g_t^{\top} = \sum_{t=1}^T \rho_t N_tN_t^{\top} +G_T.
\end{align*}

Then we can rewrite the bound of $\operatorname{tr}(\tilde{G}_T^{1/2})$ as  $\operatorname{tr}(G_T^{1/2}) + \operatorname{tr}((\sum_{t=1}^T \rho_t N_tN_t^{\top})^{1/2})$.

We just need to bound $\operatorname{tr}((\sum_{t=1}^T \rho_t N_tN_t^{\top})^{1/2})$. According to Corollary 4 in \citet{feinberg2024sketchy}, the upper bound of this term is 
\begin{equation*}
    \operatorname{tr}\left(\left(\sum_{t=1}^T \rho_t N_tN_t^{\top}\right)^{1/2}\right) \leq \sqrt{d(d-\tau)\rho_{1:T}}.
\end{equation*}

Then we can derive the bound of $\operatorname{tr}(\tilde{G}_T^{1/2})$.

\begin{align*}
    \operatorname{tr}(\tilde{G}_T^{1/2}) &\leq \operatorname{tr}(G_T^{1/2}) + \operatorname{tr} ((\sum_{t=1}^T \rho_t N_tN_t^{\top})^{1/2})\\
    & \leq \operatorname{tr}(G_T^{1/2}) + \sqrt{d(d-\tau)\rho_{1:T}}.
\end{align*}

By setting $\eta = \frac{D_1}{\sqrt{2}}$, we have
\begin{align*}
    R(T)&= \sum_{t=1}^T f_t(\x_t) - \sum_{t=1}^T f_t(\x^*)\\
    &\leq \frac{1}{2\eta} \sum_{t=1}^T \left\|\x_t - \x^*\right\|^2_{\tilde{G}_t^{1/2} - \tilde{G}_{t-1}^{1/2}}
+ \frac{\eta}{2} \sum_{t=1}^T \left\|\g_t\right\|_{\tilde{G}_t^{-1/2}}^2\\
    & \leq \frac{D_1^2}{2\eta}(\operatorname{tr}(G_T^{1/2}) + \sqrt{d(d-\tau)\rho_{1:T}}) + \eta\operatorname{tr}(G_T^{1/2})  \\
    &\leq O\left(D_1\left(\operatorname{tr}(G_T^{1/2}) + \sqrt{d(d-\tau)\rho_{1:T}}\right)\right).
\end{align*}

\subsection{Proof of Theorem \ref{thm:p2}}

Before deriving Theorem \ref{thm:p2}, we introduce two useful lemmas.
\begin{lemma}[{\citealp[Lemma 4]{an2025asgo}}]\label{lem:7} For a symmetric positive semidefinite matrix $X \in \R^{n\times n}$, it holds that
    \begin{equation*}
        \operatorname{tr}(X^{1/2})\leq \sum_{j=1}^n\sqrt{[X]_{j,j}}.
    \end{equation*}
\end{lemma}

\begin{lemma}[{\citealp[Lemma G.5]{liu2024adagradanisotropicsmoothness}}]\label{lem:8} For a non-negative sequence $\{a_j\}_{j=1}^n$ and a positive sequence $\{b_j\}_{j=1}^n$ with $B = \sum_{j=1}^n b_j$, it holds that
\begin{equation*}
    \frac{1}{B}\sum_{j=1}^n a_j \leq \sqrt{\frac{1}{B}\sum_{j=1}^n \frac{a_j^2}{b_j}}
\end{equation*}
    
\end{lemma}

\paragraph{Primal-dual framework.} We first present the proof under the primal-dual framework. Based on Theorem \ref{thm:p1}, we have 
\begin{equation}\label{eq:19}
\begin{aligned}
    &\frac{1}{T}\sum_{t=1}^T\E\left[f(\x_t)-f(\x^*)\right]\\
    \leq&\frac{\sqrt{2}D}{T}\left(C_2\operatorname{tr}(G_T^{1/2})+\sqrt{\rho_{1:T}}\right)\leq\frac{\sqrt{2}D}{T}\left(C_2\operatorname{tr}\left(\left(\sum_{t=1}^T \g_t\g_t^\top\right)^{1/2}\right)+\sqrt{\rho_{1:T}}\right)\\
    \leq&\frac{\sqrt{2}D}{T}\left(C_2\operatorname{tr}\left(\left(\sum_{t=1}^T 2\nabla f(\x_t)\nabla f(\x_t)^\top + 2\n_t\n_t^\top\right)^{1/2}\right)+\sqrt{\rho_{1:T}}\right)\\
    \leq&\frac{\sqrt{2}D}{T}\left(C_2\operatorname{tr}\left(\left(\sum_{t=1}^T 2\nabla f(\x_t)\nabla f(\x_t)^\top \right)^{1/2}\right) \right.\\
    &\left.\qquad + C_2\operatorname{tr}\left(\left(\sum_{t=1}^T 2\n_t\n_t^\top\right)^{1/2}\right)+\sqrt{\rho_{1:T}}\right).
\end{aligned}    
\end{equation}

Then we bound each term one by one. As for the first term, we use the smooth assumption and obtain
\begin{equation}\label{eq:17}
\begin{aligned}
    \operatorname{tr}\left(\left(\sum_{t=1}^T 2\nabla f(\x_t)\nabla f(\x_t)^\top \right)^{1/2}\right) &\leq  \sqrt{2}\sum_{j=1}^d\sqrt{\sum_{t=1}^T[\nabla f(\x_t)]_j^2}\\
    &\leq \sqrt{2}\sqrt{\left(\sum_{j=1}^d L_j\right)\left(\sum_{j=1}^d \sum_{t=1}^T[\nabla f(\x_t)]_j^2/L_j\right)}\\
    &\leq \sqrt{2}\sqrt{\operatorname{tr}(\L)\left( \sum_{t=1}^T\sum_{j=1}^d[\nabla f(\x_t)]_j^2/L_j\right)}\\
    &\leq \sqrt{2}\sqrt{\operatorname{tr}(\L)\left(\sum_{t=1}^T\Norm{\nabla f(\x_t)}_{\L^{-1}}^2\right)}\\
    &\leq 2\sqrt{\operatorname{tr}(\L)\sum_{t=1}^T\left(f(\x_t)-f(\x^*)\right)},
\end{aligned}    
\end{equation}
where the first inequality is due to Lemma \ref{lem:7}, the second inequality is due to Lemma \ref{lem:8} and the last inequality is based on the properties of smoothness \citep{nesterov2018lectures}.

As for the stochastic noise term, we introduce a lemma.
\begin{lemma}\label{lem:6}
    Under Assumption \ref{noise1}, it holds that
    \begin{align*}
        \E\left[ \operatorname{tr}(\n_t\n_t^\top) \right] = \Norm{\n_t}^2 \leq \frac{\Norm{\v}^2}{M}.
    \end{align*}
\end{lemma}

Based on Lemma \ref{lem:6}, we have
\begin{equation}\label{eq:18}
\begin{aligned}
    \operatorname{tr}\left(\left(\sum_{t=1}^T 2\n_t\n_t^\top\right)^{1/2}\right)&\leq \sqrt{2} \sum_{j=1}^d \sqrt{\sum_{t=1}^T [\n_{t}]_j^2}\\
    &\leq \sqrt{2}\sqrt{\left(\sum_{j=1}^d [\v]_j^2\right)\left(\sum_{j=1}^d \sum_{t=1}^T[\n_{t}]_j^2/[\v]_j^2\right)}\\
    &\leq \sqrt{2}\sqrt{\Norm{\v}^2\left( \sum_{t=1}^T\sum_{j=1}^d[\n_{t}]_j^2/[\v]_j^2\right)}\leq \sqrt{\frac{2T}{M}}\Norm{\v}.
\end{aligned}    
\end{equation}

Therefore, we can derive the convergence rate for smooth loss functions. By plugging \eqref{eq:17} and \eqref{eq:18} into \eqref{eq:19}, we have
\begin{equation}\label{eq:20}
    \begin{aligned}
    &\frac{1}{T}\sum_{t=1}^T\E\left[f(\x_t)-f(\x^*)\right]\\
    \leq&\frac{\sqrt{2}D}{T}\left(C_2\operatorname{tr}\left(\left(\sum_{t=1}^T 2\nabla f(\x_t)\nabla f(\x_t)^\top \right)^{1/2}\right) + C_2\operatorname{tr}\left(\left(\sum_{t=1}^T 2\n_t\n_t^\top\right)^{1/2}\right)+\sqrt{\rho_{1:T}}\right)\\
    \leq& \frac{\sqrt{2}D}{T}\left(2C_2\sqrt{\operatorname{tr}(\L)\sum_{t=1}^T\left(f(\x_t)-f(\x^*)\right)} + C_2\sqrt{\frac{2T}{M}}\Norm{\v}+\sqrt{\rho_{1:T}}\right).
    \end{aligned}
\end{equation}

If we denote $a = \sqrt{\frac{1}{T} \sum_{t=1}^T\left(f(\x_t)-f(\x^*)\right)}$, we can rewrite \eqref{eq:20} as 
\begin{equation}\label{eq:06222}
    a^2 \leq ba+c,
\end{equation}
where $ b = \frac{2\sqrt{2}DC_2\sqrt{\operatorname{tr}(\L)}}{\sqrt{T}}, c = \frac{2DC_2\Norm{\v}}{\sqrt{MT}} + \frac{\sqrt{2}D\sqrt{\rho_{1:T}}}{T}$. When $a\geq 0$, we have $a^2 \leq 2b^2+2c$, which means
\begin{equation}\label{eq:21}
    \frac{1}{T}\sum_{t=1}^T\E\left[f(\x_t)-f(\x^*)\right]\leq \frac{8D^2C_2^2\operatorname{tr}(\L)}{T}+\frac{4DC_2\Norm{\v}}{\sqrt{MT}}+\frac{2\sqrt{2}D\sqrt{\rho_{1:T}}}{T},
\end{equation}
which concludes the proof. 

\paragraph{Mirror descent framework.} To derive the convergence rate for the mirror descent framework, we directly incorporate Theorem \ref{thm:4} into \eqref{eq:19} and  obtain
\begin{equation}\label{eq:06221}
    \begin{aligned}
    &\frac{1}{T}\sum_{t=1}^T\E\left[f(\x_t)-f(\x^*)\right]\\
    \leq&\frac{\sqrt{2}D_1}{T}\left(\operatorname{tr}\left(\left(\sum_{t=1}^T 2\nabla f(\x_t)\nabla f(\x_t)^\top \right)^{1/2}\right) + \operatorname{tr}\left(\left(\sum_{t=1}^T 2\n_t\n_t^\top\right)^{1/2}\right)+\sqrt{d(d-\tau)\rho_{1:T}}\right)\\
    \leq& \frac{\sqrt{2}D_1}{T}\left(2\sqrt{\operatorname{tr}(\L)\sum_{t=1}^T\left(f(\x_t)-f(\x^*)\right)} + \sqrt{\frac{2T}{M}}\Norm{\v}+\sqrt{d(d-\tau)\rho_{1:T}}\right).
    \end{aligned}
\end{equation}
By using \eqref{eq:06222} again, we have
\begin{align*}
    \frac{1}{T}\sum_{t=1}^T\E\left[f(\x_t)-f(\x^*)\right]&\leq \frac{8D_1^2\operatorname{tr}(\L)}{T} + \frac{4D_1\Norm{\v}}{\sqrt{MT}}+\frac{2\sqrt{2}D_1\sqrt{d(d-\tau)\rho_{1:T}}}{T},
\end{align*}
which completes the proof.
\subsection{Proof of Theorem \ref{thm3}}


In the following, we give the proof of Theorem \ref{thm3}.

Since the update in Algorithm \ref{shampoo} is performed in the matrix space, it poses challenges for the analysis. Therefore, we first introduce an equivalent update in vector form.  

Recall the update in Algorithm \ref{shampoo}, $X_{t+1} = -\eta  \tilde{L}_t^{-1/4}\overline{G}_t^X\tilde{R}_t^{-1/4}$. We define $\tilde{H}_t =  \tilde{L}_t^{1/4} \otimes \tilde{R}_t^{1/4} \in \mathbb{R}^{mn \times mn}, \overline{L}_t = \hat{L}_t \hat{L}_t^{\top} \in \mathbb{R}^{m\times m}, \overline{R}_t = \hat{R}_t \hat{R}_t^{\top} \in \mathbb{R}^{n\times n}$, $\g_t = \overline{\operatorname{vec}}(G_t^X)$ and $\x_t = \overline{\operatorname{vec}}(X_t)$, where $\overline{\operatorname{vec}}$ denotes the row-major vectorization of a given matrix. The Kronecker product $\otimes$ obeys the following properties.

\begin{lemma}[{\citealp[Lemma 3,4]{gupta2018shampoo}}]\label{propety}
For matrices $A, A', B, B'$ of appropriate dimensions and vectors $\x, \y$, $L \in \mathbb{R}^{m \times m}$, $R \in \mathbb{R}^{n \times n}$, $G \in \mathbb{R}^{m \times n}$, the following properties hold:
\begin{enumerate}
    \item $(A \otimes B)(A' \otimes B') = (AA') \otimes (BB').$
    \item $(A \otimes B)^{\top} = A^{\top} \otimes B^{\top}.$
    \item $A, B \succeq 0, \, (A \otimes B)^{-1} = A^{-1} \otimes B^{-1}.$
    \item $A \succeq A', \, B \succeq B', \, \text{then} \, A \otimes B \succeq A' \otimes B'.$
    \item $\operatorname{tr}(A \otimes B) = \operatorname{tr}(A)\operatorname{tr}(B).$
    \item $\overline{\operatorname{vec}}(\x\y^{\top}) = \x \otimes \y.$
    \item $(L \otimes R^{\top}) \overline{\operatorname{vec}}(G) = \overline{\operatorname{vec}}(LGR).$
\end{enumerate}
\end{lemma}
By defining $\overline{\g}_t = \sum_{i=1}^t \g_i$, we can rewrite the update in Algorithm \ref{shampoo} as
\begin{equation*}
    \x_{t+1} = -\eta \tilde{H}_t^{-1}\overline{\g}_t,
\end{equation*}
which is equal to 
\begin{equation*}
    \x_{t+1} = \argmin_{\x \in \R^{mn}}\eta \left\langle\overline{\g}_t,\x  \right\rangle + \frac{1}{2}\Norm{\x}_{\tilde{H}_t}^2.
\end{equation*}
As $\tilde{L}_t$ and $\tilde{R}_t$ are monotonically increasing with $t$, it is not hard to find that $\tilde{H}_t$ is also monotonically increasing with $t$. Thus, by using Lemma \ref{duchi1}, we have a similar inequality:
\begin{equation}\label{duchi-2}
    R(T) \leq \underbrace{\frac{1}{\eta} \Psi_T(\x^*)}_{R_D}
+ \underbrace{\frac{\eta}{2} \sum_{t=1}^T \|\g_t\|^2_{\tilde{H}_{t-1}^{-1}}}_{R_G},
\end{equation}
where $\Psi_T(\x^*) = \frac{1}{2}\left\langle \x^*, \tilde{H}_T \x^*\right\rangle$.

We first give the bound of $R_D$. 
\begin{align*}
    \Psi_T(\x^*) &= \frac{1}{2}\left\langle \x^*, \tilde{H}_T \x^*\right\rangle \leq \frac{1}{2}\Norm{\tilde{H}_T}\Norm{\x^*}^2.
\end{align*}

Then we introduce a lemma to give a guarantee about the norm of Kronecker product.
\begin{lemma}[{\citealp[Theorem 8]{lancaster1972norms}}]\label{lem4} For two matrices $A$ and $B$, the following equality holds
\begin{equation*}
    \Norm{A \otimes B} = \Norm{A} \Norm{B}.
\end{equation*}
\end{lemma}

We have $\Norm{\x^*} = \Norm{X^*}_F \leq D_{F}.$ According to Lemma \ref{lem4}, $\Norm{\tilde{H}_T} = \Norm{\tilde{L}_T^{1/4} \otimes \tilde{R}_T^{1/4}} = \Norm{\tilde{L}_T^{1/4}} \Norm{ \tilde{R}_T^{1/4}}$, then we need to give the bound of  $\Norm{\tilde{L}_T^{1/4}}$ and $\Norm{\tilde{R}_T^{1/4}}$, respectively. We first define $L_T = \sum_{t=1}^{T}G_t^X(G_t^X)^{\top}$ and $R_T = \sum_{t=1}^{T}(G_t^X)^{\top}G_t^X$. Using Lemma \ref{lem1}, it is not hard to verify that $\overline{L}_t +\epsilon I_{m\times m}\preceq L_t + \epsilon I_{m\times m}, \overline{R}_t +\epsilon I_{n\times n} \preceq R_t + \epsilon I_{n \times n}$.

Therefore, we have
\begin{equation}\label{eq:11}
\begin{aligned}
     \Norm{\tilde{L}_T^{1/4}}& = \Norm{(\overline{L}_T+\epsilon I_{m\times m}+ \rho_{1:T}^LI_{m \times m})^{1/4}} \\
    &\leq \Norm{(\overline{L}_T)^{1/4}+ (\epsilon I_{m \times m}+ \rho_{1:T}^LI_{m \times m})^{1/4}} \\
     &\leq \Norm{(\overline{L}_T)^{1/4}}+ \Norm{(\rho_{1:T}^LI_{m \times m}+\epsilon I_{m\times m})^{1/4}} \\
     &\leq \Norm{(\overline{L}_T)^{1/4}}+ (\rho_{1:T}^{L}+\epsilon)^{1/4}\\
     &\leq \operatorname{tr}(L_T^{1/4}) +  (\rho_{1:T}^L)^{1/4} + \epsilon^{1/4},
\end{aligned}    
\end{equation}
where the fourth inequality is due to the fact that $\operatorname{tr}(\cdot) \geq \Norm{\cdot}$ for positive semidefinite matrices, and the last inequality is due to the monotonicity of $\operatorname{tr}(\cdot)$.

We also have
\begin{align*}
     \Norm{\tilde{R}_T^{1/4}} &= \Norm{(\overline{R}_T + \epsilon I_{n \times n}+ \rho_{1:T}^RI_{n \times n})^{1/4}} \\
     &\leq \Norm{(\overline{R}_T)^{1/4}} +  \Norm{(\rho_{1:T}^RI_{n\times n} + \epsilon I_{n \times n} )^{1/4}} \\
     &\leq \operatorname{tr}((\overline{R}_T)^{1/4}) +  (\rho_{1:T}^R)^{1/4} + \epsilon^{1/4} \\
     &\leq \operatorname{tr}(R_T^{1/4}) +  (\rho_{1:T}^R)^{1/4} + \epsilon^{1/4} .
\end{align*}

Therefore, we can get
\begin{equation}\label{rd2}
    R_D = \frac{1}{\eta}\Psi_T(\x^*)\leq \frac{D_{F}^2}{2\eta}(\operatorname{tr}(L_T^{1/4}) + (\rho_{1:T}^L)^{1/4} + \epsilon^{1/4} )(\operatorname{tr}(R_T^{1/4}) + (\rho_{1:T}^R)^{1/4} + \epsilon^{1/4} ).
\end{equation}

To give the bound of $R_G$, we first introduce a lemma.
\begin{lemma}[{\citealp[Lemma 8]{gupta2018shampoo}}] If  $G^X_t \in \mathbb{R}^{m \times n}$  has rank at most  $r$, and  $\g_t = \overline{\operatorname{vec}}(G_t^X)$, then $\forall \epsilon \geq 0, \forall t$,
\begin{equation*}
    \epsilon I_{mn \times mn} + \frac{1}{r} \sum_{i=1}^t \g_i \g_i^\top \preceq 
\left( \epsilon I_{m \times m} + \sum_{i=1}^t G_i^X (G_i^X)^\top \right)^{1/2} 
\otimes 
\left( \epsilon I_{n \times n} + \sum_{i=1}^t (G_i^X)^\top (G_i^X) \right)^{1/2}.
\end{equation*}
\end{lemma}

Then we utilize a lemma from \citet{feinberg2024sketchy}.
\begin{lemma}[{\citealp[Lemma 14]{feinberg2024sketchy}}]\label{lemma:10} Let $V_t\Sigma_t^LV_t^{\top} = \overline{L}_{t-1} + G_t^X(G_t^X)^{\top}$ be the eigendecomposition of the un-deflated sketch. We assume $\text{rank}(\Sigma_t^L)=k, k \in [\tau-1, \tau-1+r]$. Write $V_t = [V_t^{*}, V_t^{\bot}]$, where $V_t^*$ contains the first $k$ columns of $V_t$. Similarly, for the right preconditioner, let $W_t\Sigma_t^R W_t^{\top} = \overline{R}_{t-1} + (G_t^X)^{\top}G_t^X$. Write $W_t = [W_t^{*}, W_t^{\bot}]$, where $W_t^*$ contains the first $k$ columns of $W_t$. Define $N_t^L =  V_t^{\bot} ( V_t^{\bot})^{\top}$ and $N_t^R =  W_t^{\bot} ( W_t^{\bot})^{\top}$, then we have
\begin{align*}
\widetilde{L}_t &\succeq \sum_{i=1}^{t} G_i^X (G_i^X)^\top + \sum_{i=1}^{t} \rho_i^L N_i^L + \epsilon I_{m\times m}= M_t^L, \\
\widetilde{R}_t &\succeq \sum_{i=1}^{t} (G_i^X)^\top G_i^X + \sum_{i=1}^{t} \rho_i^R N_i^R + \epsilon I_{n \times n } = M_t^R.
\end{align*}
\end{lemma}

According to Lemma \ref{lem1} and Lemma \ref{lemma:10}, we have $M_t^{L} \succeq \epsilon I_{m\times m} + \sum_{i=1}^t G_i^X (G_i^X)^\top$ and $M_t^{R} \succeq \epsilon I_{n \times n} + \sum_{i=1}^t (G_i^X)^{\top} G_i^X$. Using Lemma \ref{propety}, we can derive
 \begin{equation}\label{eq:12}
\begin{aligned}
I_{m \times m} \otimes \left( \epsilon I_{n\times n} + \sum_{i=1}^t (G_i^X)^\top G_i^X \right)
&\preceq I_{m\times m} \otimes M_t^R, \\
\left( \epsilon I_{m\times m} + \sum_{i=1}^t G_i^X (G_i^X)^\top \right) \otimes I_{n \times n}
&\preceq M_t^L \otimes I_{n\times n}.
\end{aligned}
\end{equation}

Therefore, we have
\begin{equation*}
(\epsilon I_{mn\times mn} + \frac{1}{r} \sum_{i=1}^t \g_i \g_i^\top )^{1/2}
\preceq \left(M_t^L\right)^{1/4} \otimes \left(M_t^R\right)^{1/4} 
\preceq \tilde{L}_t^{1/4} \otimes \tilde{R}_t^{1/4} = \tilde{H}_t.
\end{equation*}

Then we define $\widehat{H}_t = \left( r \epsilon I_{mn\times mn} + \sum_{i=1}^t \g_i \g_i^\top \right)^{1/2}$, and  have
\begin{equation*}
    \widehat{H}_t \preceq \sqrt{r}\tilde{H}_t.
\end{equation*}

We want to give the lower bound of $\hat{H}_{t-1}$. By defining $c_t = \frac{r\epsilon}{\Norm{\g_t}^2 + r\epsilon}$, we have
\begin{align*}
    \hat{H}_{t-1}^2 & = r \epsilon I_{mn\times mn} + \sum_{i=1}^{t-1} \g_{i} \g_{i}^\top\\
    & \succeq c_t (\Norm{\g_t}^2 I_{mn\times mn} + \sum_{i=1}^{t-1} \g_{i} \g_{i}^\top + r\epsilon I_{mn\times mn})\\
    & \succeq c_t (r\epsilon I_{mn\times mn} + \sum_{i=1}^t \g_{i} \g_{i}^\top)\\
    & = c_t \hat{H}_t^2.
\end{align*}

Define $A_1 = \min_{t\in[T]}\{\sqrt{c_t}\}$. We have $A_1 \hat{H}_t \preceq \hat{H}_{t-1} \preceq \sqrt{r}\tilde{H}_{t-1}$, which means $\frac{1}{\sqrt{r}}\tilde{H}_{t-1}^{-1} \preceq \hat{H}_{t-1}^{-1} \preceq \frac{1}{A_1}\hat{H}_t^{-1}$.

We can derive
\begin{align*}
    \sum_{t=1}^T \|\g_t\|^2_{\tilde{H}^{-1}_{t-1}} & =\sum_{t=1}^T \left\langle \g_t,\tilde{H}^{-1}_{t-1}\g_t
    \right\rangle  \leq \sqrt{r}\sum_{t=1}^T \left\langle \g_t,\hat{H}^{-1}_{t-1}\g_t
    \right\rangle \\
& \leq \frac{\sqrt{r}}{A_1}\sum_{t=1}^T \left\langle \g_t,\hat{H}^{-1}_{t}\g_t
    \right\rangle.
\end{align*}

Based on the proof of Theorem 5 in \citet{feinberg2024sketchy}, we obtain the following inequality.
\begin{equation*}
    \sum_{t=1}^T \left\langle \g_t,\hat{H}^{-1}_{t}\g_t
    \right\rangle \leq 2\operatorname{tr}(\hat{H}_T).
\end{equation*}

Therefore, we have
\begin{align*}
    \sum_{t=1}^T \|\g_t\|^2_{\tilde{H}^{-1}_{t-1}} \leq \frac{\sqrt{r}}{A_1}\sum_{t=1}^T \left\langle \g_t,\hat{H}^{-1}_{t}\g_t
    \right\rangle   \leq \frac{2}{A_1}\sqrt{r}\operatorname{tr}(\hat{H}_T).
\end{align*}

Then, we need to bound the term $\hat{H}_T$.
\begin{align*}
    \hat{H}_T^2 = (r\epsilon I_{mn\times mn} + \sum_{i=1}^T \g_i\g_i^{\top}) &\preceq r
\left( \epsilon I_{m \times m} + \sum_{i=1}^T G_i^X (G_i^X)^\top \right)^{1/2} \\
&\quad \otimes
\left( \epsilon I_{n \times n} + \sum_{i=1}^T (G_i^X)^\top G_i^X \right)^{1/2}\\
 &= r \left(\epsilon I_{m \times m} + L_T\right)^{1/2} \otimes \left(\epsilon I_{n \times n} +R_T\right)^{1/2},
\end{align*}
which means $\hat{H}_T \preceq \sqrt{r}\left(\epsilon I_{m \times m} + L_T\right)^{1/4} \otimes  \left(\epsilon I_{n \times n} +R_T\right)^{1/4}$.

By defining $C_3 = \frac{1}{A_1}$, we can derive the bound of $R_G$.
\begin{align*}
    R_G & = \frac{\eta}{2} \sum_{t=1}^T \|\g_t\|^2_{\tilde{H}^{-1}_{t-1}} \leq \eta C_3 \sqrt{r} \operatorname{tr}(\hat{H}_T)\\
    & \leq \eta C_3 r \operatorname{tr}(\left(\epsilon I_{m \times m} + L_T\right)^{1/4})\operatorname{tr}(\left(\epsilon I_{n \times n} +R_T\right)^{1/4}).
\end{align*}
By setting $\eta = \frac{D_{F}}{\sqrt{r}}$, the final regret bound is 
\begin{equation}\label{eq0}
    \begin{aligned}
        &\sum_{t=1}^T f_t(\x_t) - f_t(\x^*)\\ 
         \leq& \eta C_3r \operatorname{tr}\left(\left(\epsilon I_{m \times m} + L_T\right)^{1/4}\right) \operatorname{tr}\left(\left(\epsilon I_{n \times n} +R_T\right)^{1/4}\right) \\
 &\qquad + \frac{D_{F}^2}{2\eta}\left(\operatorname{tr}(L_T^{1/4}) + (\rho_{1:T}^L)^{1/4} +\epsilon^{1/4}\right)\left(\operatorname{tr}(R_T^{1/4}) + (\rho_{1:T}^R)^{1/4}+\epsilon^{1/4}\right)\\
    \leq& \sqrt{r}D_{F}C_3\operatorname{tr}\left(\left(\epsilon I_{m \times m} + L_T\right)^{1/4}\right) \operatorname{tr}\left(\left(\epsilon I_{n \times n} +R_T\right)^{1/4}\right)\\
 &\qquad + \frac{\sqrt{r}D_{F}}{2}\left(\operatorname{tr}(L_T^{1/4}) + (\rho_{1:T}^L)^{1/4}+\epsilon^{1/4}\right)\left(\operatorname{tr}(R_T^{1/4}) + (\rho_{1:T}^R)^{1/4} +\epsilon^{1/4}\right)\\
     = &O\left(D_F\sqrt{r}(\operatorname{tr}(L_T^{1/4}) + (\rho_{1:T}^L)^{1/4})(\operatorname{tr}(R_T^{1/4}) + (\rho_{1:T}^R)^{1/4})\right).
    \end{aligned}
\end{equation}

\subsection{Proof of Theorem \ref{thm:jmlr1}}
The proof of Theorem \ref{thm:jmlr1} follows the same structure as 
that of Theorem \ref{thm3}, with a refined analysis of the 
preconditioning matrix.

Define 
\begin{equation*}
    \tilde{H}_t = \tilde{L}_t^{1/2} \otimes I_{n\times n} 
    \in \R^{mn \times mn}.
\end{equation*}
Applying Lemma \ref{duchi1} yields the decomposition
\begin{equation}\label{duchi-3}
    R(T) \leq \underbrace{\frac{1}{\eta} \Psi_T(\x^*)}_{R_D}
+ \underbrace{\frac{\eta}{2} \sum_{t=1}^T \|\g_t\|^2_{\tilde{H}_{t-1}^{-1}}}_{R_G},
\end{equation}
where $\Psi_T(\x^*) = \frac{1}{2}\left\langle \x^*, \tilde{H}_T \x^*\right\rangle$.

We first bound $R_D$. By using the Cauchy-Schwarz inequality,
\begin{equation*}
    \Psi_T(\x^*) 
    = \tfrac{1}{2}\left\langle \x^*, \tilde{H}_T \x^*\right\rangle 
    \leq \tfrac{1}{2}\Norm{\tilde{H}_T}_2 \Norm{\x^*}^2.
\end{equation*}
Invoking Lemma \ref{lem4} together with \eqref{eq:11} gives
\begin{equation*}
    \Psi_T(\x^*) 
    \leq \tfrac{1}{2}\Norm{\tilde{L}_T^{1/2}}_2 \Norm{I_{n\times n}}_2 
         \Norm{\x^*}^2
    \leq \frac{D_F^2}{2}\left(\operatorname{tr}(L_T^{1/2}) 
         +\sqrt{\rho_{1:T}^L} + \epsilon\right).
\end{equation*}

We next turn to $R_G$. Using the identity 
$\operatorname{vec}(A)^\top (I \otimes B) \operatorname{vec}(A) 
= \operatorname{tr}(A^\top B A)$, we have
\begin{equation*}
    \sum_{t=1}^T \Norm{\g_t}^2_{\tilde{H}^{-1}_{t-1}} 
    = \sum_{t=1}^T \left\langle \g_t, \tilde{H}^{-1}_{t-1} \g_t \right\rangle 
    = \sum_{t=1}^T \tr\left((G_t^X)^\top \tilde{L}_{t-1}^{-1/2} G_t^X \right).
\end{equation*}
We now establish a lower bound on $L_{t-1}$ in terms of $L_t$. 
Define $c_t' = \frac{\epsilon}{\Norm{G_t^X}_2^2 + \epsilon}$. 
Since $c_t' \Norm{G_t^X}_2^2 I_{m\times m} \succeq c_t' G_t^X (G_t^X)^\top$ 
and $c_t' \leq 1$, we obtain
\begin{align*}
    \epsilon I_{m\times m} 
      + \sum_{i=1}^{t-1} G^X_{i} (G^X_{i})^\top &\succeq c_t' \left( \Norm{G_t^X}_2^2 I_{m\times m} 
      + \sum_{i=1}^{t-1} G^X_{i} (G^X_{i})^\top 
      + \epsilon I_{m\times m} \right) \\
    &\succeq c_t' \left( \epsilon I_{m\times m} 
      + \sum_{i=1}^{t} G^X_{i} (G^X_{i})^\top \right).
\end{align*}

Define $A_4 = \max_{t \in [T]} \{1/c_t'\}$. Combining the above 
inequality with \eqref{eq:12} yields
\begin{equation}\label{eq:15}
    \begin{aligned}
    \sum_{t=1}^T \tr\left((G_t^X)^\top \tilde{L}_{t-1}^{-1/2} G_t^X \right) &\leq \sum_{t=1}^T \tr\left((G_t^X)^\top (\epsilon I_{m\times m} + L_{t-1})^{-1/2} G_t^X \right)\\
    &\leq A_4 \sum_{t=1}^T \tr\left((G_t^X)^\top(\epsilon I_{m\times m} + L_{t})^{-1/2} G_t^X \right),
\end{aligned}
\end{equation}
where the first inequality follows from $\tilde{L}_{t-1} \succeq\epsilon I_{m\times m} +  L_{t-1}$ and the fact that $\tr(A^{\top}BA) \leq\tr(A^{\top}CA)$ if $C \succeq B$. The right-hand side is bounded by the following lemma.

\begin{lemma}[{\citealp[Proof of Theorem 1]{an2025asgo}}]\label{lem:10}
With the notation of Algorithm \ref{alg:sasgo},
\begin{equation*}
    \sum_{t=1}^T \tr\left((G_t^X)^\top (\epsilon I_{m\times m} + L_{t})^{-1/2} G_t^X \right) 
    \leq 2 \operatorname{tr}(L_T^{1/2}).
\end{equation*}
\end{lemma}

Substituting Lemma \ref{lem:10} into \eqref{duchi-3} and setting
$\eta = \frac{D_F}{\sqrt{2}}$, we obtain the following guarantee
for S-ASGO under the primal-dual framework
\begin{align*}
    R(T) 
    &\leq \frac{D_F^2}{2\eta} \left(\operatorname{tr}(L_T^{1/2}) +\sqrt{\rho_{1:T}^L}+\epsilon \right) 
      + \eta A_4 \operatorname{tr}(L_T^{1/2}) \\
    &= O\!\left(D_F \left(\operatorname{tr}(L_T^{1/2}) +\sqrt{\rho_{1:T}^L}\right)\right).
\end{align*}
This completes the proof of Theorem \ref{thm:jmlr1}.

\subsection{Proof of Theorem \ref{thm:jmlr1md}}
We now turn to the mirror descent framework. Applying the standard 
mirror descent analysis yields
\begin{equation}\label{eq:13}
R(T) \leq \underbrace{\frac{1}{2\eta} \sum_{t=1}^T 
    \left(\Norm{\x_t - \x^*}^2_{\tilde{H}_t} 
    - \Norm{\x_{t+1} - \x^*}^2_{\tilde{H}_t}\right)}_{R_D} 
    + \underbrace{\frac{\eta}{2} \sum_{t=1}^T 
    \Norm{\g_t}^2_{\tilde{H}_t^{-1}}}_{R_G}.
\end{equation}

We first derive the bound of $R_D$.
\begin{equation}\label{eq:14}
\begin{aligned}
    &\sum_{t=1}^T \left( \Norm{\x_t - \x^*}^2_{\tilde{H}_t} 
        - \Norm{\x_{t+1} - \x^*}^2_{\tilde{H}_t} \right) \\
    =\;& \sum_{t=1}^T \Norm{\x_t - \x^*}^2_{\tilde{H}_t - \tilde{H}_{t-1}} 
        + \Norm{\x_1 - \x^*}^2_{\tilde{H}_0} \\
    =\;& \sum_{t=1}^T \operatorname{tr}\!\left( 
        (X_t - X^*)^\top 
        (\tilde{L}_t^{1/2} - \tilde{L}_{t-1}^{1/2}) 
        (X_t - X^*) \right) 
        + \Norm{\tilde{H}_0}_2 \Norm{\x_1 - \x^*}^2 \\
    \leq\;& \sum_{t=1}^T \operatorname{tr}\!\left( 
        (X_t - X^*)(X_t - X^*)^\top 
        (\tilde{L}_t^{1/2} - \tilde{L}_{t-1}^{1/2}) \right) 
        + \epsilon D_F^2 \\
    \leq\;& \sum_{t=1}^T \Norm{X_t - X^*}_2^2 
        \operatorname{tr}\!\left( \tilde{L}_t^{1/2} - \tilde{L}_{t-1}^{1/2} \right) 
        + \epsilon D_F^2 \\
    \leq\;& D_2^2 \operatorname{tr}(\tilde{L}_T^{1/2}) + \epsilon D_F^2
    \leq D_2^2 \left( \operatorname{tr}(L_T^{1/2}) + m\sqrt{\rho_{1:T}^L} \right)
        + \epsilon D_F^2.
\end{aligned}
\end{equation}
where the first inequality is due to $\operatorname{tr}(AB) = \operatorname{tr}(BA)$ and $\Norm{\tilde{H}_0}_2\Norm{\x_1 - \x^*}^2 \leq \epsilon D_F^2$, and the second inequality is due to $\operatorname{tr}(AB) \leq \Norm{A}_2\operatorname{tr}(B)$ for positive semi-definite matrices.

Combining \eqref{eq:15}, \eqref{eq:13} and \eqref{eq:14}, and setting 
$\eta = D_2/\sqrt{2}$, we obtain the following guarantee for S-ASGO under the mirror  descent framework
\begin{align*}
    R(T) 
    &\leq \frac{1}{2\eta}\left( D_2^2 \left( \operatorname{tr}(L_T^{1/2}) 
        + m\sqrt{\rho_{1:T}^L} \right) + \epsilon D_F^2 \right) 
        + \frac{\eta}{2} \operatorname{tr}(L_T^{1/2}) \\
    &= O\!\left( D_2 \left( \operatorname{tr}(L_T^{1/2}) 
        + m\sqrt{\rho_{1:T}^L} \right) \right),
\end{align*}
which completes the proof of Theorem \ref{thm:jmlr1md}.

\subsection{Proof of Theorem \ref{thm:jmlr2}}

In this section, we present the proof for smooth loss functions. We introduce a lemma from the proof in \citet{an2025asgo}.
\begin{lemma}
    Under Assumptions \ref{smooth2} and \ref{noise2}, it holds that
    \begin{equation*}
        \E\left[\Norm{\left(\sum_{t=1}^T \nabla f(X_t) \nabla f(X_t)^\top\right)^{1/2}}_{*}\right] \leq \sqrt{\Norm{\L}_{*}\sum_{t=1}^T 2\left(f(X_t) - f(X^*)\right)}
    \end{equation*}
    \begin{equation*}
        \E\left[\Norm{\left(\sum_{t=1}^T N_t N_t^\top\right)^{1/2}}_{*}\right] \leq \sqrt{\frac{T}{M}}\Norm{V}_{*}.
    \end{equation*}
\end{lemma}

\paragraph{Primal-dual framework.} We first present the proof for the primal-dual framework. We have
\begin{align*}
     &\frac{1}{T}\sum_{t=1}^T\E\left[f(X_t)-f(X^*)\right]\\
     \leq& \frac{2 A_4 D_F}{T}\operatorname{tr}(L_T^{1/2}) + \frac{\sqrt{2}D_F}{2T}\left(\sqrt{\rho_{1:T}^L}+\epsilon\right)\\
     =& \frac{2 A_4 D_F}{T}\Norm{\left(\sum_{t=1}^T G_t^X (G_t^X)^{\top}\right)^{1/2}}_{*} + \frac{\sqrt{2}D_F}{2T}\left(\sqrt{\rho_{1:T}^L}+\epsilon\right)\\
     \leq& \frac{2\sqrt{2} A_4 D_F}{T}\Norm{\left(\sum_{t=1}^T \nabla f(X_t) (\nabla f(X_t))^{\top} + N_t N_t^\top\right)^{1/2}}_{*} + \frac{\sqrt{2}D_F}{2T}\left(\sqrt{\rho_{1:T}^L}+\epsilon\right)\\
     \leq& \frac{2\sqrt{2} A_4 D_F}{T}\Norm{\left(\sum_{t=1}^T \nabla f(X_t) (\nabla f(X_t))^{\top}\right)^{1/2}}_{*} + \frac{2\sqrt{2} A_4 D_F}{T}\Norm{\left(\sum_{t=1}^T N_t N_t^\top\right)^{1/2}}_{*} \\
     &+ \frac{\sqrt{2}D_F}{2T}\left(\sqrt{\rho_{1:T}^L}+\epsilon\right)\\
     \leq& \frac{4 A_4 D_F}{T}\sqrt{\Norm{\L}_{*}\sum_{t=1}^T\E\left[f(X_t)-f(X^*)\right]} + \frac{2\sqrt{2} A_4 D_F\Norm{V}_{*}}{\sqrt{MT}} + \frac{\sqrt{2}D_F}{2T}\left(\sqrt{\rho_{1:T}^L}+\epsilon\right).
\end{align*}

If we denote $a = \sqrt{\frac{1}{T} \sum_{t=1}^T\E\left[f(X_t)-f(X^*)\right]}$, we can rewrite it as
\begin{equation*}
    a^2\leq ba+c,
\end{equation*}
where $b = \frac{4A_4D_F\sqrt{\Norm{\L}_{*}}}{\sqrt{T}}, c=\frac{2\sqrt{2} A_4 D_F\Norm{V}_{*}}{\sqrt{MT}} + \frac{D_F(\sqrt{\rho_{1:T}^L}+\epsilon)}{T}$. By using $a^2\leq 2b^2+2c$, we can derive
\begin{align*}
    \frac{1}{T}\sum_{t=1}^T\E\left[f(X_t)-f(X^*)\right] \leq \frac{16A_4^2D_F^2\Norm{\L}_{*}}{T} + \frac{4\sqrt{2}A_4D_F\Norm{V}_*}{\sqrt{MT}} + \frac{2\sqrt{2}D_F\left(\sqrt{\rho_{1:T}^L}+\epsilon\right)}{2T}.
\end{align*}

\paragraph{Mirror-descent framework.}  It is not hard to verify that, for S-ASGO under the mirror descent framework, by using Theorem \ref{thm:p4}, we can obtain
\begin{align*}
     &\frac{1}{T}\sum_{t=1}^T\E\left[f(X_t)-f(X^*)\right]\\
     \leq& \frac{\sqrt{2}D_2}{T}\operatorname{tr}(L_T^{1/2}) + \frac{\sqrt{2}mD_2}{2T}\sqrt{\rho_{1:T}^L}+ \frac{\sqrt{2}\epsilon D_F^2}{2TD_2}\\
     \leq& \frac{2 D_2}{T}\Norm{\left(\sum_{t=1}^T \nabla f(X_t) (\nabla f(X_t))^{\top}\right)^{1/2}}_{*} + \frac{\sqrt{2} D_2}{T}\Norm{\left(\sum_{t=1}^T N_t N_t^\top\right)^{1/2}}_{*} \\
     &+ \frac{\sqrt{2}D_2m}{2T}\sqrt{\rho_{1:T}^L}+\frac{\sqrt{2}\epsilon D_F^2}{2TD_2}\\
     \leq& \frac{2D_2}{T}\sqrt{\Norm{\L}_{*}\sum_{t=1}^T\E\left[f(X_t)-f(X^*)\right]} + \frac{\sqrt{2}D_2\Norm{V}_{*}}{\sqrt{MT}} + \frac{\sqrt{2}mD_2}{2T}\sqrt{\rho_{1:T}^L}+\frac{\sqrt{2}\epsilon D_F^2}{2TD_2}.
\end{align*}

If we denote $a = \sqrt{\frac{1}{T} \sum_{t=1}^T\E\left[f(X_t)-f(X^*)\right]}$, we can rewrite it as
\begin{equation*}
    a^2\leq ba+c,
\end{equation*}
where $b = \frac{2D_2\sqrt{\Norm{\L}_{*}}}{\sqrt{T}}, c=\frac{\sqrt{2} D_2\Norm{V}_{*}}{\sqrt{MT}} + \frac{\sqrt{2}mD_2\sqrt{\rho_{1:T}^L}}{T}+ \frac{\sqrt{2}\epsilon D_F^2}{2TD_2}$. By using $a^2\leq 2b^2+2c$, we can derive

    \begin{equation*}
        \frac{1}{T}\sum_{t=1}^T \E\left[f(X_t)- f(X^*)\right] \leq O\left(\frac{D_2^2 \Norm{\L}_{*}}{T} + \frac{D_2 \Norm{V}_{*}}{\sqrt{MT}}+ \frac{mD_2\sqrt{\rho_{1:T}^L}}{T}\right).
    \end{equation*}

\section{Experiments}\label{section7}
In this section, we assess the performance of the proposed methods via numerical experiments, ranging from online classification and image classification to the pretraining of language models. All experiments are conducted on a server equipped with $4$ NVIDIA RTX 4090 GPUs and an Intel Xeon Platinum 8358P CPU (2.60 GHz).

\begin{figure*}[t]
	\begin{center}
		\subfigure{
			\includegraphics[width=0.29\textwidth]{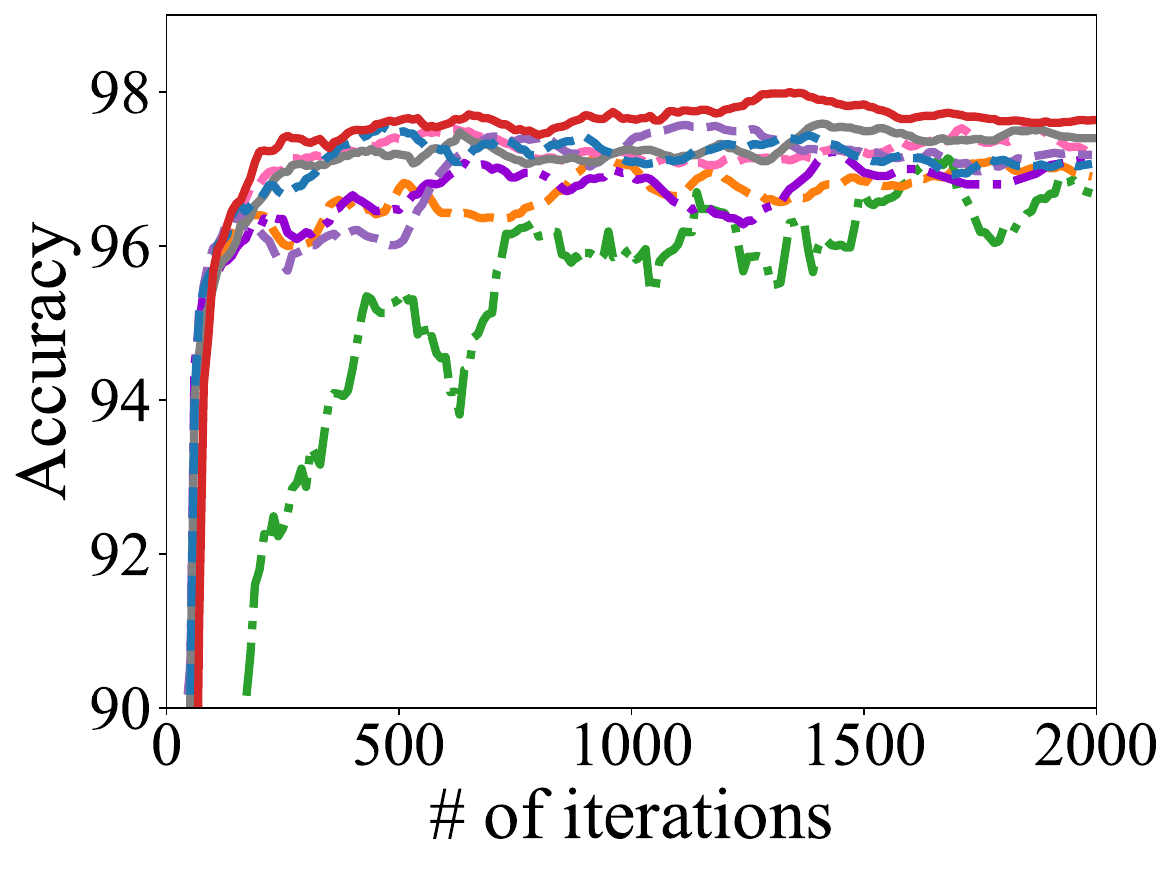}}
   \setcounter{subfigure}{0}
   \subfigure{
			\includegraphics[width=0.29\textwidth]{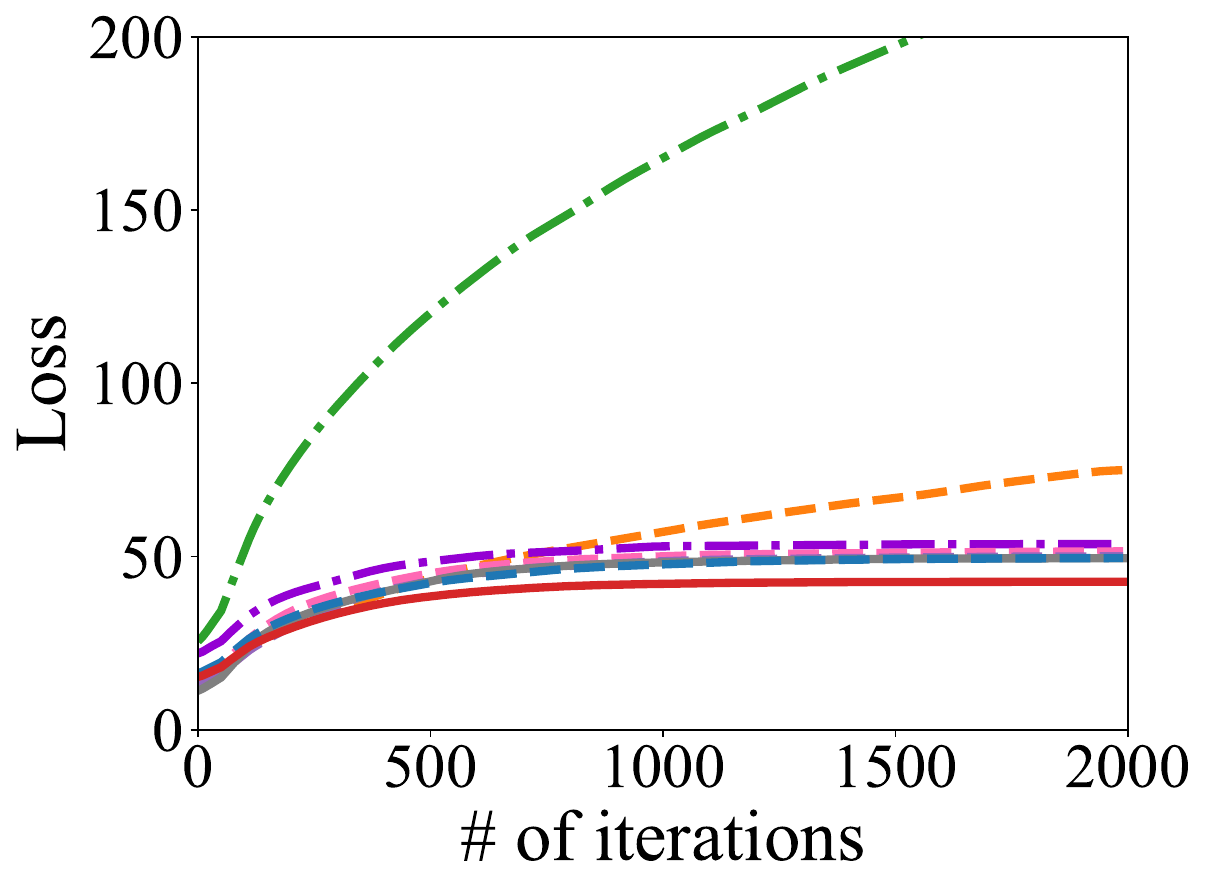}}
   \subfigure{
			\includegraphics[width=0.29\textwidth]{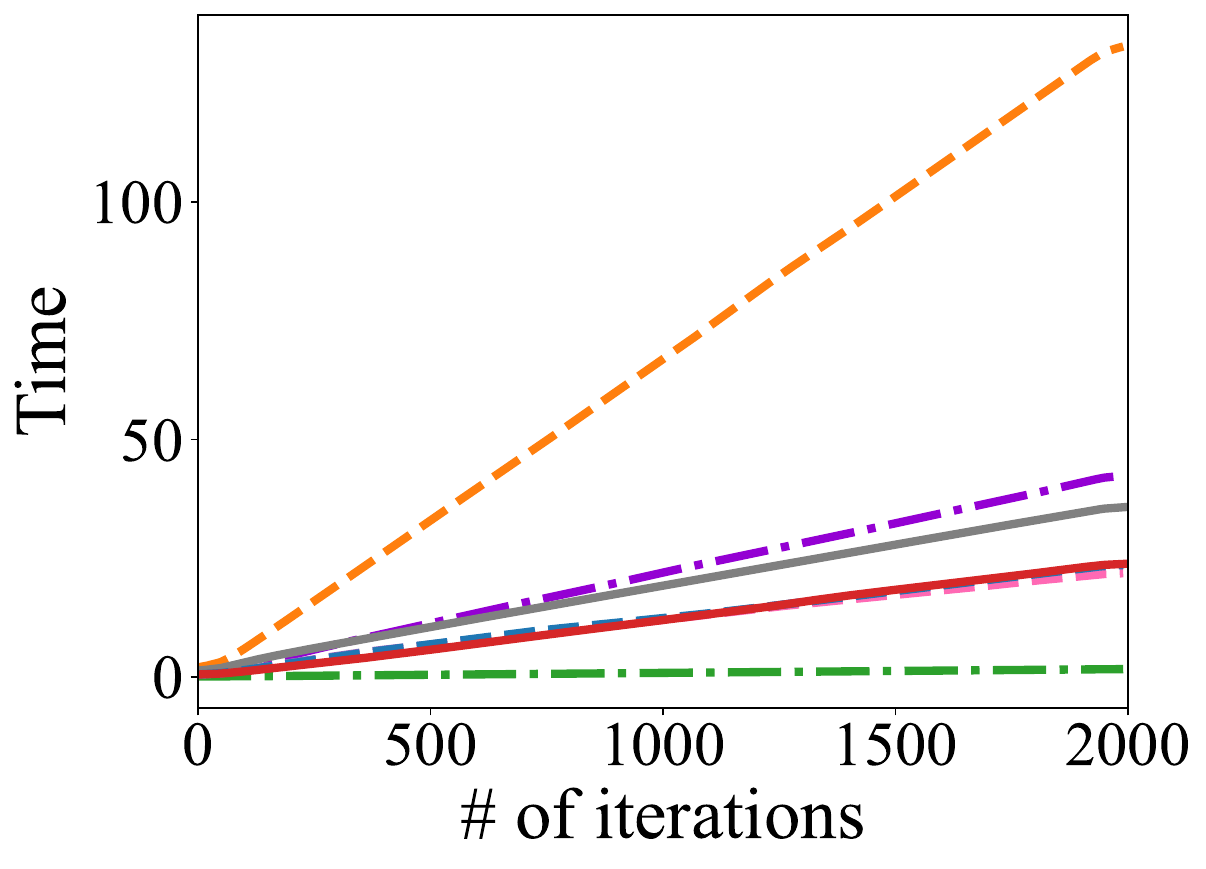}
		}
		\subfigure{
			\includegraphics[width=0.95\textwidth]{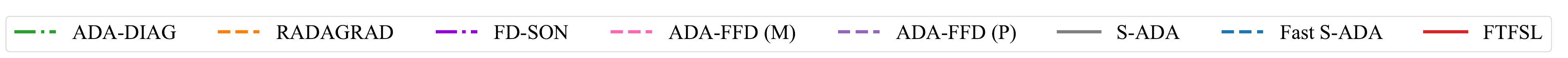}
		}
		\caption{Results for the Gisette dataset.}
		\label{fig:1}
	\end{center}
\end{figure*}
\begin{figure*}[th]
	\begin{center}
		\subfigure{
			\includegraphics[width=0.29\textwidth]{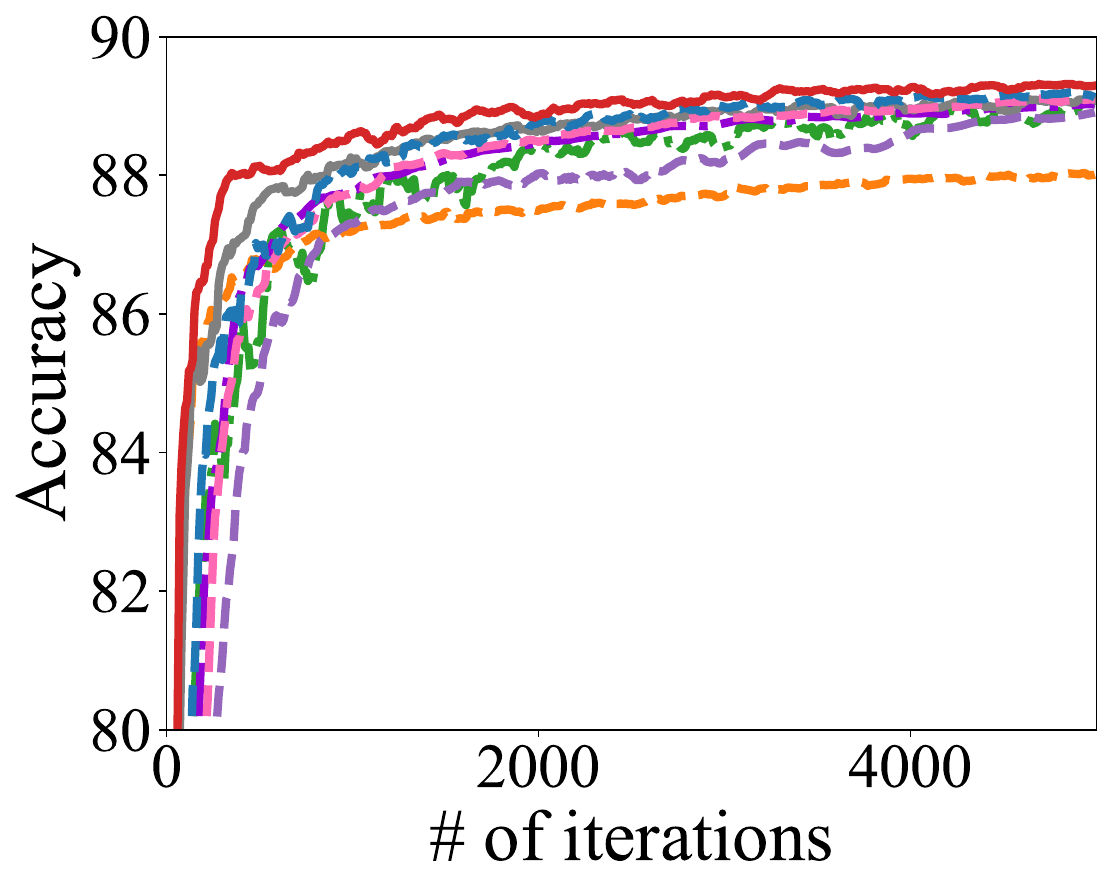}}
   \setcounter{subfigure}{0}
   \subfigure{
			\includegraphics[width=0.29\textwidth]{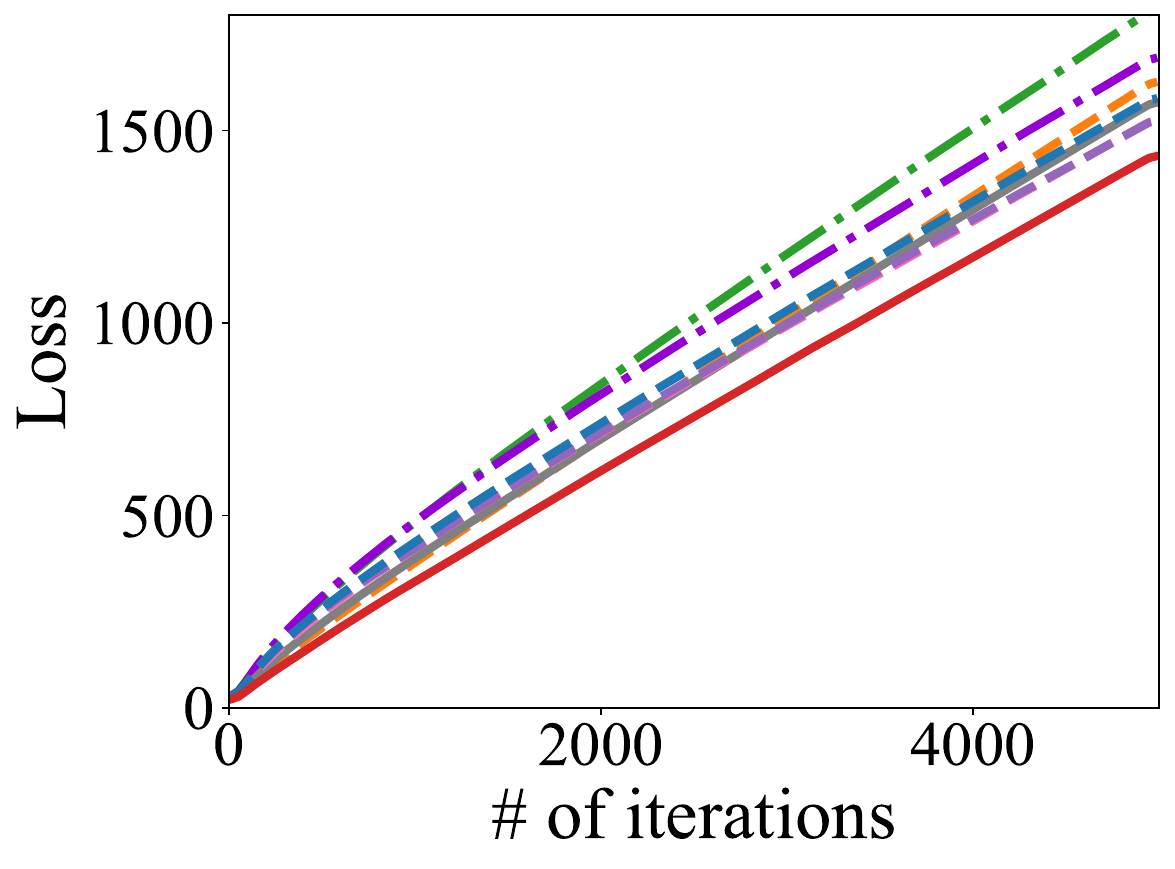}}
   \subfigure{
			\includegraphics[width=0.29\textwidth]{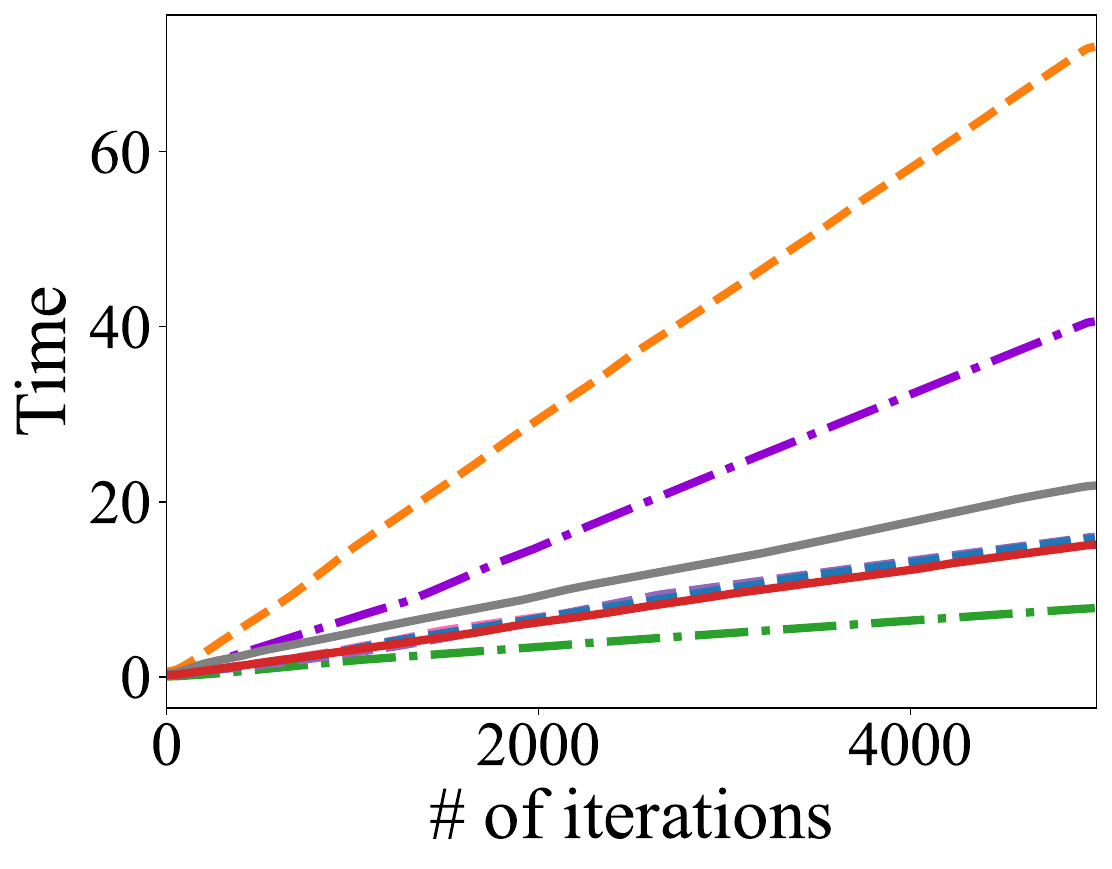}
		}
		\subfigure{
			\includegraphics[width=0.95\textwidth]{./exp/exp1/legend.pdf}
		}
		\caption{Results for the Epsilon dataset.}
		\label{fig:2}
	\end{center}
\end{figure*}

\subsection{Online Classification}
First, we perform online classification to evaluate the performance of our methods with two real-world datasets from the LIBSVM \citep{chang2011libsvm} repository: Gisette and Epsilon, which are high-dimensional and dense. In particular, the Gisette dataset contains $6000$ training samples and $1000$ testing samples, each with $5000$ features. The Epsilon dataset consists of $400,000$ training samples and $100,000$ testing samples,  each with $2000$ features.  

\paragraph{Experiment Setup.} In each round $t \in [T]$,  a batch of  data $\left\{\left(\mathbf{w}_{t, 1}, y_{t, 1}\right), \ldots,\left(\mathbf{w}_{t, n}, y_{t, n}\right)\right\}$ arrives, where $(\w_{t,i},y_{t,i}) \in [-1,1]^{d} \times\{-1,1\}, i=1, \ldots, n$.  The online learner aims to predict a linear model $\x_t$ and suffers  the hinge loss $f_t(\x_t) = \frac{1}{n}\sum_{i=1}^n \max\{0,1-y_{t,i}\x_t^{\top}\w_{t,i}\}$. For the Gisette dataset, we set the batch size $n = 32$, the sketching size $\tau=50$ to be $1\%$ of the original dimensionality, and $T=2000$. For the Epsilon dataset, we set the batch size $n=128$, $\tau=20$  and $T=5000$. We compare our methods with ADA-DIAG \citep{duchi2011adaptive}, RADAGRAD \citep{krummenacher2016scalable}, FD-SON \citep{luo2016efficient}, ADA-FFD under two frameworks \citep{wan2021efficient} and  S-ADA \citep{feinberg2024sketchy}.

\paragraph{Experiment Results.} Following \citet{duchi2011adaptive}, we adopt the test accuracy to compare different methods. To better demonstrate the improvements of our methods, we additionally plot the training loss and runtime of various methods.   From Figure \ref{fig:1} and Figure \ref{fig:2}, we observe that FTFSL outperforms all other methods in both loss and testing accuracy, aligning with its superior regret bound. Moreover, FTFSL and Fast S-ADA exhibit significantly lower runtimes compared to S-ADA, owing to their superior time complexities.

\subsection{Image Classification}\label{sec:image-exp}

In this section, we conduct numerical experiments on multi-class image classification tasks to evaluate the performance of our proposed methods. We compare FTFSL, FTSL-Shampoo, S-ASGO under the mirror descent framework (S-ASGO (M)) and S-ASGO under the primal-dual framework (S-ASGO (P)) with several baseline methods, including ADA-DIAG \citep{duchi2011adaptive}, ADA-FFD under primal-dual framework \citep{wan2021efficient},  S-ADA \citep{feinberg2024sketchy}, Shampoo \citep{gupta2018shampoo}, S-Shampoo \citep{feinberg2024sketchy} and ASGO \citep{an2025asgo}. The experiments involve training ResNet18 and ResNet34 models \citep{Resnet18} on the CIFAR-10 and CIFAR-100 datasets \citep{Krizhevsky2009Cifar10}, respectively, for $200$ epochs with a batch size of $128$.

A direct implementation of FD requires a full SVD of the sketching matrix at every round, which becomes the dominant computational bottleneck in high-dimensional regimes. SVD remains notoriously difficult to parallelize on modern accelerators, and its wall-clock cost in practice is often orders of magnitude higher than its theoretical $O(\tau^2 d)$ complexity would suggest. 
To address this, ASGO \citep{an2025asgo} circumvents the SVD by computing the preconditioner $M^{-1/2}$ directly through Newton--Schulz iterations or the Polar Express \citep{amsel2026the}, albeit at the cost of additional memory for the intermediate iterates. However, this trick cannot be extended to our method, since our algorithm requires both the eigenvalues and the eigenvectors of the matrix at every round. To render our methods practical at scale, we instead adopt the ARO scheme \citep{gong2026aronewlensmatrix}, which computes the eigenvectors of the matrix and then recovers the eigenvalues from them. This substantially reduces the per-iteration cost while preserving the empirical performance of our methods.

Consider the problem of computing the leading $\tau$ left singular vectors of a matrix 
$A \in \R^{p \times \tau}$ with $p \gg \tau$, given a previous 
approximation $\hat{U}^{\mathrm{prev}} \in \R^{p \times \tau}$ from a 
closely related iterate. Following 
\citet{gong2026aronewlensmatrix}, we approximate the leading left 
singular subspace by
\begin{equation}\label{eq:approx-svd}
    \hat{U} = \mathrm{QR}\!\left( A \cdot 
        f_{\mathrm{RN}}\!\bigl( (\hat{U}^{\mathrm{prev}})^\top A \bigr)^{\!\top} 
    \right),
\end{equation}
where $\mathrm{QR}(\cdot)$ denotes the orthonormal factor of the thin QR decomposition, and  $f_{\mathrm{RN}}(\cdot) : \R^{\tau\times \tau} \to \R^{\tau \times \tau}$ 
is the row-normalization function defined as 
\begin{equation}\label{eq:row-norm}
    f_{\mathrm{RN}}(X) = \sqrt{\tau} \, Q(X)^{-1} X, 
    \qquad 
    Q(X) = \mathrm{Diag}\bigl( \Norm{X_{1,:}}, \ldots, \Norm{X_{\tau,:}} \bigr).
\end{equation}
After obtaining $\hat{U}$, we rotate $A$ into this approximate
left singular subspace and compute
\begin{equation}
    \hat{X} = \hat{U}^\top A .
\end{equation}
Following the same row-norm interpretation as in ARO, the singular values
are estimated by
\begin{equation}
    \hat{\sigma}_i = \Norm{\hat{X}_{i,:}}, \qquad i=1,\ldots,\tau .
\end{equation}

Two ingredients are central to the efficiency of this scheme. First, 
the warm start $\hat{U}^{\mathrm{prev}}$ allows a single 
power-iteration step to suffice whenever $A$ varies smoothly across 
consecutive invocations of the procedure. Second, the QR 
factorization is realized through the Shifted Cholesky-QR (SCQR) \citep{fukaya2020shifted}
variant: for an input matrix $C \in \R^{p \times \tau}$, one forms 
the regularized Gram matrix $P = C^\top C + \lambda I_{\tau\times \tau}$, computes 
its Cholesky factorization $P = L L^\top$, and obtains the 
orthonormal basis as $Q = C L^{-1}$.


\paragraph{Application to FD.} We instantiate the above procedure 
with $A = B_t$ and $\hat{U}^{\mathrm{prev}} = \hat{U}_{t-1}$, where 
$\hat{U}_{t-1}$ denotes the approximate left singular subspace 
maintained from the previous round. The output $\hat{U}_t$ serves as 
an approximation of the leading $\tau$ left singular vectors of 
$B_t$ and the FD truncation step proceeds on the recovered singular values $\hat{\Sigma}_t$. Since 
$\hat{U}_t \in \R^{d \times \tau}$ has the same shape as $B_t$, the 
additional storage incurred by the procedure is of the same order 
as that of the sketching matrix itself, and thus does not affect 
the overall space complexity of FD.

\paragraph{Experimental Setup.} For sketching-based methods, we set the sketching size $\tau$ according to the dimensionality of the gradient,
\begin{equation*}
    \tau = \min\{\lceil d \times \omega \rceil, 50\},
\end{equation*}
where $d$ denotes the total number of parameters in the layer and $\omega$ is the sketching ratio. For methods that exploit the matrix structure, which are more memory-efficient, we  set $\tau = \lceil \omega \times d_i \rceil$ along each dimension, where $d_i$ is the size of the $i$-th dimension of the gradient. We use a sketching ratio of $\omega = 0.1$ in all experiments. For a fair comparison, we disable momentum and apply only learning rate decay. We use the original version of ASGO for this task.

\begin{wrapfigure}[13]{r}{0.41\textwidth} 
  \centering
  \includegraphics[width=0.3\textwidth]{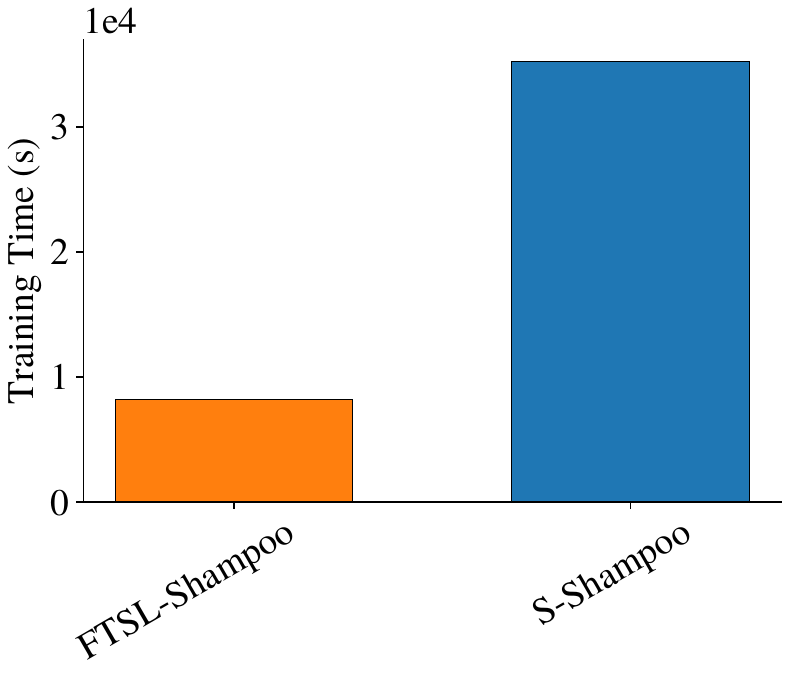}
  \caption{Running time of two types of SVD implementation.}
  \label{fig:svd-time}
\end{wrapfigure}

\paragraph{Experimental Results.} 
We first examine the wall-clock benefit of the approximate SVD technique. Figure~\ref{fig:svd-time} reports the training time during ResNet18 training on CIFAR-10, where the SVD of S-Shampoo is computed via the original PyTorch implementation. Across all training iterations, our approximation delivers a substantial reduction in the per-iteration cost.

\begin{figure*}[t!]
	\begin{center}
			\includegraphics[width=0.3\textwidth]{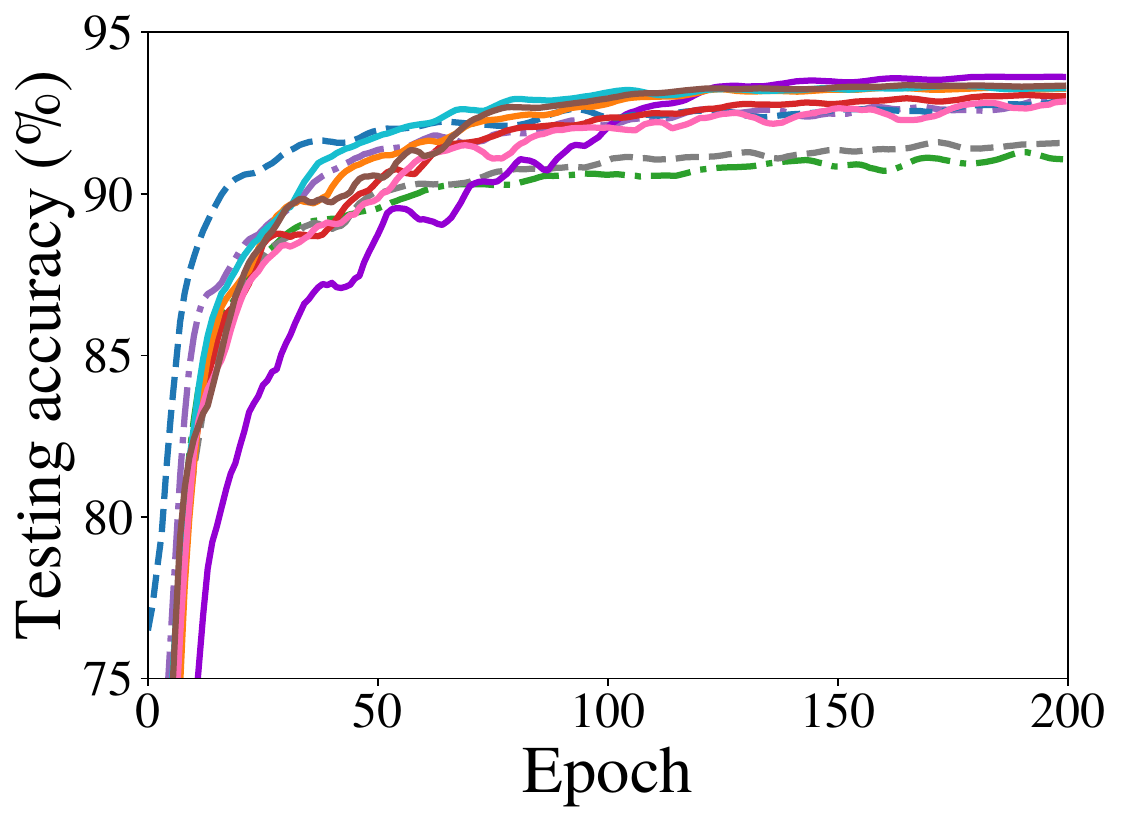}
            \includegraphics[width=0.3\textwidth]{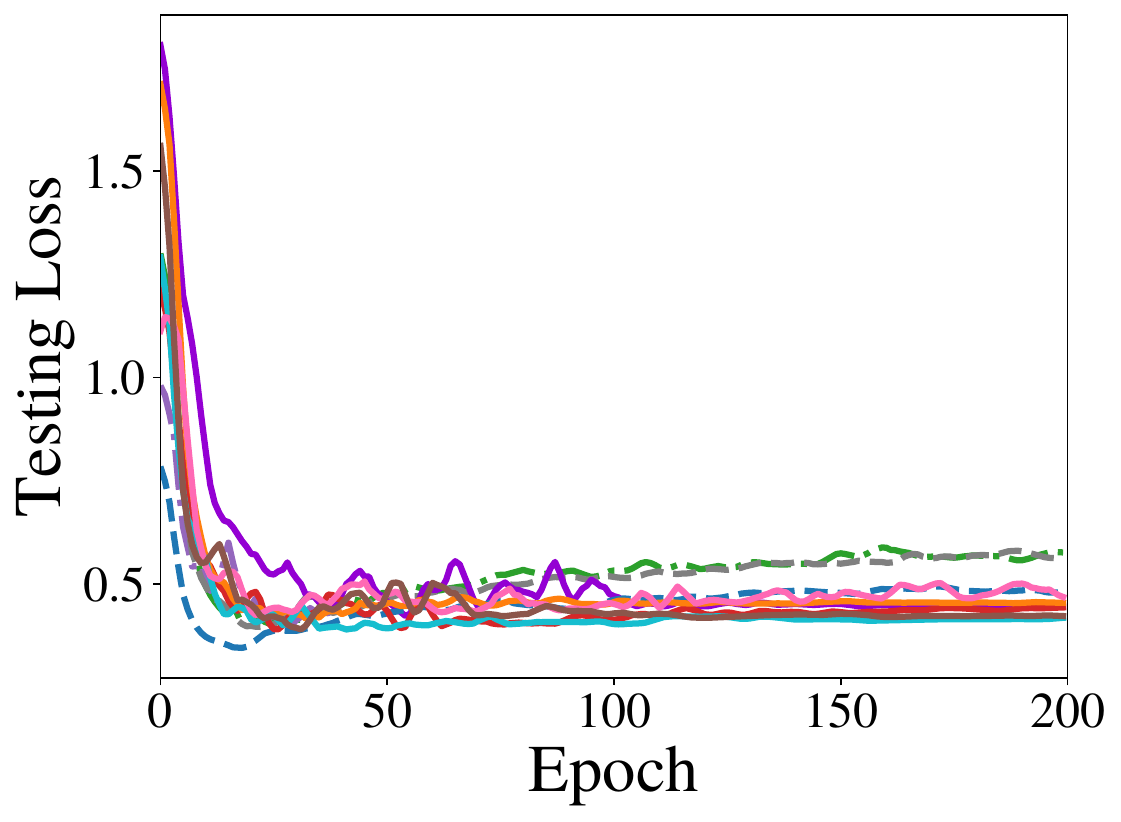}
			\includegraphics[width=0.3\textwidth]{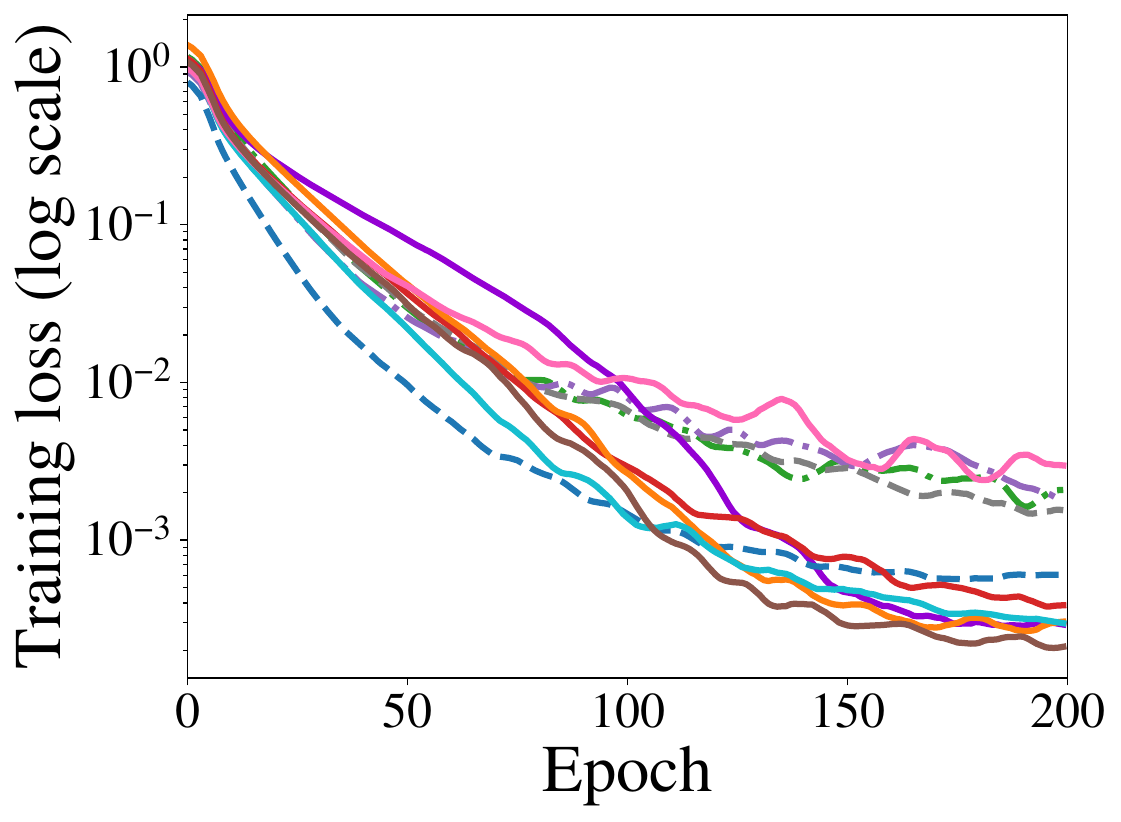}
		\subfigure{
			\includegraphics[width=0.98\textwidth]{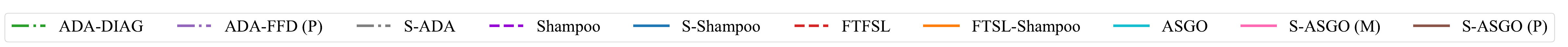}
		}
		\caption{Results for the CIFAR-10 dataset.}
		\label{fig:5}
	\end{center}
\end{figure*}

\begin{figure*}[t!]
	\begin{center}{
			\includegraphics[width=0.3\textwidth]{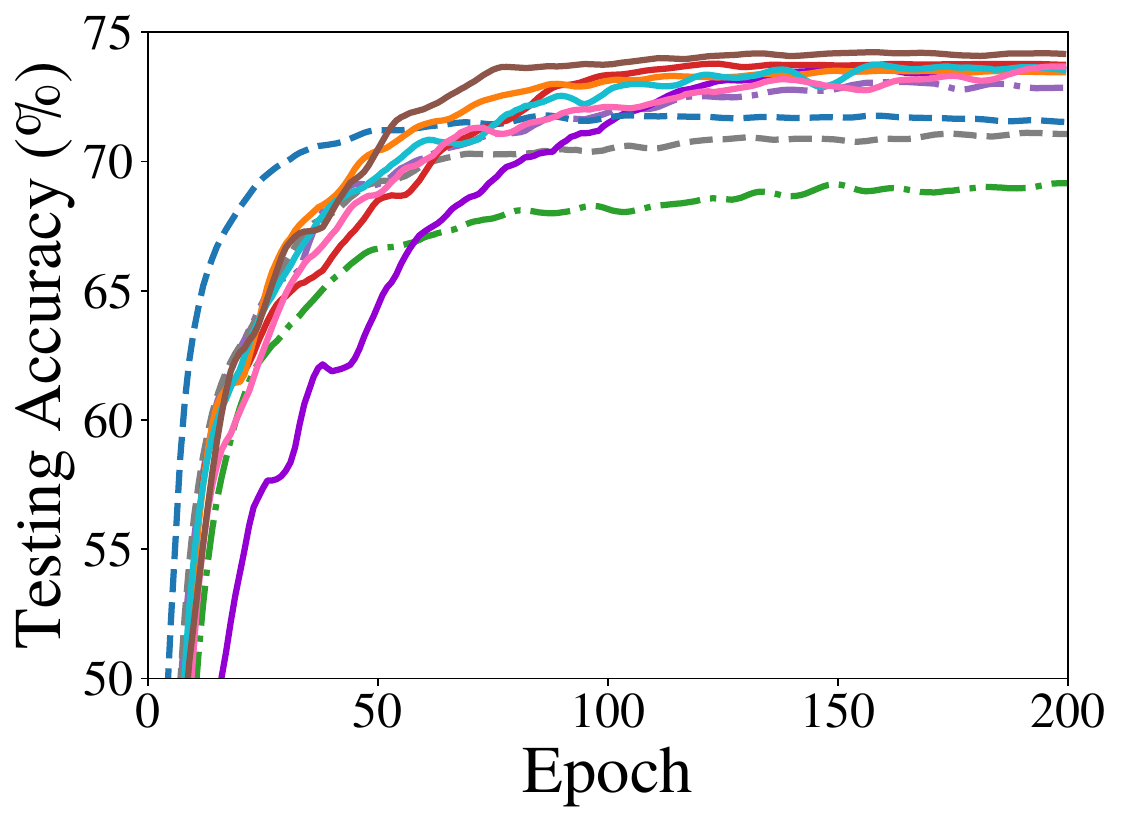}}
            \includegraphics[width=0.3\textwidth]{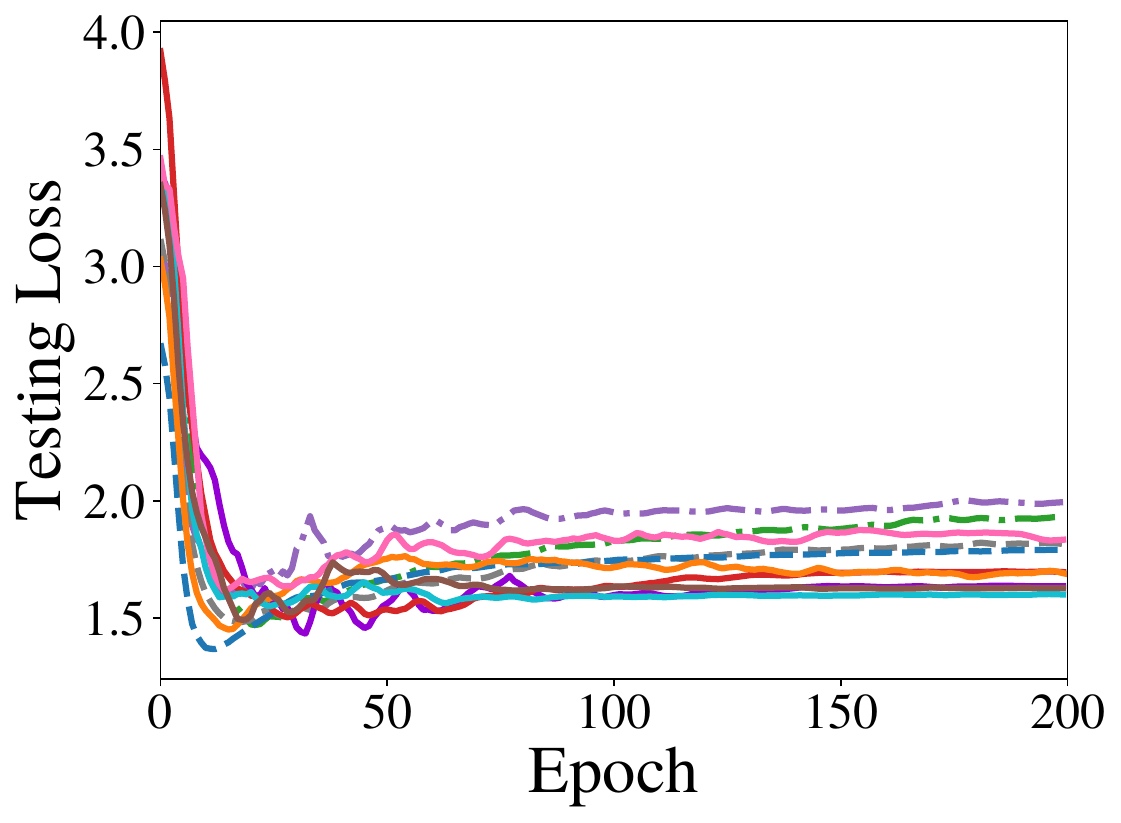}
{\includegraphics[width=0.3\textwidth]{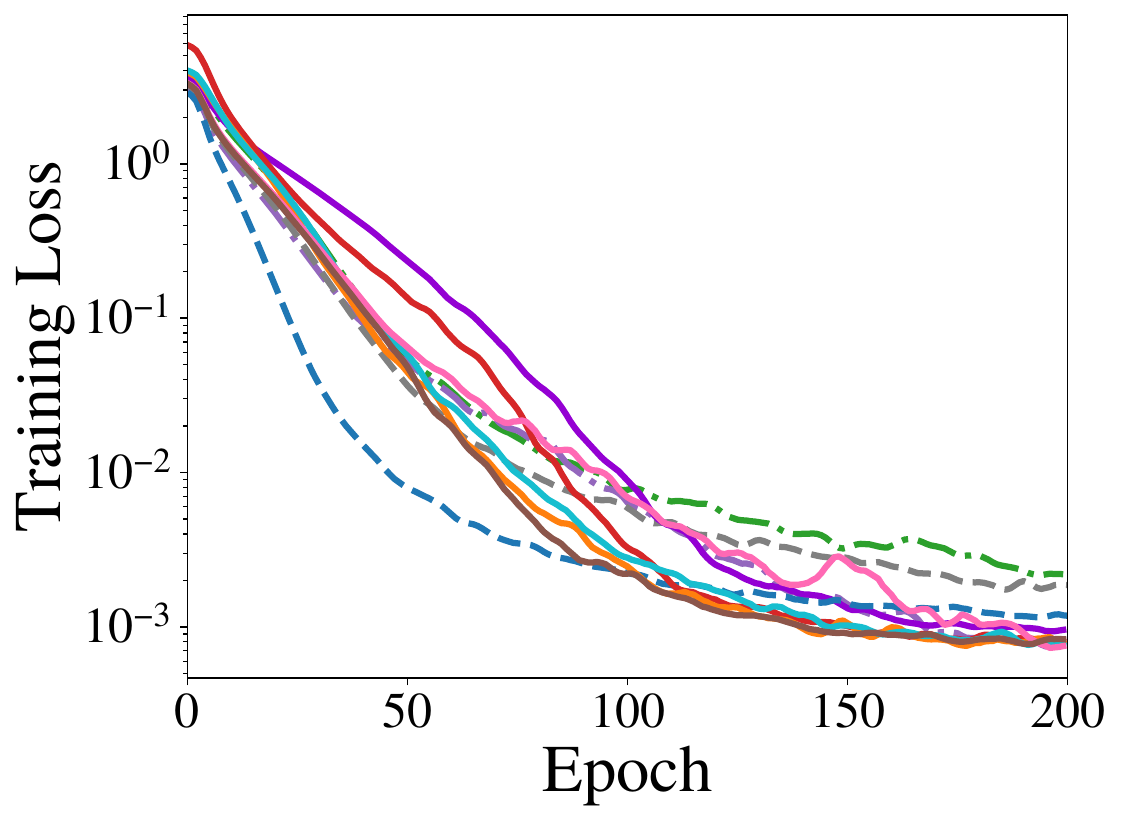}}
		\subfigure{
			\includegraphics[width=0.98\textwidth]{./exp/legend2.pdf}
		}
		\caption{Results for the CIFAR-100 dataset.}
		\label{fig:6}
	\end{center}
\end{figure*}

Figures~\ref{fig:5} and~\ref{fig:6} display the training loss and 
test accuracy against the number of iterations on CIFAR-10 and 
CIFAR-100, respectively. In both metrics, FTSL-Shampoo attains 
performance comparable to that of the full Shampoo, while 
substantially reducing memory consumption and running time, in line 
with our theoretical guarantees. Meanwhile, FTFSL converges faster 
than competing sketching-based methods, corroborating the 
effectiveness of the primal-dual formulation. We further report the 
per-iteration running time of each method: FTFSL exhibits a clear 
reduction over S-ADA, consistent with the improved time complexity. Finally, although S-ASGO 
maintains only a single preconditioner along one axis of the 
gradient, it remains competitive in both training loss and test 
accuracy with its two-preconditioner counterpart FTSL-Shampoo, and 
achieves a further reduction in wall-clock time. This observation 
is consistent with our theoretical analysis, which predicts that 
one-sided preconditioning suffices to capture the dominant curvature 
information while lowering the per-iteration cost.

\begin{figure*}[t!]
	\begin{center}{\includegraphics[width=0.3\textwidth]{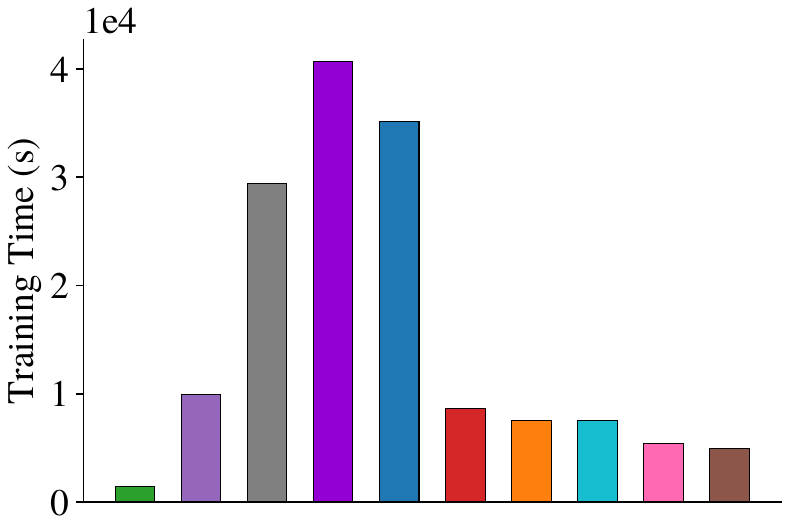}}
   {\includegraphics[width=0.3\textwidth]{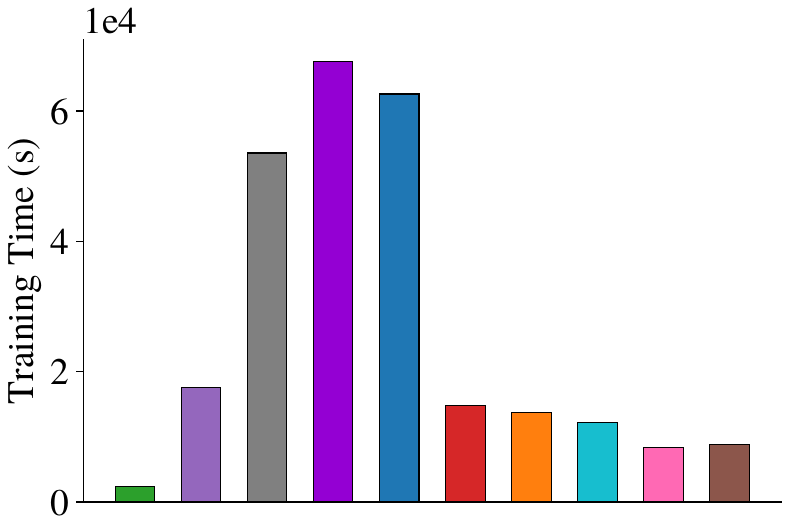}}
		\subfigure{
			\includegraphics[width=0.98\textwidth]{./exp/legend2.pdf}
		}
		\caption{Results of the running time.}
		\label{fig:7}
	\end{center}
\end{figure*}

\subsection{Language Model Pretraining}\label{pllm}

In this section, we empirically evaluate the efficiency of S-ASGO on a large language model (LLM) pretraining task. We compare our method with Muon \citep{jordan2024muon}, Shampoo \citep{gupta2018shampoo}, ASGO \citep{an2025asgo}, and AdamW \citep{kingma2014adam,loshchilov2018decoupled}. All experiments are implemented based on the official implementation of ASGO.\footnote{\url{https://github.com/infinity-stars/ASGO}}

Following prior work \citep{an2025asgo,xie2025talegeometriesadaptiveoptimizers}, we develop a practical variant of S-ASGO that adopts several modifications commonly used in modern optimizers, presented in Algorithm \ref{alg:sasgo2}. Specifically, we maintain the preconditioner with an exponential moving average and incorporate momentum. We use an approximate SVD to decompose the sketching matrix, which reduces the per-step computation. Moreover, we normalize the update of S-ASGO by its root mean square (RMS) norm and then multiply it by $0.2$, so that its RMS norm matches that of the Adam update \citep{liu2025muon}.

\begin{algorithm}[t]
   \caption{A Practical Implementation of S-ASGO}
	\label{alg:sasgo2}
	\begin{algorithmic}[1]
     \REQUIRE Learning rate schedule $\{\eta_t\}_t$, sketching size $\tau$, $\epsilon >0$, momentum parameters $\beta_1, \beta_2 \in [0,1)$, batch size $M$
     \STATE Initialize $X_1 = \textbf{0}_{m \times n}, \hat{L}_0 = \textbf{0}_{m\times \tau}, \tilde{L}_0 = \textbf{0}_{m\times m}, \overline{G}_0^{X} = \textbf{0}_{m\times n}$
     \FOR{ $t=1$ to $T$}
     \STATE Sample a batch of $M$ samples uniformly
     \STATE Query the gradient $G_t^X = \frac{1}{M}\sum_{i=1}^M \nabla f(X_t;\xi_{t,i})$ 
     \STATE Update $M_t = \beta_1M_{t-1}+(1-\beta_1)G_t^X$
     \STATE Send $\beta_2\hat{L}_{t-1}$ and $(1-\beta_2) G_t^X(G_t^X)^{\top}$ to Algorithm \ref{alg:fd2} and receive $\hat{L}_t, \rho^L_{t}$
     \STATE Denote $\tilde{L}_t =  \hat{L}_t \hat{L}_t^{\top} +(\epsilon + \rho_{1:t}^L)I_{m\times m}$
     \STATE Update $X_{t+1} = X_t - \eta_t \frac{0.2\sqrt{mn}\tilde{L}_t^{-1/2}M_t}{\Norm{\tilde{L}_t^{-1/2}M_t}_F}$
     \ENDFOR
	\end{algorithmic}
\end{algorithm}

\begin{figure*}[t!]
	\begin{center}{
			\includegraphics[width=0.45\textwidth]{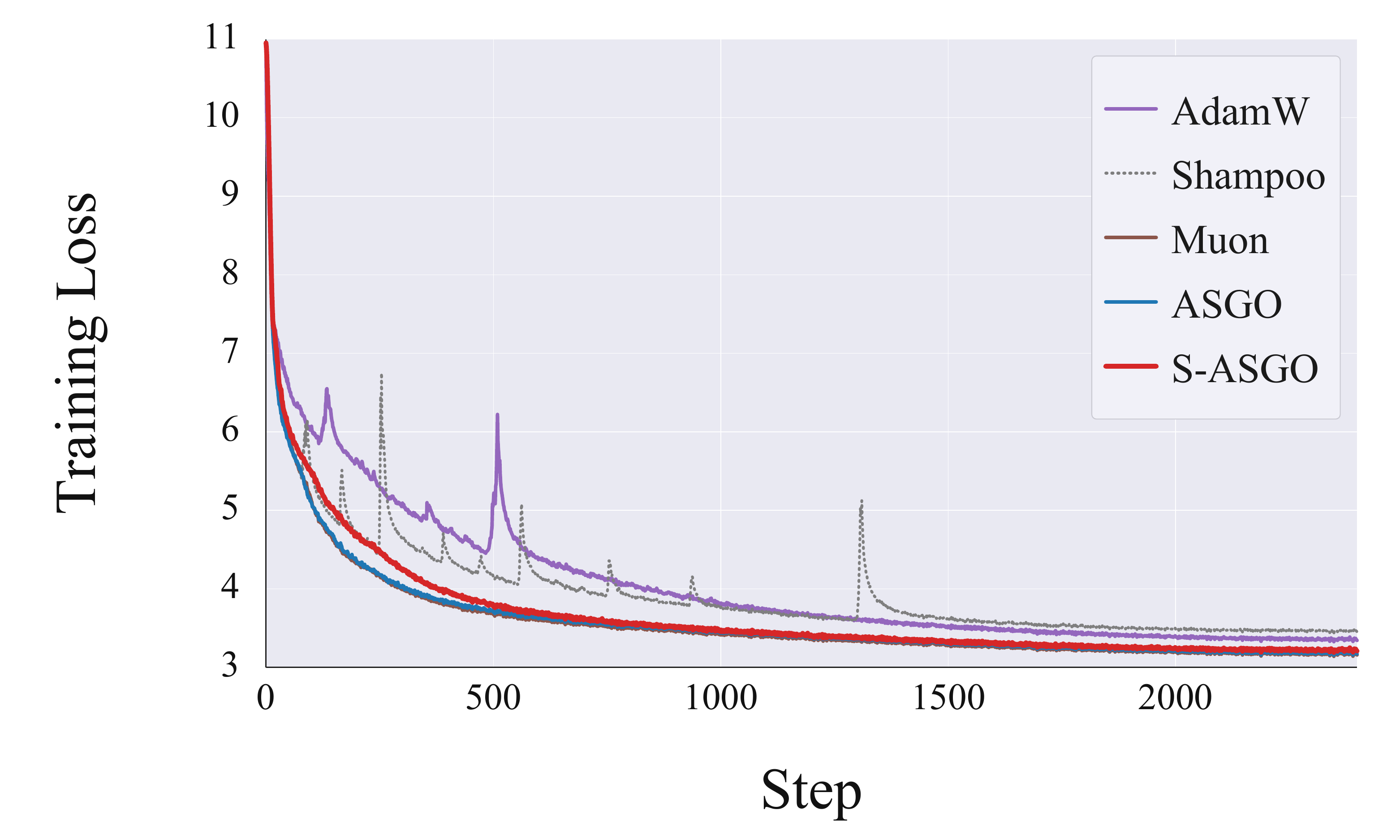}
            \includegraphics[width=0.45\textwidth]{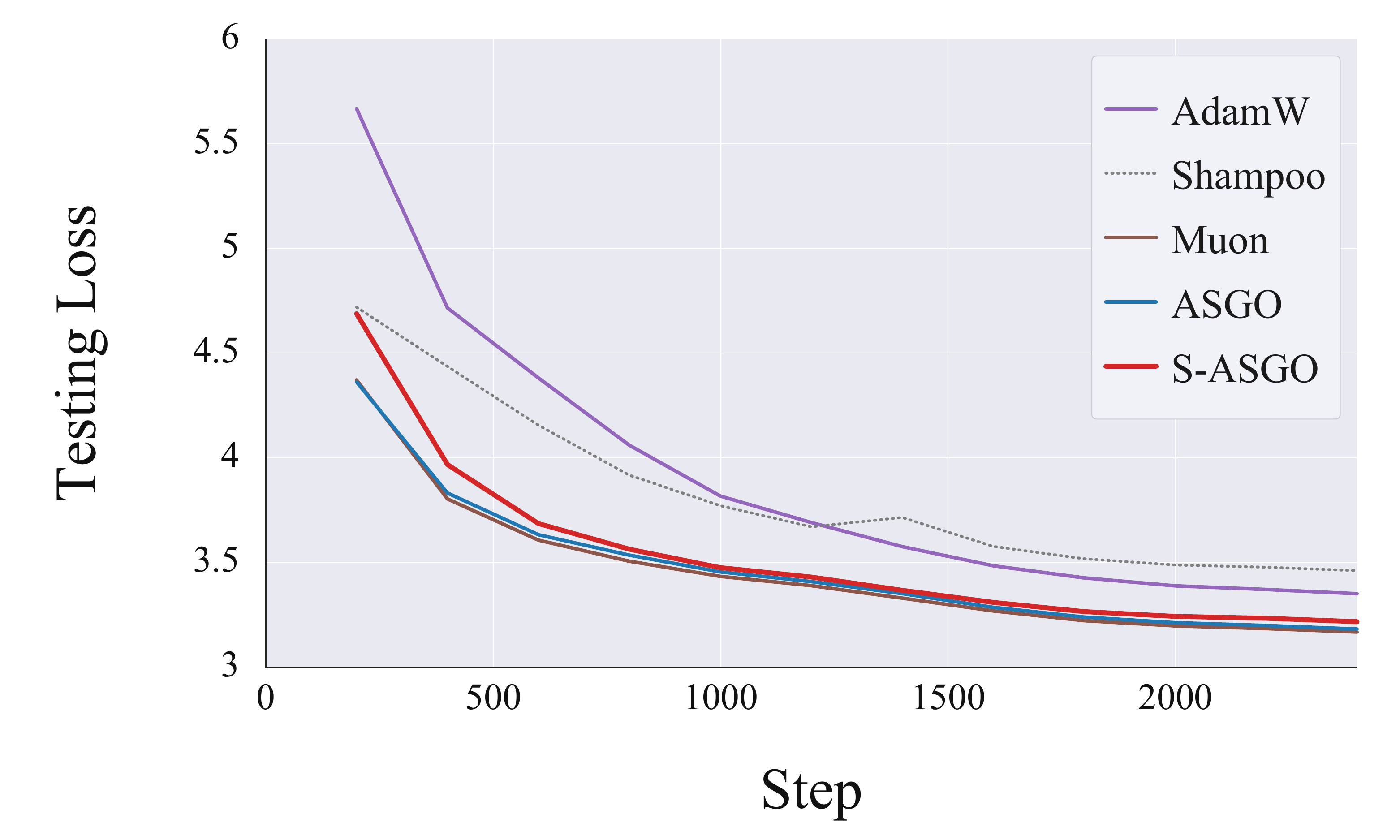}}
		\caption{Results for GPT-2 Pretraining with a sequence length of $512$.}
		\label{fig:11}
	\end{center}
\end{figure*}

\begin{figure*}[t!]
	\begin{center}{
			\includegraphics[width=0.45\textwidth]{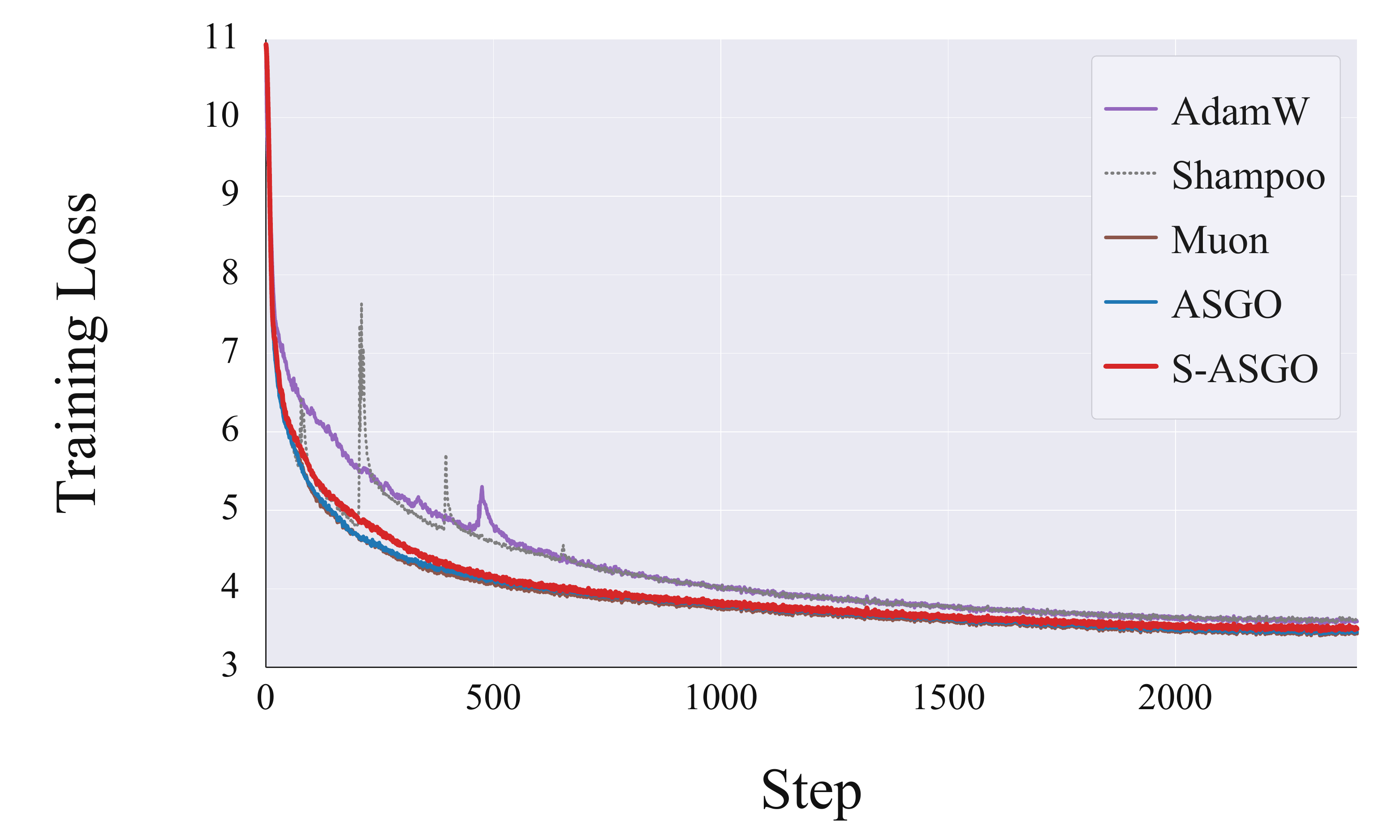}
            \includegraphics[width=0.45\textwidth]{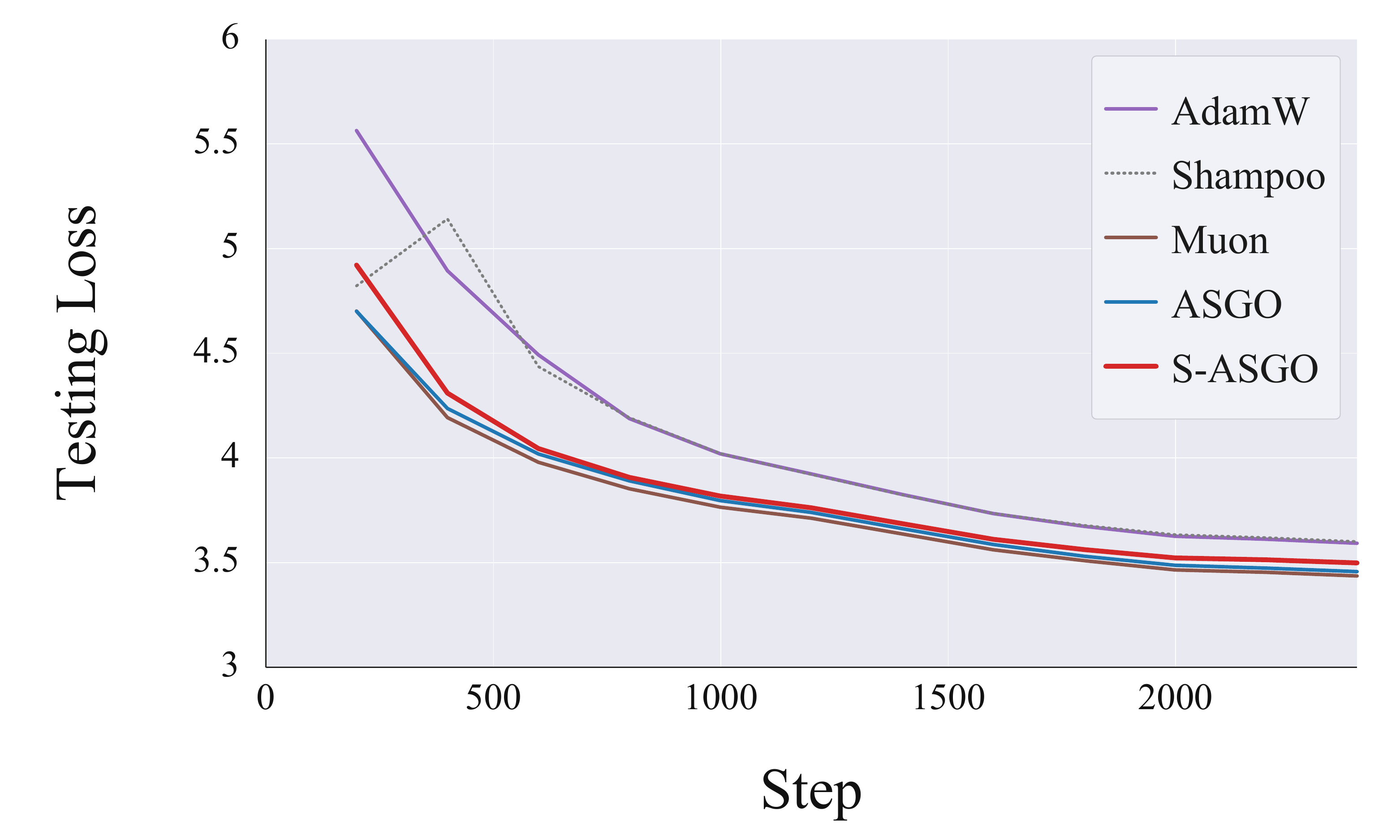}}
		\caption{Results for GPT-2 Pretraining with a sequence length of $256$.}
		\label{fig:12}
	\end{center}
\end{figure*}

\paragraph{Experimental Setup.} We adopt the configuration of the GPT-2 model from  \citet{karpathy2022nanogpt}, which consists of $12$ Transformer layers, $12$ attention heads, and an embedding dimension of $768$. We train on the OpenWebText dataset for a total of $2400$ steps, using Distributed Data Parallel (DDP) across $4$ NVIDIA RTX 4090 GPUs. The learning rate follows a linear warmup over the first $200$ steps and is then decayed by cosine annealing to a final value of $10^{-5}$. We consider two settings, one with a sequence length of $256$ tokens, a batch size of $32$ and gradient accumulation over $8$ steps, and the other with a sequence length of $512$ tokens, a batch size of $16$ and gradient accumulation over $32$ steps. Following common practice \citep{jordan2024muon,an2025asgo}, we train the 1D and embedding-layer parameters with AdamW \citep{loshchilov2018decoupled}. For S-ASGO, we set
\begin{equation*}
    \tau = \max\{\lceil m \times \omega \rceil, 2\},
\end{equation*}
where $\omega$ is the sketching ratio and $\omega=0.2$ in our experiments. For the other methods, we follow the implementation from \citet{an2025asgo}. For the hyperparameters, For hyperparameters, we adopt the recommended settings from \citet{an2025asgo} and perform a grid search for S-ASGO (M). Specifically, we set the learning rate $0.01$ for AdamW and $0.005$ for S-ASGO (M), with $\beta_1 = 0.9$ and $\beta_2 = 0.8$.

\paragraph{Experimental Results.} We evaluate S-ASGO (M) against several baselines on the GPT-2 pretraining task. As shown in Figures \ref{fig:11} and \ref{fig:12}, S-ASGO (M) attains performance comparable to that of Muon and ASGO. More importantly, S-ASGO (M) maintains a much smaller preconditioner than ASGO, because it keeps only a low-rank preconditioning matrix rather than the full one. Consequently, its optimizer memory is governed by the sketching size $\tau$ rather than by the full preconditioner dimension, which reduces the memory footprint while leaving the pretraining loss essentially unchanged. 
These results show that the sketched preconditioner retains the benefit of axis-wise preconditioning at a substantially lower memory cost.

\subsection{Ablation Study on the Sketching Size}\label{sec:ablation-tau}
In this section, we study the sensitivity of our methods to the sketching size $\tau$. As $\tau$ controls the rank of the FD matrix, it governs a trade-off between approximation quality and cost. On one hand, a larger $\tau$ yields a smaller discarded mass $\rho_{1:T}$ and hence a tighter sketching-error term in our regret bounds. On the other hand, a smaller $\tau$ lowers the per-round memory and running time. To examine this trade-off empirically, we adopt the same experimental setup as the GPT-2 pretraining in Section~\ref{pllm}.

\begin{figure*}[t]
	\begin{center}{
			\includegraphics[width=0.45\textwidth]{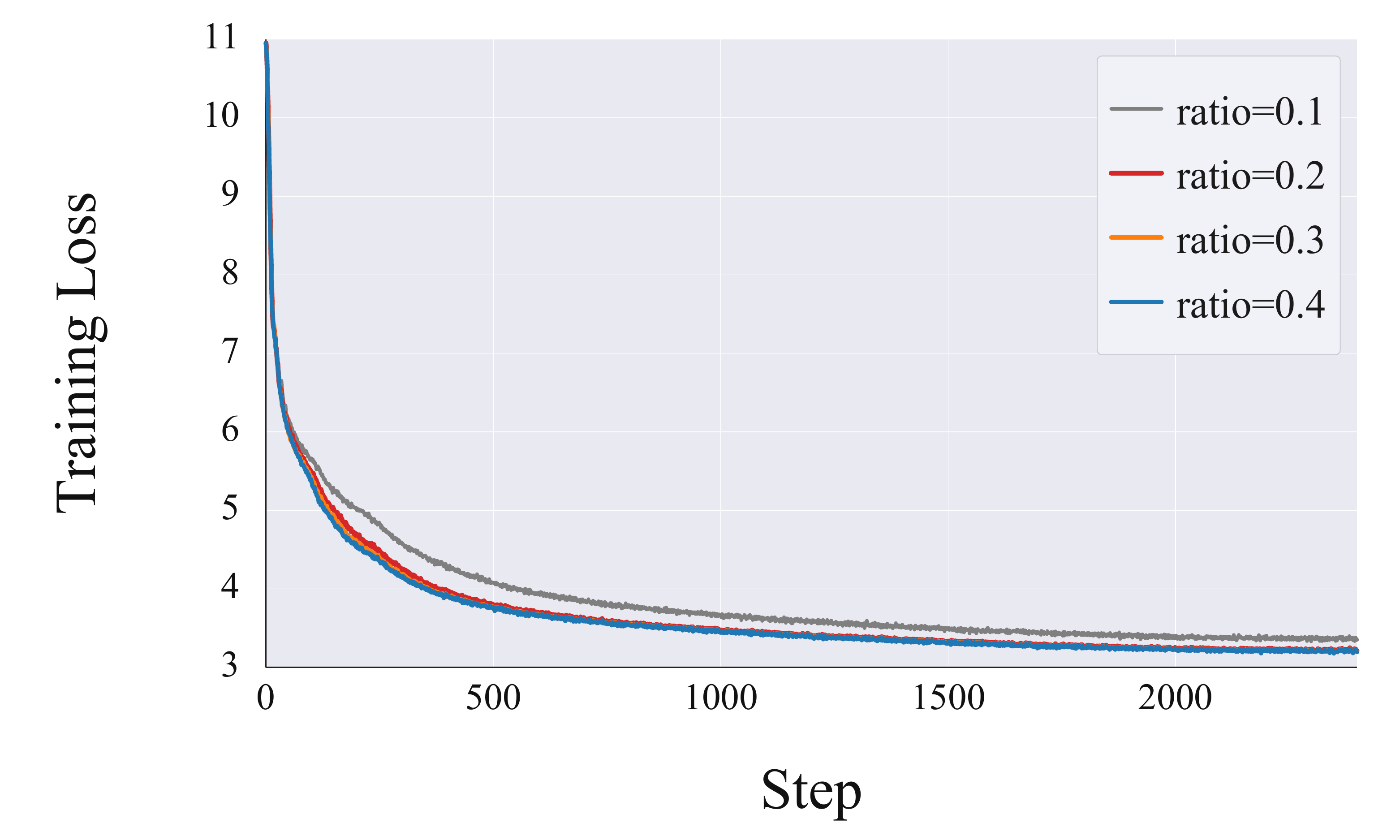}
            \includegraphics[width=0.45\textwidth]{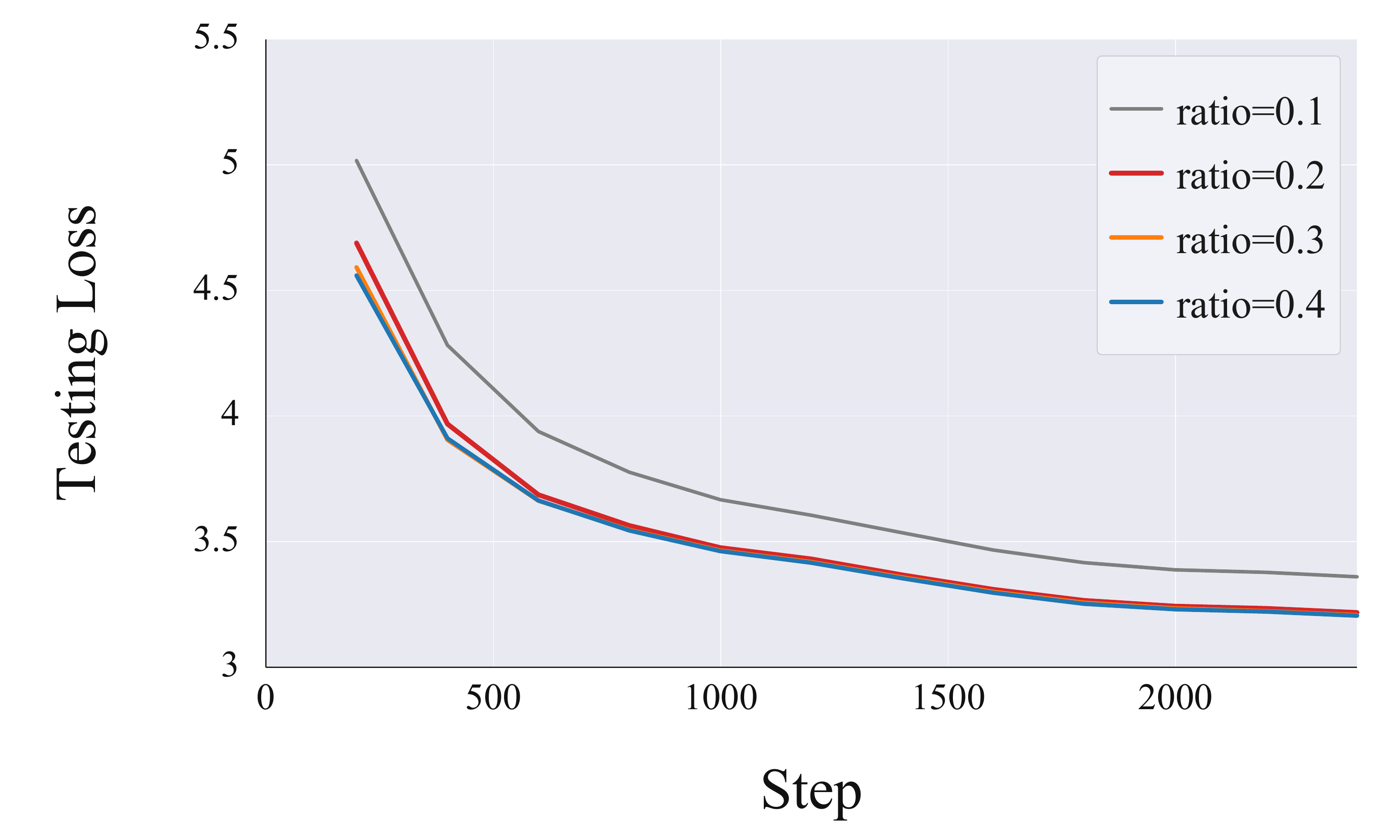}}
    \caption{Ablation Study on the Sketching Size.}
    \label{fig:ablation}
    	\end{center}
\end{figure*}
As shown in Figure~\ref{fig:ablation}, a larger sketching ratio improves the performance on the GPT-2 pretraining task, with the gain becoming marginal once the ratio exceeds $0.2$. This indicates that a sketching ratio of $0.2$ can capture the principal components of the gradients. Although a smaller ratio of $0.1$ performs slightly worse on this task, it is already sufficient on the CIFAR-10 and CIFAR-100 tasks. We conjecture that this difference stems from the rank of the gradients in the two domains. In image classification, only a small portion of each image is relevant to the label, so the gradients tend to be low-rank, and a small $\tau$ already captures them. In contrast, the gradients in language modeling exhibit a higher effective rank, which calls for a larger $\tau$.

\section{Conclusion and Future Work}\label{section8}
In this paper, we investigate adaptive preconditioned optimization methods via frequent directions (FD) and develop several efficient adaptive optimization algorithms. We first introduce FTSL, which achieves  a tighter dimension-free regret bound of  $O(\tr(G_T^{1/2}) + \sqrt{\rho_{1:T}})$ by integrating FD with ADA-FULL under the primal-dual framework. We then propose an  accelerated variant of FTSL that reduces the time complexity to $O(\tau d)$ while preserving the same order regret bound. 
We next consider a more general setting of optimization over matrix-valued decision variables, where the matrix structure can be exploited to design more efficient adaptive preconditioners. In this setting, we first adapt FD to Shampoo under the primal-dual framework, leading to an improved dimension-free regret bound. We further propose S-ASGO and analyze it under both the mirror descent and primal-dual frameworks, establishing improved regret guarantees with lower computational complexity. Furthermore, we extend our analysis to offline stochastic optimization for convex non-smooth losses and convex smooth losses. Finally, we design practical variants of our methods and verify the effectiveness and efficiency of all methods empirically.

Two directions are particularly worth pursuing in future work. 
First, for sparse gradient matrices, the sparse frequent directions technique 
\citep{spafd} attains a better time complexity than standard FD. This suggests 
a natural way to further accelerate our algorithms when the observed gradients 
are sparse.  Second, our theoretical analysis is developed for convex loss functions in the offline stochastic optimization setting, whereas the loss functions of many practical machine learning applications are inherently non-convex. Developing an analysis for non-convex smooth objectives remains an important question.

\appendix

\section{Proof of Theorem \ref{thm:p1}}

The proof of  Theorem \ref{thm:p1} can be derived from the proof of Theorems \ref{thm:ffd1} and \ref{thm:4}.

According to Assumption \ref{convex}, we have
\begin{equation}\label{eq:16}
    \frac{1}{T}\sum_{t=1}^T\E\left[f(\x_t)-f(\x^*)\right]\leq\frac{1}{T}\underbrace{\sum_{t=1}^T\E\left[\left\langle\g_t,\x_t - \x^*\right\rangle\right]}_{R(T)},
\end{equation}
which means that we can derive the convergence rate by using the regret bound of the online methods. Therefore, by incorporating Theorems~\ref{thm:ffd1} and~\ref{thm:4} into~\eqref{eq:16}, we can obtain Theorem~\ref{thm:p1}.

\subsection{Proof of Theorem \ref{thm:p4}}

The proof of  Theorem \ref{thm:p4} can be derived from the proofs of Theorems \ref{thm:jmlr1} and \ref{thm:jmlr1md}.

According to Assumption \ref{convex}, we have
\begin{equation}\label{eq:22}
    \frac{1}{T}\sum_{t=1}^T\E\left[f(X_t)-f(X^*)\right]\leq\frac{1}{T}\underbrace{\sum_{t=1}^T\E\left[\left\langle \hat{G}_t^X,X_t - X^*\right\rangle\right]}_{R(T)},
\end{equation}
which means that we can derive the convergence rate by using the regret bound of the online methods. Therefore, by incorporating Theorems~\ref{thm:jmlr1} and~\ref{thm:jmlr1md} into~\eqref{eq:22}, we can obtain Theorem~\ref{thm:p4}.

\section{Calculation of $\tilde{G}_t^{-1/2}$}
In FTSL, we need to calculate $\tilde{G}_t^{-1/2}$, which cannot be directly derived from the Woodbury formula  \citep{hager1989updating}. We provide the following process.

Assume  that $U_t^{\bot}$ is the complementary subspace of $U_t \in \mathbb{R}^{d\times \tau}$, then we have $I_{d\times d} = U_t U_t^{\top} + U_t^{\bot}(U_t^{\bot})^{\top}$. We denote $\operatorname{diag}(\lambda_{[1:\tau]}^{(t)} - \lambda_{\tau}^{(t)}) = \Sigma_t \in \mathbb{R}^{\tau \times \tau}$ and have

\begin{align*}
    \rho_{1:t}I_{d\times d} +  U_t \Sigma_tU_t^{\top} &= U_t(\Sigma_t+\rho_{1:t}I_{\tau\times \tau}) U_t^{\top} + \rho_{1:t}U_t^{\bot}(U_t^{\bot})^{\top}\\
   U_t(\Sigma_t+\rho_{1:t}I_{\tau\times \tau}) U_t^{\top} &= \rho_{1:t}I_{d\times d} +  U_t \Sigma_tU_t^{\top} - \rho_{1:t}U_t^{\bot}(U_t^{\bot})^{\top}.
\end{align*}

Then we can get 
\begin{align*}
   \rho_{1:t} I_{d\times d} + U_t\Sigma_tU_t^{\top}& = U_t(\Sigma_t+\rho_{1:t}I_{\tau\times \tau}) U_t^{\top} + \rho_{1:t}U_t^{\bot}(U_t^{\bot})^{\top} = [U_t;U_t^{\bot}] \Sigma^{\prime}_t[U_t;U_t^{\bot}]^{\top},
\end{align*}

where $\Sigma_t^{\prime} = \left[\begin{matrix}
 \Sigma_t+\rho_{1:t}I_{\tau \times \tau} & \textbf{0} \\
\textbf{0} & \rho_{1:t}I_{d-\tau \times d-\tau}
\end{matrix}\right]$.

Therefore, we have 
\begin{align*}
    \sqrt{\rho_{1:t} I_{d\times d} + U_t\Sigma_tU_t^{\top}} &= U_t\sqrt{(\Sigma_t+\rho_{1:t}I_{\tau\times \tau})}U_t^{\top} + \sqrt{\rho_{1:t}}U_t^{\bot}(U_t^{\bot})^{\top}\\
    &= \sqrt{\rho_{1:t}}I_{d\times d} + U_t(\sqrt{\Sigma_t+\rho_{1:t}I_{\tau \times \tau}}-\sqrt{\rho_{1:t}I_{\tau \times \tau}})U_t^{\top}.
\end{align*}

By applying the Woodbury formula, we can derive 
\begin{equation*}
\begin{aligned}
    &(\rho_{1:t} I_{d\times d} + U_t\Sigma_tU_t^{\top})^{-1/2}\\
    &= \frac{1}{\sqrt{\rho_{1:t}}}\left(I_{d\times d} - U_t\left(\sqrt{\Sigma_t+\rho_{1:t}I_{\tau \times \tau}}\right)^{-1}\left(\sqrt{\Sigma_t+\rho_{1:t}I_{\tau \times \tau}}-\sqrt{\rho_{1:t}I_{\tau\times \tau}}\right)U_t^{\top}\right).
\end{aligned}
\end{equation*}
\newpage
\bibliography{ref}

\end{document}